\PassOptionsToPackage{table}{xcolor}
\documentclass[conference]{IEEEtran}
\usepackage{cite}
\usepackage{xcolor}
\usepackage{hyperref}
\hypersetup{
  pdfauthor={},
  colorlinks,
  linkcolor={green!80!black},
  citecolor={red!70!black},
  urlcolor={blue!70!black}
}
\usepackage{xurl}
\usepackage{placeins}
\usepackage{graphicx}
\usepackage{booktabs}
\usepackage{amsmath}
\usepackage{amsfonts}
\usepackage{nicefrac}
\usepackage{algorithm}
\usepackage{algorithmic}
\usepackage{float}
\makeatletter
\let\IEEEoriginalmakecaption\@makecaption
\makeatother
\usepackage{subcaption}
\makeatletter
\let\@makecaption\IEEEoriginalmakecaption
\makeatother
\usepackage{comment}
\usepackage{placeins}

\usepackage{multirow}
\usepackage{array}
\usepackage[capitalize,noabbrev]{cleveref}

\begin{document}

\newcommand{\tit}[1]{\smallbreak\noindent\textbf{#1.}}
\newcommand{\tinytit}[1]{\noindent\textbf{#1.}}
\newcommand{\ie}{i.e.\ }
\newcommand{\eg}{e.g.\ }
\newcommand{\myverbatim}[1]{\texttt{\small\detokenize{#1}}}
\date{}
\title{Certification of Real Images through Calibrated Content Authentication}
\author{\IEEEauthorblockN{Sarim Hashmi, Abdelrahman Elsayed, Mohammed Talha Alam, Samuele Poppi, Nils Lukas}
\IEEEauthorblockA{\textit{Mohamed bin Zayed University of Artificial Intelligence}\\
Abu Dhabi, UAE\\
\{sarim.hashmi, abdelrahman.elsayed, mohammed.alam, samuele.poppi, nils.lukas\}@mbzuai.ac.ae}}

\maketitle

\begin{abstract}
Generative models can synthesize high-quality inauthentic multimedia content that is already being misused at scale.
We evaluate twenty deepfake detectors against ten generators released in the last four years and find accuracy decreasing over time, from near-perfect 99.5\% to 76\%.
Adversarial perturbations further reduce every baseline detector to below 2\% accuracy, effectively inverting the detector's assigned label.
We argue that this unreliability reflects a fundamental ambiguity: generators can reproduce authentic content exactly (e.g., through memorization), so content alone cannot reveal the true provenance label.
For this reason, content produced by a generator must admit a faithful reconstruction by that same generator, and finding such a reconstruction makes synthetic provenance plausible and authenticity \emph{plausibly deniable}.
We therefore propose and evaluate a detection paradigm that outputs a calibrated prediction of whether authenticity is \emph{plausibly deniable}: a faithful reconstruction by any known generator establishes plausible deniability, while calibration bounds how often content from known generators fails to be reproduced.
Our evaluation shows that (i) our detector can be calibrated so that at most 1\% of generated content is wrongly certified, an operating point at which most baseline detectors reach near-zero recall, including the strongest with 93\% accuracy; (ii) calibrating a stricter security threshold on attacked samples preserves this bound against adaptive adversaries within the evaluated bounded-perturbation attack space, whose perturbations break every baseline, but does not cover arbitrary adversarial transformations; and (iii) post-hoc verifiability is eroding, as 1{,}116 of 3{,}000 Reddit images resist reproduction by a 2022 generator, but only 55 to 79 resist reproduction by 2024 generators.
We release our \href{https://github.com/Sarim-MBZUAI/content-authentication}{code} to enable calibration against future generators.
\end{abstract}

\section{Introduction}
\label{sec:intro}

Generative models can synthesize \emph{inauthentic} multimedia content that can be falsely claimed to be authentic.
Open-weight releases such as SD3~\cite{esser2024sd3} and FLUX~\cite{flux2024} put this capability in anyone's hands, with no provider observing how it is used.
For example, during a June 2025 conflict in the Middle East, AI-generated videos of missile strikes flooded social media, and the three most viral clips alone drew over 100 million views~\cite{bbc2025aiwar}.
When users asked X's AI assistant, Grok, about the clips' authenticity, it returned wrong or contradictory answers for the same video~\cite{dfrlab2025grok}.
However, verification at this scale can only be automated, yet unreliable verifiers might be worse than none, as each wrong prediction lends credibility to inauthentic content.

\begin{figure}
  \centering
  \includegraphics[width=\columnwidth]{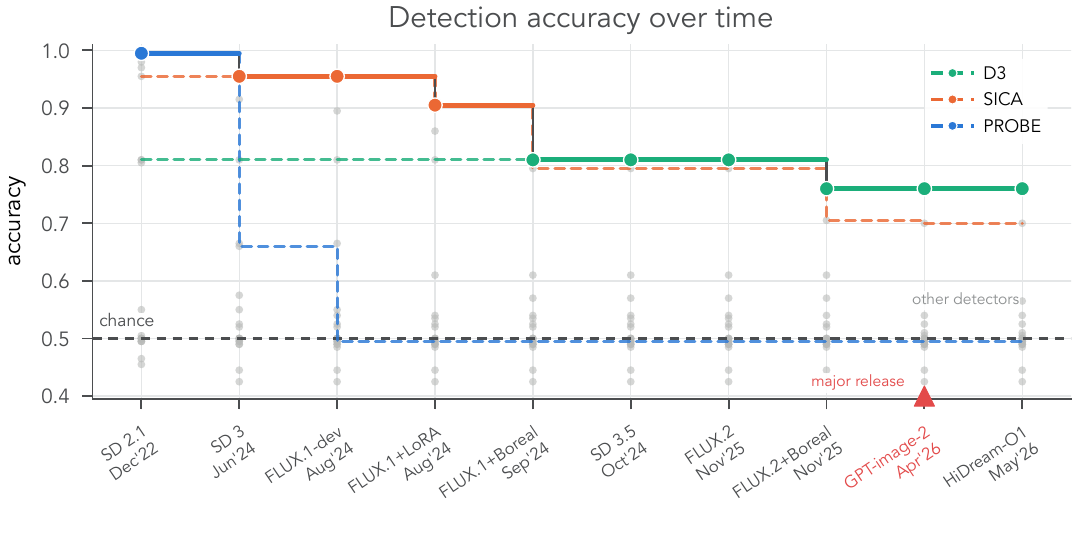}
  \caption{\textbf{The detection game increasingly favors generators.}
  Among detectors available at each generator's release, the best accuracy falls from 99.5\% on 2022 generators to 76\% on 2026 generators, suggesting that generators are winning.}
  \label{fig:teaser}
\end{figure}

To establish content provenance, providers have begun to attach provenance signals at generation time to detect content generated by their platforms.
\begin{figure*}[!t]
  \centering
  \includegraphics[width=0.99\textwidth]
  {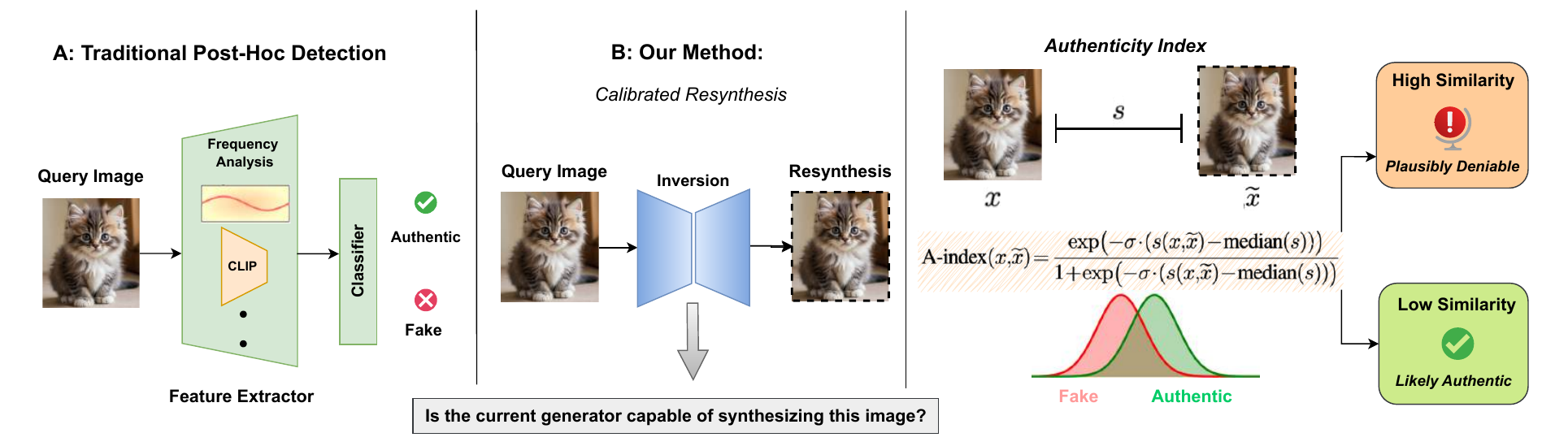}
  \caption{
    \textbf{Conceptual illustration of our method.}
    (A) Traditional post-hoc detectors separate real from fake using feature cues but struggle as generators improve.
    (B) Our method instead tests whether a generator can resynthesize the query image $x$.
    The similarity $s(x,\tilde{x})$ between $x$ and its resynthesis $\tilde{x}$ is calibrated into an authenticity score, where high similarity implies \emph{plausible deniability} and low similarity indicates \emph{likely authentic}. In this and subsequent figures, \emph{A-index} denotes the authenticity score.
  }
  \label{teaserv2}
\end{figure*}
Leading US AI providers have adopted provenance measures: Google embeds SynthID watermarks in generated content~\cite{googleSynthID}, and OpenAI attaches Content Credentials to generated images~\cite{openai2026provenance}.
In 2023, Amazon, Anthropic, Google, Inflection, Meta, Microsoft, and OpenAI committed to developing and deploying provenance or watermarking mechanisms for AI-generated audio and visual content~\cite{whitehouse2023commitments}.
The EU AI Act imposes marking and disclosure requirements for AI-generated content from August 2026~\cite{euaiact2024}, but these measures (i) require provider cooperation, (ii) can reduce output quality when embedding watermarks~\cite{freqmark2024}, and (iii) rely on watermarks whose robustness is not assured, with demonstrated vulnerabilities to model fine-tuning~\cite{stableSignatureUnstable2024}.
In fact, watermarks can be removed by fine-tuning open-weight models~\cite{lukas2023leveraging,diaa2025optimizing}, metadata is stripped by common processing, and anyone can run an unwatermarked, open-weight generator.
The threat is a user who synthesizes fabricated content with such a generator and distributes it as authentic, leaving no provenance signal to detect.
For content from uncooperative or open-source generators, post-hoc detection remains the only viable defense.

Deepfake detectors are a promising defense as they do not require the provider's cooperation~\cite{ufd,d3,lin2024survey}, but their accuracy drops over time as new, more capable, generators are being developed.
Our benchmark (see \Cref{app:benchmark}) spans ten generators released between December 2022 and May 2026, from SD2.1~\cite{rombach2021ldm} through SD3~\cite{esser2024sd3}, SD3.5~\cite{stability_sd35}, and the FLUX.1~\cite{flux2024} and FLUX.2~\cite{flux2_dev} families, including LoRA-adapted variants~\cite{xlabs_flux_realismlora,boreal_flux1,boreal_flux2}, to GPT~Image-2~\cite{openai2026chatgptimages2} and HiDream-O1~\cite{hidream_o1}.
Across these four years, we evaluate the top twenty detectors, and find that their accuracy drops from ~99.5\% to ~76\% (\Cref{fig:teaser}).
The best detector itself changes twice, from PROBE~\cite{probe} to SICA~\cite{sica} to D3~\cite{d3}, while most of the remaining detectors perform close to the 50\% chance level on the newest generators (\Cref{fig:teaser}).
To make things worse, under $\ell_\infty$-bounded perturbations with $\epsilon=8/255$, all twenty detectors fall below 2\% accuracy, and even the strongest, D3 \cite{d3}, retains only 1.75\%.
More fundamentally, authenticity cannot be decided from content alone, since a more capable generator may reproduce authentic content exactly, e.g., through memorization~\cite{carlini2023extracting,somepalli2023copying}, so asserting authenticity requires knowledge of both the model and its output.
A prediction that cannot be trusted is worse than no prediction, since each error feeds the \emph{liar's dividend}, where authentic content is dismissed as fake~\cite{deepfakedefense2025}.
We cannot expect complete detectors, but we can build \emph{sound} ones, i.e., detectors that only make claims with a known error rate and abstain otherwise.

We propose a detector that meets this requirement by certifying content as \emph{authentic}.
Unlike the provenance label of a sample, which may not be predictable from the content alone, the reconstruction fidelity under known generators on the sample is measurable and predictive.
Our method, shown in \Cref{teaserv2}, inverts a suspect sample through known generators~\cite{rf_inversion} and measures how faithfully the best reconstruction matches it.
If the reconstruction is faithful, then anyone with that generator could have created the sample, its authenticity is \emph{plausibly deniable}, and our method abstains.
If every generator fails to reconstruct it faithfully, our method certifies the sample as \emph{authentic} relative to the tested generators.
Prior work shows that generated images are reconstructed more faithfully than authentic images~\cite{wang2023dire,ricker2024aeroblade}, which motivates reconstruction fidelity as our authentication signal.
Since our detector can abstain and we fix the actions a user can take in advance, e.g., a single generation, we calibrate the decision threshold so that at most 1\% of generated calibration samples are certified (\Cref{fig:methodology_a_calib}).

We also consider an \emph{adaptive} but restricted adversary who knows the algorithm and perturbs generated disinformation within a bounded set until no generator reconstructs it faithfully, turning a correct abstention into a false certificate.
We therefore calibrate two error-bounding thresholds: a \emph{safety} threshold for non-adversarial users and a stricter \emph{security} threshold for bounded perturbations.
For \texttt{SD3-Medium}, attacking the calibration samples raises the threshold from 0.0365 to 0.038 and preserves the 1\% bound under the same attack.
These perturbations undermine all twenty existing detectors without invalidating our certificates.

Prior work uses reconstruction error for binary classification that labels every input~\cite{wang2023dire,ricker2024aeroblade,luo2024lare2,chu2025fire}.
Selective prediction abstains under uncertainty~\cite{geifman2017selective}, whereas our method abstains when reconstruction shows that a generator can reproduce the sample.
Our contribution is a methodology that calibrates this signal into a sound detector, preserves calibration against adaptive adversaries, and applies to any generator and, as shown for video, multiple modalities.
Prior detectors output uncalibrated scores and require retraining for new generators, whereas our framework requires only recalibration.
Certification takes 11.67 seconds per image and generator on one NVIDIA RTX 5000 Ada versus milliseconds for feed-forward detectors, making it suitable for forensic verification rather than feed-scale screening.

However, soundness has a cost, as every abstention gives up coverage, i.e., the share of authentic content that the detector certifies instead of abstaining, and this cost grows with generator capability.
On 3,000 images collected from Reddit, a generator from 2022 fails to reproduce 1,116 images, while generators from 2024 fail on only 55 to 79.
The certifiable share of this corpus therefore falls from 37.2\% to at most 2.63\%, i.e., by more than $14\times$, within two years of generator releases.
Every generator release therefore shrinks the set of content that anyone can still certify, and we forecast that this erosion will continue.
Our method doubles as the instrument for tracking it, since recalibrating against each new generator measures how much content remains verifiable.

We model efficient adversaries, constrained in the compute available to them, as best-of-100 attackers: they draw 100 images at random from a generator, here from one prompt, keep the one with the highest authenticity score, and refine it with PGD.
Our evaluation shows that (i) our calibrated thresholds hold at 1\% FPR across five generator configurations while most baseline certifies almost nothing at this rate, (ii) twenty-three of twenty-four semantic transformations stay below the safety threshold and the remaining one stays below the security threshold, (iii) a best-of-100 attacker who spends 19.5 minutes of search raises its best candidate only from 0.0148 to 0.0154, and (iv) on 100 in-the-wild videos, frame aggregation preserves the reconstruction signal while no video baseline exceeds 0.615 AUC.
While these thresholds withstand the attacks we calibrate with, a stronger optimizer or a new attack class may exceed them, which a defender counters by recalibrating.

\subsection{Contributions}
\begin{enumerate}
    \item We evaluate twenty deepfake detectors against ten generators released over four years, showing that the best accuracy declines from 99.5\% to 76\% and that all twenty fall below 2\% under adversarial perturbations.
    \item We propose a sound detector that certifies content as authentic only when no known generator can faithfully reproduce it, calibrated so at most 1\% of generated content is wrongly certified.
    At this error rate, most existing detectors reach near-zero recall despite up to 93\% accuracy.
    \item We are the first to calibrate against adaptive adversaries, distinguishing a safety threshold for unaware users from a security threshold for efficient attackers with limited compute and a perturbation budget of $\epsilon=8/255$.
    \item We study 3,000 Reddit images and forecast the erosion of post-hoc verifiability, showing that each generator release shrinks the certifiable set from 1,116 to at most 79 images, and we extend our method to video.
\end{enumerate}

\section{Background \& Related Work}
\label{sec:background}

\tit{Active Content Provenance}
Watermarking embeds a detectable signal at generation time, e.g., in the latent decoder~\cite{fernandez2023stable} or the initial noise~\cite{wen2023treering,yang2024gaussianshading}, and C2PA attaches signed metadata~\cite{c2pa2024}.
Both methods can enable highly accurate and precise detection, but only for providers that cooperate and voluntarily implement these mechanisms.
Content from uncooperative or open-weight generators carries no active provenance signal.
Neither watermarks nor C2PA are adversarially robust and can be removed~\cite{lukas2023leveraging,diaa2025optimizing}.

\tit{Passive Deepfake Detection}
Passive detectors classify content from forensic cues and fall into three families.
Artifact methods detect low-level traces such as spectral anomalies or sensor noise~\cite{frank2020leveraging,lukas2006digital,npr,freqnet}, feature classifiers train a decision head on foundation-model embeddings~\cite{ufd,cozzolino2024clipbar,c2pclip,fatformer}, and reconstruction methods score how well an autoencoder or an older diffusion model reproduces the content~\cite{ricker2024aeroblade,wang2023dire,luo2024lare2,chu2025fire}.
AEROBLADE~\cite{ricker2024aeroblade}, DIRE~\cite{wang2023dire}, and FIRE~\cite{chu2025fire} threshold this reconstruction error into a real-or-fake label for every input.
We instead invert the most capable open-weight generators and calibrate their reconstruction fidelity into a one-sided certificate with a measured error rate.
We evaluate twenty recent detectors (\Cref{app:baselines}), of which three performed best on our benchmark (\Cref{fig:teaser}):
\begin{itemize}
    \item \textbf{PROBE}~\cite{probe} improves a trained detector by generating hard training examples, similar to adversarial training.
    It edits a generator's internal representations to create realistic images that the detector misclassifies and then retrains the detector on them.
    \item \textbf{SICA}~\cite{sica} trains one detector for several types of image forgery, which is hard because every type leaves different subtle patterns that blur together during training.
    It adapts a pretrained vision model with a constraint, based on the image content, that keeps these patterns separated.
    \item \textbf{D3}~\cite{d3} trains one detector on many generators at once, which normally lowers accuracy on some of them.
    To avoid this, it compares every image with a distorted copy of itself and uses the difference as its detection signal.
\end{itemize}
All twenty are binary detectors, i.e., they output a real-or-fake decision for every input.

\tit{Video Deepfake Detection}
Video detectors exploit cues over time rather than within a single frame.
FTCN~\cite{ftcn} removes most spatial processing so the network must learn temporal inconsistencies, GenConViT~\cite{genconvit} combines convolutional and transformer features with a learned latent representation of the frame, and StyleFlow~\cite{styleflow} tracks how style features change across frames.
Like their image counterparts, all three output a real-or-fake decision for every video.

\tit{Generative Models and Inversion}
A text-to-image generator $G$ maps an input $\mathbf{z}$, i.e., a noise latent and a prompt $c$, to content $x = G(\mathbf{z})$.
Diffusion models learn to reverse a gradual noising of data~\cite{ho2020denoising}, typically in the latent space of an autoencoder~\cite{rombach2021ldm}, and flow matching simplifies this into an ordinary differential equation with a learned velocity field $v_\theta$ that moves noise to data along near-straight paths~\cite{lipman2023flow,liu2023rectifiedflow},
\begin{equation}
\frac{\mathrm{d}x_t}{\mathrm{d}t} = v_\theta(x_t, t, c), \qquad x_1 \sim \mathcal{N}(0, I),
\label{eq:flow}
\end{equation}
a design used by current open-weight generators such as SD3~\cite{esser2024sd3} and FLUX~\cite{flux2024}.
Inversion asks the reverse question, i.e., which input would make $G$ reproduce a given sample $x$.
Earlier diffusion inversion required per-sample optimization or fragile backward integration over curved paths~\cite{mokady2023nulltext,wallace2023edict}.
The near-straight paths of \Cref{eq:flow} integrate backwards stably, and RF-Inversion~\cite{rf_inversion} does so without per-sample optimization, yielding a reconstruction
\begin{equation}
\tilde{x} = G\big(\widetilde{G}^{-1}(x)\big),
\label{eq:inversion}
\end{equation}
whose fidelity to $x$ is scored with the similarity metrics of \Cref{sec:methodology}.
This pipeline also admits a differentiable approximation, which our calibration attacks require (\Cref{sec:adversial_attack}).

\section{Threat Model}
\label{sec:threat_model}

\tit{Setting}
Consider a platform that hosts millions of images and videos and wants to label content as authentic or inauthentic, e.g., with a community note.
A wrong label that nobody can trust is worse than saying ``I don't know'' (abstention), so the platform wants to calibrate its false positive rate, and it certifies the few items that go viral or get flagged.
A newsroom verifying footage or a court weighing evidence faces the same trade-off.
The error costs are asymmetric, as a wrong certificate lends fabricated content undeserved credibility, while an abstention only leaves content unlabeled.
The \emph{defender} certifies content as authentic or abstains, and the \emph{adversary} is the party who wants generated content certified.
We consider the certification of images and, in a preliminary study, of video (\Cref{sec:video}).
Our detector exposes two functions:
\begin{itemize}
    \item $a_G(x) \leftarrow \textsc{Score}(G, x)$: inverts a sample $x$ under a generator $G$ and maps the fidelity of the reconstruction $\tilde{x}_G$ to an authenticity score (the A-index) in $[0,1]$.
    \item $\textsc{Certify}(x)$: certifies $x$ as authentic relative to a set of generators $\mathcal{G}$ if $a_G(x) \geq \tau_G$ for every $G \in \mathcal{G}$, and otherwise abstains with the evidence $(\tilde{x}_G, G)$.
\end{itemize}

\tit{Defender Capabilities}
The defender holds a set of open-weight generators $\mathcal{G}$ and an inversion method for each $G \in \mathcal{G}$.
The defender can generate content from every $G \in \mathcal{G}$, limited only by their compute, and knows the class of attacks the adversary may use, i.e., the perturbation type and budget, but not the attacked samples themselves.
The defender needs no access to the adversary's prompt, seed, or generator choice, as certification tests a suspect sample against every $G \in \mathcal{G}$.
This lets the defender apply the adversary's restricted set of perturbations to their own generated content before deployment and calibrate decision thresholds to a chosen error rate $\alpha$.

Certificates are scoped to $\mathcal{G}$, which is meaningful only if $\mathcal{G}$ contains the most capable generators. We therefore assume that they are publicly available, as inversion requires white-box access, and each new generator release requires recalibration.

\tit{Adversary Capabilities}
We consider (1) a normal user who is unaware of the detector and does not try to evade it, which the safety threshold covers, and (2) an adaptive attacker with access to a restricted set of transformations, which the security threshold covers. The adversary is \emph{adaptive}, i.e., they know the detection method, the generator set $\mathcal{G}$, and all calibrated thresholds, as we assume no security through obscurity.
This is the strongest knowledge assumption, and every weaker adversary is a special case of it.
They generate content with a generator $G \in \mathcal{G}$, may select the best among a limited number of candidates, and may add an imperceptible perturbation $\delta$ with $\|\delta\|_\infty \leq \epsilon$ before publishing.
We restrict perturbations to the $\ell_\infty$ norm, as an adversary spreading disinformation must keep the depicted content intact, and visible modifications change what the content shows (\Cref{sec:adversial_attack}).
We consider \emph{efficient} adversaries with limited compute who cannot fine-tune generators, since an adversary with unbounded compute defeats any post-hoc detector.
We price the adversary's search in our evaluation, i.e., selecting the best of $N=100$ candidates costs approximately 19.5 minutes of sequential inversion on one GPU.

\tit{Objectives}
The adversary's success is the rate at which their generated samples are certified as authentic.
We measure the defender's error as the false positive rate, i.e., the probability that a generated sample is certified after a transformation $T \in \mathcal{T}$,
\begin{equation}
\mathrm{FPR} \;=\; \Pr_{G \in \mathcal{G},\, x \sim G}\Big[\,\exists\, T \in \mathcal{T} \colon \textsc{Certify}(T(x))\,\Big],
\label{eq:fpr}
\end{equation}
where $\mathcal{T}$ contains only the identity for the normal user (safety threshold) and the restricted set of the adaptive attacker, e.g., perturbations $\delta$ with $\|\delta\|_\infty \leq \epsilon$, for the security threshold.
We measure the defender's utility as the true positive rate, i.e., the share of authentic content $\mathcal{D}_{\mathrm{auth}}$ that they certify,
\begin{equation}
\mathrm{TPR} \;=\; \Pr_{x \sim \mathcal{D}_{\mathrm{auth}}}\big[\textsc{Certify}(x)\big].
\label{eq:tpr}
\end{equation}
The defender maximizes TPR subject to $\mathrm{FPR} \leq \alpha$, and we compare all detectors by their TPR at a fixed FPR of $\alpha = 1\%$.
An adversary may also perturb authentic content to force an abstention, which lowers TPR but never raises FPR.

\begin{figure*}[!t]
    \centering
    \includegraphics[width=0.99\linewidth]{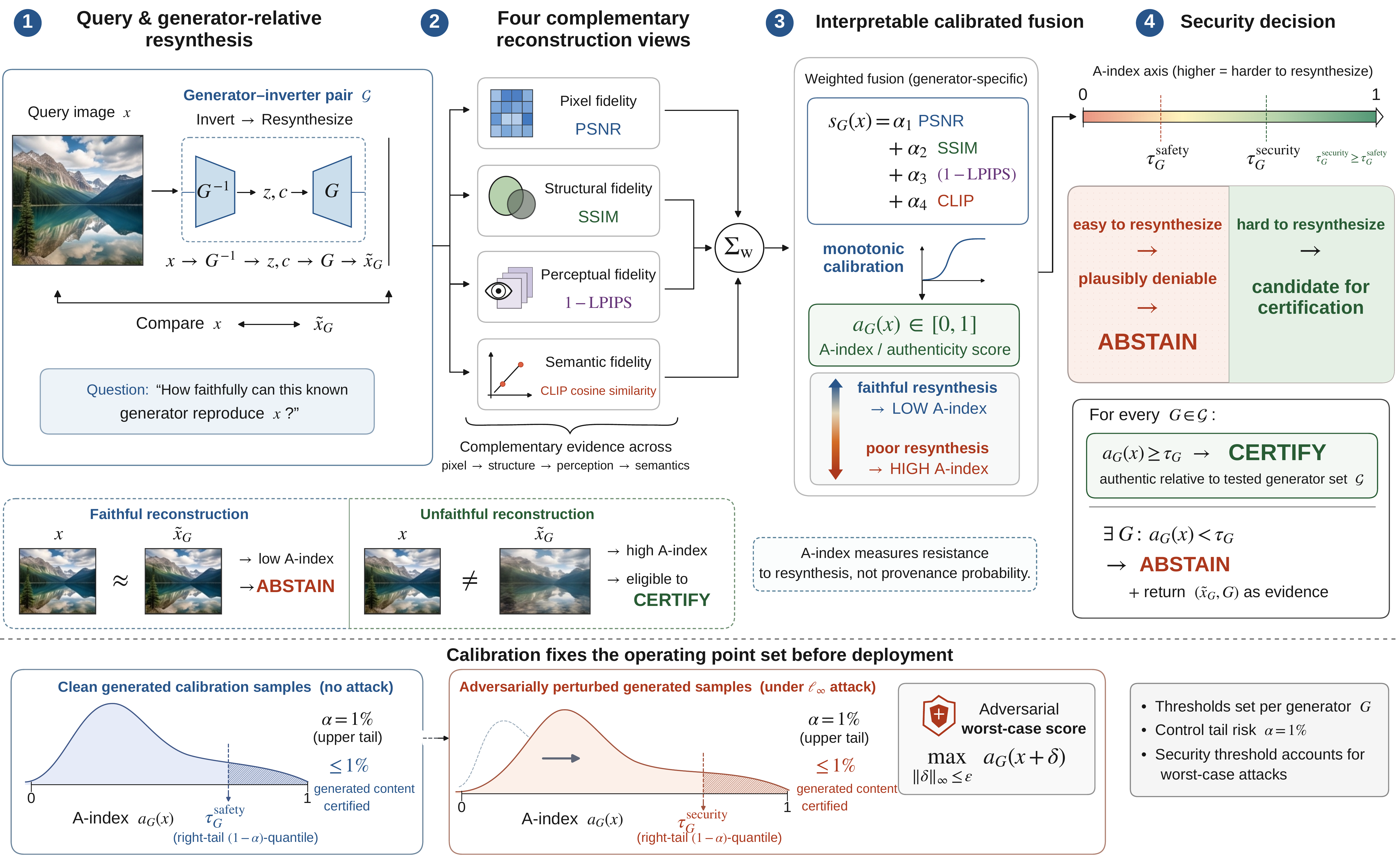}
    \caption{\textbf{Computing of the Authenticity Score.} Given an input image $x$, a reconstruction-free inverter $G_e^{-1}$ produces an inverted reconstruction $\tilde{x}$. We then compute complementary similarities between $x$ and $\tilde{x}$: pixel fidelity\texttt{(PSNR)}, structural fidelity\texttt{(SSIM)}, perceptual distance\texttt{($1{-}$LPIPS)}, and semantic consistency (\texttt{CLIP} cosine). A calibrated weighted combiner (learned $\alpha_1, \alpha_2, \alpha_3, \alpha_4$) produces a scalar $s$($x$, $\tilde{x}$), yielding the A-index in $[0,1]$. A safety threshold $\tau$ certifies content as \emph{Authentic} when $\text{A-index} \geq \tau$, and otherwise labels it as \emph{Plausibly Deniable}. We further analyze robustness by applying $\ell_\infty$-bounded perturbations $\delta$ (PGD-style) through $G_e^{-1}$ to maximize or minimize the A-index.
}
\label{fig:concept}
\end{figure*}

\section{Conceptual Approach}
\label{sec:methodology}

We state four design goals, G1 to G4.
(G1)~Every certificate must come with a measured error rate (\Cref{eq:fpr}).
(G2)~Every abstention must come with evidence that anyone can check.
(G3)~Thresholds must be set on attacked samples, not only clean ones.
(G4)~A new generator must only require recalibration, not retraining.
The rest of this section builds a detector with these four properties.

\subsection{Decision Rule}
Given a suspect sample $x$, our detector inverts $x$ under every generator $G \in \mathcal{G}$ (\Cref{eq:inversion}), using a caption of $x$ as the prompt, which in our experiments is the image's COCO caption, and computes an \emph{authenticity score} $a_G(x) \in [0,1]$, which is low when the reconstruction $\tilde{x}_G$ is faithful.
The caption is required because inversion integrates \Cref{eq:flow} backwards conditioned on a prompt, which the defender does not know for a suspect sample; in deployment, a captioning model such as BLIP-2~\cite{blip2} would provide it.

If $a_G(x) < \tau_G$ for some generator, the detector abstains and returns $\tilde{x}_G$ and $G$ as evidence (G2), since anyone with $G$ could have created $x$.
If $a_G(x) \geq \tau_G$ for every generator, the detector certifies $x$ as authentic relative to $\mathcal{G}$.
Each certificate names the generator set, thresholds, and calibration date, so certificates against an outdated $\mathcal{G}$ are recognizable.
\Cref{alg:procedure} summarizes calibration and deployment, and \Cref{fig:concept} shows the pipeline.

\begin{figure*}[!t]
    \centering

    \begin{subfigure}[t]{0.23\textwidth}
        \centering
        \begin{minipage}[t][0.18\textheight][c]{\linewidth}
            \centering
            \includegraphics[width=\linewidth,height=0.17\textheight,keepaspectratio]{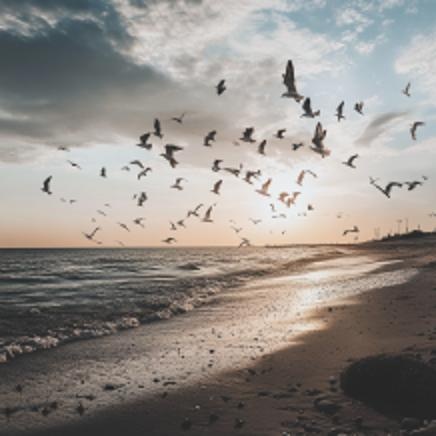}
        \end{minipage}
        \caption*{\footnotesize (a) Original (Easy to invert)}
    \end{subfigure}
    \hfill
    \begin{subfigure}[t]{0.23\textwidth}
        \centering
        \begin{minipage}[t][0.18\textheight][c]{\linewidth}
            \centering
            \includegraphics[width=\linewidth,height=0.17\textheight,keepaspectratio]{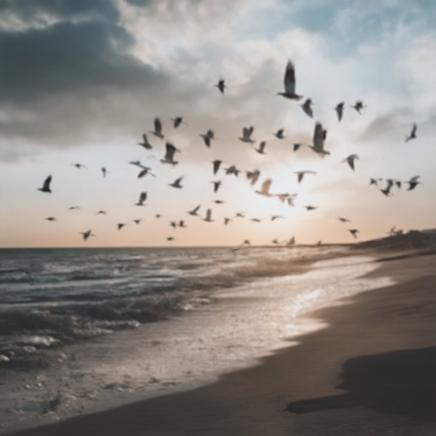}
        \end{minipage}
        \caption*{\footnotesize (b) Inverted (Easy to invert)}
    \end{subfigure}
    \hfill
    \begin{subfigure}[t]{0.23\textwidth}
        \centering
        \begin{minipage}[t][0.18\textheight][c]{\linewidth}
            \centering
            \includegraphics[width=\linewidth,height=0.17\textheight,keepaspectratio]{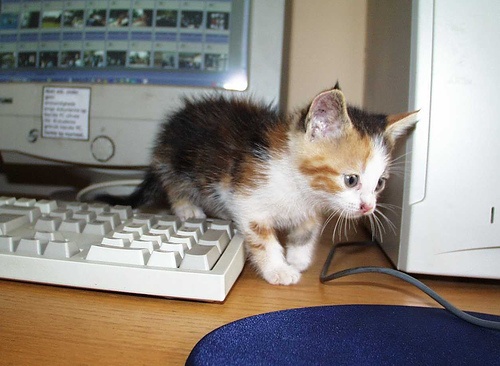}
        \end{minipage}
        \caption*{\footnotesize (c) Original (Easy to invert)}
    \end{subfigure}
    \hfill
    \begin{subfigure}[t]{0.23\textwidth}
        \centering
        \begin{minipage}[t][0.18\textheight][c]{\linewidth}
            \centering
            \includegraphics[width=\linewidth,height=0.17\textheight,keepaspectratio]{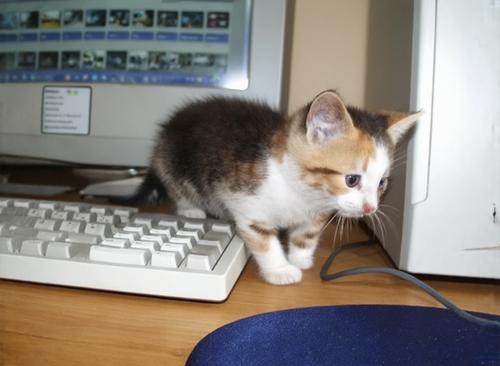}
        \end{minipage}
        \caption*{\footnotesize (d) Inverted (Easy to invert)}
    \end{subfigure}

    \vspace{0.25cm}

    \begin{subfigure}[t]{0.23\textwidth}
        \centering
        \begin{minipage}[t][0.18\textheight][c]{\linewidth}
            \centering
            \includegraphics[width=\linewidth,height=0.17\textheight,keepaspectratio]{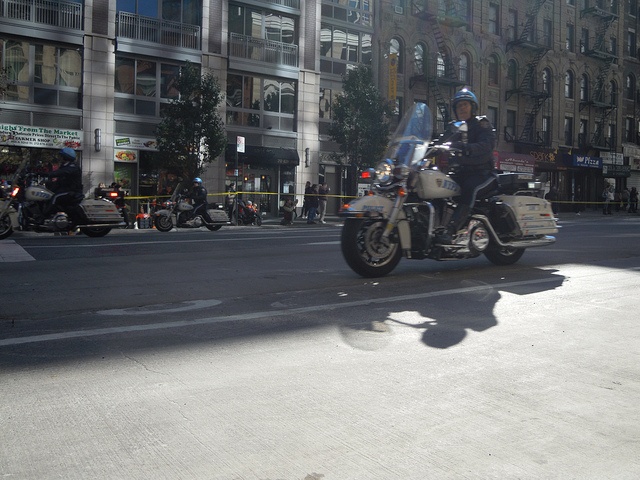}
        \end{minipage}
        \caption*{\footnotesize (e) Original (Hard to invert)}
    \end{subfigure}
    \hfill
    \begin{subfigure}[t]{0.23\textwidth}
        \centering
        \begin{minipage}[t][0.18\textheight][c]{\linewidth}
            \centering
            \includegraphics[width=\linewidth,height=0.17\textheight,keepaspectratio]{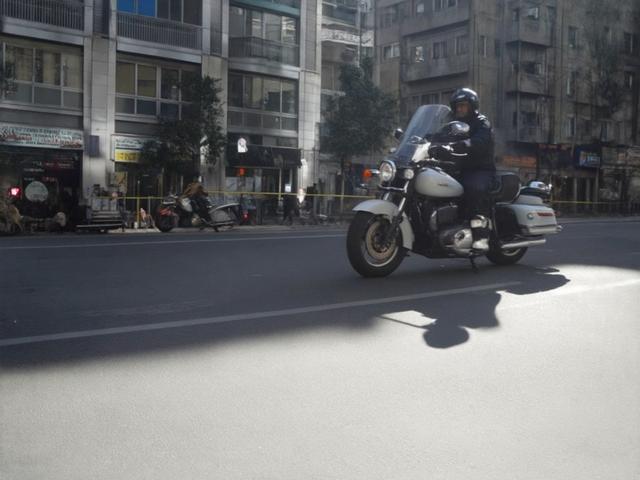}
        \end{minipage}
        \caption*{\footnotesize (f) Inverted (Hard to invert)}
    \end{subfigure}
    \hfill
    \begin{subfigure}[t]{0.23\textwidth}
        \centering
        \begin{minipage}[t][0.18\textheight][c]{\linewidth}
            \centering
            \includegraphics[width=\linewidth,height=0.17\textheight,keepaspectratio]{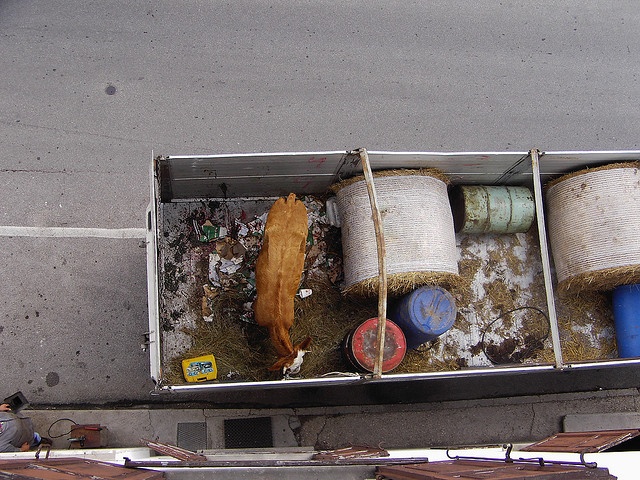}
        \end{minipage}
        \caption*{\footnotesize (g) Original (Hard to invert)}
    \end{subfigure}
    \hfill
    \begin{subfigure}[t]{0.23\textwidth}
        \centering
        \begin{minipage}[t][0.18\textheight][c]{\linewidth}
            \centering
            \includegraphics[width=\linewidth,height=0.17\textheight,keepaspectratio]{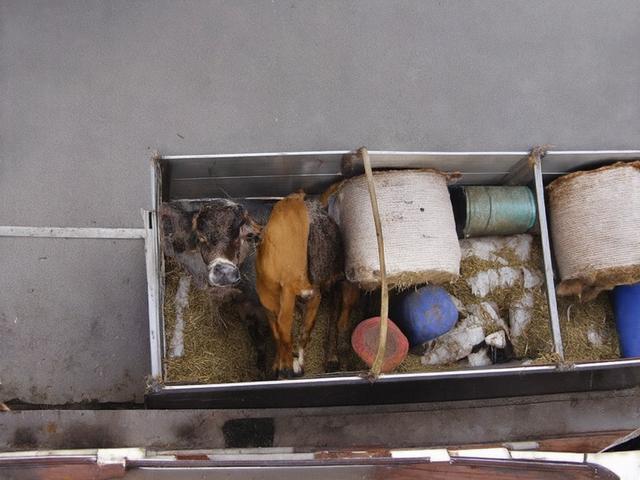}
        \end{minipage}
        \caption*{\footnotesize (h) Inverted (Hard to invert)}
    \end{subfigure}

    \caption{
    \textbf{Examples of easy- and hard-to-invert images.}
    Panels (a), (c), (e), and (g) are real inputs, while panels (b), (d), (f), and (h) are their generated reconstructions.
    The top row shows easy-to-invert images, and the bottom row shows hard-to-invert images.
    Visual complexity generally makes an image harder to invert because fine objects, overlapping structures, and spatial relationships are difficult to reproduce, as shown in (e)--(h).
    However, the visually complex input in (c) is reconstructed faithfully in (d) because its content is well represented within the generator's latent space.
    Inversion difficulty therefore depends on both image complexity and how well the generator's latent space represents the image.
    }
    \label{fig:inversion_examples_main}
\end{figure*}

In \Cref{alg:procedure}, lines 1--4 calibrate, i.e., for each generator we invert held-out samples and set the safety threshold at the $(1{-}\alpha)$-quantile of their scores (line 2), then attack the same samples and set the security threshold likewise (line 3).
The remaining lines deploy, i.e., we score the suspect sample under each generator, abstain with evidence on the first faithful reconstruction (line 7), and certify only when every generator fails (line 10).
\Cref{fig:inversion_examples_main} shows examples of this reconstruction evidence, contrasting easy-to-invert samples with hard-to-invert samples.

\begin{algorithm}[t]
\caption{Sound detection via reconstruction}
\label{alg:procedure}
\begin{algorithmic}[1]
\renewcommand{\algorithmicrequire}{\textbf{Calibration}}
\renewcommand{\algorithmicensure}{\textbf{Deployment}}
\REQUIRE generators $\mathcal{G}$, held-out samples $\mathcal{D}_G$ for each $G$, rate $\alpha$
\FOR{$G \in \mathcal{G}$}
\STATE $\tau_G^{\text{safety}} \leftarrow (1{-}\alpha)$-quantile of $\{a_G(x) : x \in \mathcal{D}_G\}$
\STATE $\tau_G^{\text{security}} \leftarrow (1{-}\alpha)$-quantile of $\{\max_{\|\delta\|_\infty \leq \epsilon} a_G(x{+}\delta) : x \in \mathcal{D}_G\}$
\ENDFOR
\ENSURE suspect $x$, threshold choice $\tau_G \in \{\tau_G^{\text{safety}}, \tau_G^{\text{security}}\}$
\FOR{$G \in \mathcal{G}$}
\IF{$a_G(x) < \tau_G$}
\RETURN \textsc{Abstain} with evidence $(\tilde{x}_G, G)$
\ENDIF
\ENDFOR
\RETURN \textsc{Certify} as authentic relative to $\mathcal{G}$
\end{algorithmic}
\end{algorithm}

\subsection{Authenticity Score}
The score must be measurable on generator outputs (G1), computable for every generator we can invert (G4), and differentiable, so the defender can attack it during calibration (G3).
Two images can differ at four levels, i.e., in pixels, structure, perception, and semantics, and we pick one established metric per level, as no single level suffices (\Cref{app:score} ablates all fifteen metric subsets).
We combine PSNR (pixels), SSIM (structure), $1{-}$LPIPS (perception), and CLIP similarity (semantics),
\begin{equation}
\begin{split}
s(x,\tilde{x}) =\;& \alpha_1 \cdot \text{PSNR}(x,\tilde{x}) + \alpha_2 \cdot \text{SSIM}(x,\tilde{x}) \\
&+ \alpha_3 \cdot (1 - \text{LPIPS}(x,\tilde{x})) + \alpha_4 \cdot \text{CLIP}(x,\tilde{x}),
\end{split}
\label{eq:score}
\end{equation}
and normalize the result with a logistic function of scale $\sigma$,
\begin{equation}
a(x, \tilde{x}) \;=\; \frac{\exp\!\big({-}\sigma \cdot s(x,\tilde{x})\big)}{1 + \exp\!\big({-}\sigma \cdot s(x,\tilde{x})\big)},
\label{eq:aindex}
\end{equation}
so faithful reconstructions receive low scores.
The weights $\alpha_1, \dots, \alpha_4$ are fitted to separate authentic from generated reference images\footnote{We fit the weights on a held-out split $D_{\mathrm{fit}}$ that is disjoint from the calibration and test data (\Cref{sec:experiments}).} (\Cref{fig:methodology_a_calib}), yielding $\alpha_1 = -0.0181$ (PSNR), $\alpha_2 = 1.380$ (SSIM), $\alpha_3 = -4.058$ ($1{-}$LPIPS), and $\alpha_4 = 8.066$ (CLIP).
The fitted weights concentrate on semantics, as CLIP similarity is the strongest single metric with an effective AUC of 0.8546, whereas PSNR receives a near-zero weight.
The fitted weight of $1{-}$LPIPS is negative, as perceptual reconstruction can be disproportionately faithful for generated images, so high perceptual similarity itself indicates generated origin.
Removing CLIP from the combination raises the distribution overlap from 0.3436 to 0.5284, the largest degradation among the four metrics, with formulas and full ablations in \Cref{app:score}.
Our figures abbreviate the authenticity score as A-index.

\subsection{Calibration}
For each $G \in \mathcal{G}$, we draw $n$ held-out samples from $G$, invert them, and set the \emph{safety threshold} $\tau_G^{\text{safety}}$ such that
\begin{equation}
\Pr_{x \sim G}\big[\,a_G(x) \geq \tau_G^{\text{safety}}\,\big] \leq \alpha,
\label{eq:safety}
\end{equation}
i.e., at most a fraction $\alpha$ of $G$'s outputs is certified (we use $\alpha = 1\%$).
Since certification requires every generator's threshold to be met, \Cref{eq:fpr} stays bounded for any mixture of generators in $\mathcal{G}$.
The quantile of $n$ samples estimates \Cref{eq:safety} with sampling error, so we report each threshold with a confidence bound.
The defender controls this sampling error, as they generate the calibration samples themselves and can increase $n$ until the bound is tight enough for their deployment.
No component is trained, which is why calibration is all a new generator costs (G4).
Calibration happens entirely offline, before deployment, as the defender generates all calibration images and knows the transformations the attacker may apply.

\subsection{Calibration under Attack}
\label{sec:adversial_attack}
The adversary of \Cref{sec:threat_model} perturbs generated content to raise its score above the threshold.
Transformed generated content has a different score distribution than clean outputs, so the attacker needs its own threshold. We therefore repeat the calibration on attacked samples, i.e., for each held-out $x \sim G$ we solve
\begin{equation}
\max_{\|\delta\|_\infty \leq \epsilon} \; a_G(x + \delta), \qquad \epsilon = 8/255,
\label{eq:attack}
\end{equation}
with projected gradient ascent through a differentiable approximation of the inversion pipeline, and set the \emph{security threshold} $\tau_G^{\text{security}}$ as the $(1{-}\alpha)$-quantile of the attacked scores.
The maximization includes the unperturbed sample as a feasible point, so the security threshold is stricter, i.e., $\tau_G^{\text{security}} \geq \tau_G^{\text{safety}}$, and it bounds \Cref{eq:fpr} against the calibration attack (G3).
A defender who anticipates a different attack recalibrates with it, using the same procedure.
We calibrate against $\ell_\infty$ perturbations because they are a standard attack that leaves the content intact, i.e., an attacker who wants their fabrication seen cannot change what it shows.
Visible modifications change the content instead, and none of the twenty-four transformations we test, which commonly occur when content is shared, e.g., cropping, text overlays, compression, and noise, pushes generated content past the security threshold (\Cref{app:semantic}).

\subsection{Recalibration and Video}
When a new generator is released, the defender obtains its weights, verifies that inversion reconstructs its outputs, generates and attacks held-out samples, and computes both thresholds.
Recalibration costs one inversion per calibration image and generator, so its price scales linearly with the calibration set size and the number of generators.
Recalibration also gives certificates an expiry date, i.e., since each certificate names the generator set it was issued against, the defender can re-check certified content once $\mathcal{G}$ grows and revoke certificates that a new generator can reproduce.
Our study in \Cref{sec:social-media} measures how many certificates each generator release revokes.
For video, we sample $k$ frames per clip, invert and score each frame with the image pipeline, and average the per-frame scores,
\begin{equation}
a_{\text{video}} \;=\; \frac{1}{k}\sum_{i=1}^{k} a(f_i, \tilde{f}_i),
\label{eq:video}
\end{equation}
which we calibrate with the identical procedure (we use $k = 8$).

\begin{figure}[!t]
    \centering
    \includegraphics[width=\linewidth]{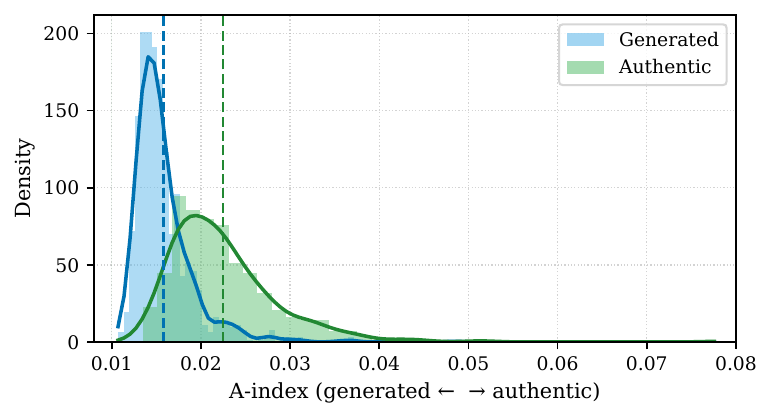}
   \caption{
\textbf{Reconstruction fidelity separates real and fake images.}
Distribution of the A-index \(a(x,\tilde{x})\) obtained by inverting both real and fake images with \texttt{SD3 Medium}.
Fake images are concentrated at lower scores because they are reconstructed more faithfully, whereas real images are shifted toward higher scores.
This separation provides the signal used to calibrate the generator-specific authentication threshold.
}
    \label{fig:methodology_a_calib}
\end{figure}

\section{Experiments}
\label{sec:experiments}

\tit{Generators and Data}
We evaluate five open-weight generator configurations: Stable Diffusion 2.1~\cite{rombach2021ldm}, SD3 Medium~\cite{stability_sd3_medium,esser2024sd3}, SD3.5 Medium~\cite{stability_sd35}, FLUX.1 Dev~\cite{flux2024}, and FLUX.1 Dev with a Realism LoRA adapter~\cite{xlabs_flux_realismlora}.
All five configurations use RF-Inversion~\cite{rf_inversion} for reconstruction, with $28$ inference steps, a guidance scale of $3.5$, float16 precision, and a fixed random seed.
We calibrate on 2{,}000 images from De-Factify 4~\cite{defactify4,malviya2025skdu}, containing 1{,}000 authentic and 1{,}000 generated images at $512\times512$ resolution.
The generated calibration set contains 200 images from each of SD2.1, SDXL, SD3, DALL\textperiodcentered E~3, and Midjourney~6.

\tit{Data Separation}
We fit the A-index weights $\alpha_1,\ldots,\alpha_4$ and the sigmoid scale on $D_{\mathrm{fit}}$, estimate safety thresholds on generated samples in $D_{\mathrm{cal}}$, and compute every reported result on $D_{\mathrm{test}}$.
The three datasets are mutually disjoint, and the security thresholds are estimated on adversarially perturbed versions of the same calibration samples in $D_{\mathrm{cal}}$.
We optimize the weights with Differential Evolution to minimize the overlap between the authentic and generated score distributions, using a population of $20$, a mutation factor of $0.6$, a crossover probability of $0.7$, and at most $300$ iterations.
All AUC, TPR, distribution-overlap, ablation, and attack results are computed on $D_{\mathrm{test}}$, which is used neither for weight fitting nor for threshold calibration.
We set the sigmoid scale to $\sigma=0.9$ based on a sensitivity analysis, and \Cref{app:score} reports the fitted weights and the sensitivity of separability to weight perturbations.

\tit{Baselines and Calibration}
We compare against the twenty post-hoc detectors in \Cref{tab:attack_results} using their original inference code, thresholds, and publicly available weights.
For each generator, we set the safety threshold to the $99$th percentile of its generated calibration scores, corresponding to an FPR of $\alpha=1\%$.
We calibrate the security threshold with the same procedure after attacking the generated calibration samples, as defined in \Cref{eq:safety,eq:attack}.
All timings are measured on one NVIDIA RTX 5000 Ada, on which inverting and scoring one image under one generator takes $11.67$ seconds.
Additional fitting, hardware, and hyperparameter details are provided in the appendix.

\tit{Calibrated Thresholds}
The resulting safety thresholds are generator-specific: $0.015$ for SD2.1, $0.0365$ for SD3 Medium, $0.0365$ for SD3.5 Medium, $0.035$ for FLUX.1 Dev, and $0.038$ for FLUX.1 Dev with the Realism LoRA.
Thresholds rise with generator capability, as the 2022 generator admits a threshold of $0.015$ while every 2024 configuration requires at least $0.035$, reflecting improved reconstruction fidelity.
The Realism LoRA adapter requires the strictest threshold of all five configurations, although it modifies the same FLUX.1 base model.
\Cref{fig:methodology_a_calib} shows the score separation underlying this calibration for SD3 Medium.

\tit{Experimental Questions}
Our experiments address six questions: (i) how much does combining complementary similarity metrics improve separation over any single metric, (ii) how much authentication recall do existing detectors retain on unseen generators at $1\%$ FPR, (iii) can semantic transformations or bounded PGD perturbations cross the calibrated thresholds, (iv) can an attacker combining best-of-$100$ search with PGD obtain a certificate, (v) how many unverified internet images exceed the calibrated thresholds, and (vi) does frame aggregation preserve the reconstruction signal on video?

\tit{Combining Metrics Improves Separation}
\tinytit{Protocol}
We fit weights independently for all fifteen non-empty metric subsets and measure the overlap of the authentic and generated A-index distributions, the effective AUC, and Cohen's $d$.
We also compare the linear combination of \Cref{eq:score} against alternative aggregation strategies, including logistic regression, an MLP, a random forest, and an RBF-kernel SVM.

\tinytit{Results}
Generated images reconstruct more faithfully than authentic images on every metric, e.g., with a mean PSNR of $23.69$~dB versus $22.00$~dB and a mean CLIP similarity of $0.9666$ versus $0.9293$.
The full four-metric combination reduces the distribution overlap to $0.3436$ at an effective AUC of $0.8852$, compared with $0.4401$ for CLIP similarity as the best single metric.
The best three-metric subset, i.e., SSIM, $1{-}$LPIPS, and CLIP, reaches an overlap of $0.3476$ and nearly matches the full combination.
Removing SSIM from the full combination raises the overlap to $0.3936$ and removing $1{-}$LPIPS raises it to $0.4035$, while removing PSNR changes it by only $0.0040$.
The full combination also raises Cohen's $d$ from $1.2996$ for CLIP similarity alone to $1.5870$.
An RBF-kernel SVM reaches a higher effective AUC of $0.9125$, while the linear combination performs within $0.0273$ AUC of it using four parameters.
Multiplicative perturbations of $30\%$ on the fitted weights raise the overlap only from $0.3436$ to $0.3917$, and \Cref{app:score} reports all subsets and aggregation strategies.

\subsection{Results on Generalizability and Robustness to Adversarial Attacks}

\tit{Default Accuracy Hides Asymmetric Errors}
\tinytit{Protocol}
We evaluate all twenty detectors on a balanced test set of 1{,}000 authentic and 1{,}000 generated images from generators excluded from their training data.
The generated test images stem from six such generators: SD2.1, SDXL, SD3 Medium, SD3.5 Medium, DALL\textperiodcentered E~3, and Midjourney~6.
This zero-shot setting measures how each detector behaves when the generation method changes after its training.
We report default-threshold accuracy and class-wise accuracy, then recalibrate every score to measure TPR at the fixed FPR of $1\%$ defined in \Cref{eq:fpr,eq:tpr}.
We report class-wise results because aggregate accuracy can conceal substantially different behavior across the two classes.
This operating point measures how many authentic images remain certifiable when at most $1\%$ of generated images may be incorrectly certified, and it separates the quality of the underlying scores from the choice of the original decision threshold.

\begin{figure}[!t]
    \centering
    \includegraphics[width=0.95\linewidth]{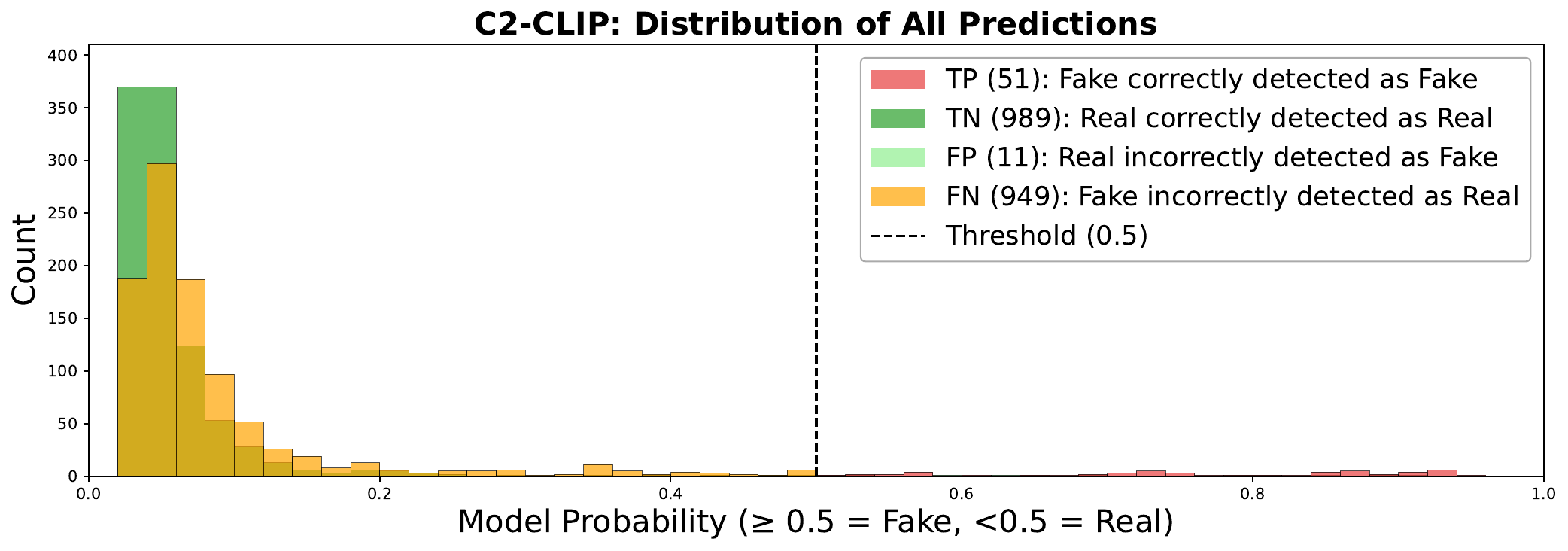}
    \caption{
    \textbf{C2PClip score distributions overlap under zero-shot distribution shift.}
    Predictions on a balanced test set of 1{,}000 real and 1{,}000 fake images from unseen generators.
    C2PClip correctly classifies most real images but assigns most fake images to the real class.
    The resulting overlap prevents the detector from retaining useful recall when its threshold is calibrated to limit false authenticity claims.
    }
    \label{fig:c2_clip}
\end{figure}

\tinytit{Results}
Default-threshold accuracy ranges from $36.95\%$ to $93.25\%$, but seven detectors correctly classify $96.2\%$-$99.5\%$ of authentic images and only $0.1\%$-$5.5\%$ of generated images.
UFD~\cite{ufd} and FatFormer~\cite{fatformer} each classify exactly $1$ of the $1{,}000$ generated images correctly, while FreqNet~\cite{freqnet}, FerretNet~\cite{ferretnet}, AllPatchesMatter~\cite{allpatchesmatter}, IAPL~\cite{iapl}, and ForensicConcept~\cite{forensicconcept} exhibit the same imbalance.
FreqNet, for example, classifies $53$ generated but $995$ authentic images correctly, FerretNet $40$ and $962$, AllPatchesMatter $34$ and $987$, and IAPL $14$ and $980$.
C2PClip~\cite{c2pclip} classifies $51$ of the $1{,}000$ generated images correctly while assigning $949$ to the authentic class, although the test set contains both classes in equal numbers.
\Cref{fig:c2_clip} shows the score-level behavior underlying this asymmetry, as the predictions concentrate on the authentic side of the decision boundary.
The imbalance is not universal, as D3~\cite{d3} classifies $736$ generated and $942$ authentic images correctly, and SICA~\cite{sica} classifies $716$ and $924$.
NPR~\cite{npr} fails differently, as its accuracy decreases on both classes to $36.95\%$ overall instead of retaining high recall on authentic images, and DGS-Net~\cite{dgsnet} classifies only $10$ generated and $847$ authentic images correctly.
The zero-shot prediction distributions of all baselines are reported in \Cref{fig:zero_shot_all}.
At $1\%$ FPR, the overlapping score distributions force most baselines to near-zero TPR, including OmniAID~\cite{omniaid}, which reaches $93.25\%$ accuracy at its default threshold.
Satisfying the $1\%$ FPR constraint requires moving each detector's threshold to an operating point at which almost no authentic image remains certifiable.
In contrast, the authentic and generated A-index distributions remain separated in the same zero-shot setting, so our detector retains recall on authentic images after calibration to the same $1\%$ FPR.
Because the same FPR constraint is applied to every method, this difference follows from the separation of the underlying scores rather than from the calibration procedure.
These results show that default accuracy does not measure authentication recall at a deployment threshold.

\tit{Bounded Attacks Break Binary Detectors}
\tinytit{Protocol}
We apply PGD~\cite{pgd} with an $\ell_\infty$ budget of $\epsilon=8/255$ to the samples that each detector classifies correctly before the attack.
The adaptive adversary knows the detection procedure, the evaluated generator set $\mathcal{G}$, and the calibrated thresholds.
Against our detector, the attack succeeds only if the perturbed image satisfies the calibrated threshold for every generator in $\mathcal{G}$, whereas flipping a binary detector requires crossing a single decision boundary.
If $a_G(x+\delta) < \tau_G$ holds for any generator, the detector still abstains and returns the corresponding reconstruction as evidence.
We report before-attack accuracy, post-attack accuracy, and the attack success rate as the fraction of correctly classified samples that the attack flips.
We also evaluate the A-index under twenty-four semantic transformations covering photometric changes, spatial transformations, text and box overlays, resampling, JPEG compression, blur, and Gaussian noise.
\Cref{tab:attack_results} reports the PGD results, while \Cref{tab:semantic_attacks} reports the semantic transformations.

\begin{table}[!t]
\caption{Results before/after PGD ($\epsilon=\frac{8}{255}$) on \textbf{2,000 Images} (1{,}000 fake + 1{,}000 real, $512{\times}512$). Following standard practice, only samples classified correctly \emph{before} the attack are perturbed, so accuracy can only drop. Attack success is the fraction of those correct samples flipped. $\dagger$: threshold-degenerate detector (score has no natural $0.5$ cut). ``$-$'': no sample of that class was correct pre-attack.}
\label{tab:attack_results}
\setlength{\tabcolsep}{2pt}
    \centering
        \resizebox{\linewidth}{!}{%
        \begin{tabular}{l ccc ccc cc}
        \toprule
        & \multicolumn{3}{c}{\textbf{Before attack}}
        & \multicolumn{3}{c}{\textbf{After attack}}
        & \multicolumn{2}{c}{\textbf{Attack success (\%)}} \\
        \cmidrule(lr){2-4}\cmidrule(lr){5-7}\cmidrule(lr){8-9}
        \textbf{Model}
        & \shortstack{Correct\\fake} & \shortstack{Correct\\real} & \shortstack{Acc.\\(\%)}
        & \shortstack{Correct\\fake} & \shortstack{Correct\\real} & \shortstack{Acc.\\(\%)}
        & \shortstack{on\\fake} & \shortstack{on\\real} \\
        \midrule
        UFD~\cite{ufd} & 1 & 974 & 48.75 & 0 & 0 & \cellcolor{red!12}\textbf{0.00} & 100.0 & 100.0 \\
        FreqNet~\cite{freqnet} & 53 & 995 & 52.40 & 0 & 0 & \cellcolor{red!12}\textbf{0.00} & 100.0 & 100.0 \\
        NPR~\cite{npr} & 86 & 653 & 36.95 & 0 & 0 & \cellcolor{red!12}\textbf{0.00} & 100.0 & 100.0 \\
        FatFormer~\cite{fatformer} & 1 & 987 & 49.40 & 0 & 0 & \cellcolor{red!12}\textbf{0.00} & 100.0 & 100.0 \\
        AEROBLADE$^\dagger$~\cite{ricker2024aeroblade} & 778 & 778 & 77.80 & 0 & 30 & \cellcolor{yellow!20}1.50 & 100.0 & 96.1 \\
        C2PClip~\cite{c2pclip} & 51 & 949 & 50.00 & 0 & 2 & \cellcolor{red!12}\textbf{0.10} & 100.0 & 99.8 \\
        D3~\cite{d3} & 736 & 942 & \cellcolor{green!15}83.90 & 24 & 11 & \cellcolor{yellow!20}1.75 & 96.7 & 98.8 \\
        FIRE$^\dagger$~\cite{chu2025fire} & 998 & 0 & 49.90 & 1 & 0 & \cellcolor{red!12}\textbf{0.05} & 99.9 & - \\
        DDA~\cite{dda} & 576 & 975 & 77.55 & 0 & 4 & \cellcolor{red!12}\textbf{0.20} & 100.0 & 99.6 \\
        FerretNet~\cite{ferretnet} & 40 & 962 & 50.10 & 0 & 0 & \cellcolor{red!12}\textbf{0.00} & 100.0 & 100.0 \\
        WaRPAD$^\dagger$~\cite{warpad} & 689 & 689 & 68.90 & 15 & 0 & \cellcolor{yellow!20}0.75 & 97.8 & 100.0 \\
        APM~\cite{allpatchesmatter} & 34 & 987 & 51.05 & 0 & 0 & \cellcolor{red!12}\textbf{0.00} & 100.0 & 100.0 \\
        OmniAID~\cite{omniaid} & 917 & 948 & \cellcolor{green!15}93.25 & 0 & 0 & \cellcolor{red!12}\textbf{0.00} & 100.0 & 100.0 \\
        PGC~\cite{pgc} & 304 & 951 & 62.75 & 0 & 0 & \cellcolor{red!12}\textbf{0.00} & 100.0 & 100.0 \\
        PROBE~\cite{probe} & 539 & 990 & 76.45 & 0 & 1 & \cellcolor{red!12}\textbf{0.05} & 100.0 & 99.9 \\
        DEAR~\cite{dear} & 417 & 993 & 70.50 & 0 & 0 & \cellcolor{red!12}\textbf{0.00} & 100.0 & 100.0 \\
        DGS-Net~\cite{dgsnet} & 10 & 847 & 42.85 & 0 & 0 & \cellcolor{red!12}\textbf{0.00} & 100.0 & 100.0 \\
        SICA~\cite{sica} & 716 & 924 & \cellcolor{green!15}82.00 & 0 & 0 & \cellcolor{red!12}\textbf{0.00} & 100.0 & 100.0 \\
        IAPL~\cite{iapl} & 14 & 980 & 49.70 & 0 & 0 & \cellcolor{red!12}\textbf{0.00} & 100.0 & 100.0 \\
        ForensicCpt~\cite{forensicconcept} & 55 & 982 & 51.85 & 0 & 2 & \cellcolor{red!12}\textbf{0.10} & 100.0 & 99.8 \\
        \bottomrule
        \end{tabular}
        }%
\end{table}

\begin{figure*}[!t]
    \centering
    \begin{minipage}[b]{0.185\textwidth}
        \begin{subfigure}[b]{\linewidth}
            \includegraphics[width=\linewidth]{diagrams/a_brown_and_white_dog_is_sitting_in_some_grass_and_a_red_and_white_fire_hydrant_original.png}
            \caption{Original}
        \end{subfigure}\\[2pt]
        \begin{subfigure}[b]{\linewidth}
            \includegraphics[width=\linewidth]{diagrams/a_brown_and_white_dog_is_sitting_in_some_grass_and_a_red_and_white_fire_hydrant_adversarial.png}
            \caption{PGD-perturbed}
        \end{subfigure}\\[2pt]
        \begin{subfigure}[b]{\linewidth}
            \includegraphics[width=\linewidth]{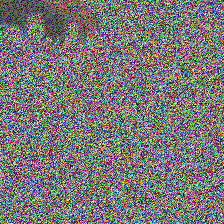}
            \caption{Perturbation $\delta$ ($16\times$)}
        \end{subfigure}
    \end{minipage}\hfill
    \begin{minipage}[b]{0.79\textwidth}
        \begin{subfigure}[b]{\linewidth}
            \includegraphics[width=\linewidth]{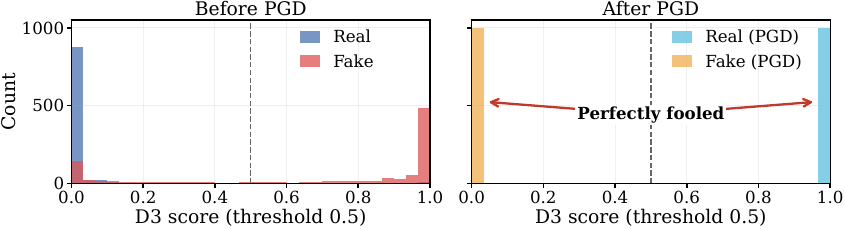}
            \caption{D3 predictions before and after PGD}
        \end{subfigure}\\[2pt]
        \begin{subfigure}[b]{\linewidth}
            \includegraphics[width=\linewidth]{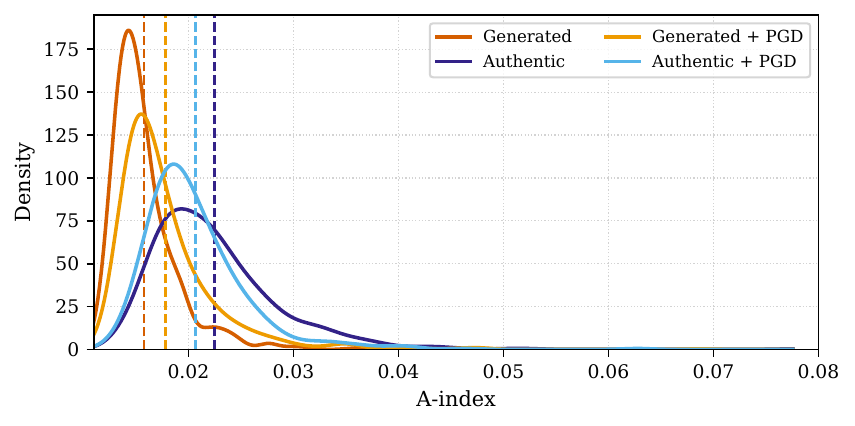}
            \caption{A-index distributions before and after PGD}
        \end{subfigure}
    \end{minipage}
    \caption{
    \textbf{PGD perturbations collapse binary detectors but preserve separated A-index distributions.}
    Panels (a)--(c) show an image before and after an $\ell_\infty$-bounded perturbation with $\epsilon=8/255$, and the perturbation $\delta$ amplified $16\times$ for visibility.
    Panel (d) shows that D3 declines from $83.90\%$ to $1.75\%$ accuracy, the highest post-attack accuracy among the twenty baselines.
    Panel (e) shows that PGD shifts both A-index distributions, while the distributions retain distinct peaks and motivate the security threshold.
    }
    \label{fig:d3_pgd}
\end{figure*}

\tinytit{Results}
\Cref{fig:d3_pgd} illustrates the visual perturbation and the corresponding score-distribution changes under PGD.
Twelve of the twenty baselines fall to $0.00\%$ accuracy after PGD, and none retains more than $1.75\%$ accuracy.
UFD~\cite{ufd} degrades from $48.75\%$ to $0.00\%$ accuracy, misclassifying every image after the attack, with a $100.0\%$ attack success rate on both classes.
FreqNet~\cite{freqnet}, NPR~\cite{npr}, and FatFormer~\cite{fatformer} show the same outcome, falling from $52.40\%$, $36.95\%$, and $49.40\%$ accuracy, respectively, to $0.00\%$.
The collapse includes the detectors with the highest pre-attack accuracy, as OmniAID~\cite{omniaid} falls from $93.25\%$ to $0.00\%$ and SICA~\cite{sica} falls from $82.00\%$ to $0.00\%$.
DEAR~\cite{dear}, PGC~\cite{pgc}, and DGS-Net~\cite{dgsnet} likewise fall from $70.50\%$, $62.75\%$, and $42.85\%$ accuracy, respectively, to $0.00\%$.
The baselines above $0.00\%$ remain close to it, as PROBE~\cite{probe} falls from $76.45\%$ to $0.05\%$, DDA~\cite{dda} from $77.55\%$ to $0.20\%$, and AEROBLADE~\cite{ricker2024aeroblade} from $77.80\%$ to $1.50\%$.
C2PClip~\cite{c2pclip} and ForensicConcept~\cite{forensicconcept} both fall to $0.10\%$ accuracy, and WaRPAD~\cite{warpad} drops from $68.90\%$ to $0.75\%$ with a $97.8\%$ attack success rate on generated images.
The attack success rate on authentic images is undefined for FIRE~\cite{chu2025fire}, as it classifies no authentic image correctly before the attack.
The post-attack prediction distributions of all baselines are shown in \Cref{fig:pgd_attacks}.
D3~\cite{d3} provides the highest post-attack accuracy but declines from $83.90\%$ to $1.75\%$, retaining $24$ generated and $11$ authentic correct classifications, with attack success rates of $96.7\%$ and $98.8\%$ on the two classes.
For SD3 Medium, attacked calibration raises the security threshold from the clean value of $0.0365$ to $0.038$, while the attacked A-index distributions retain distinct peaks.
The A-index is similarly robust to transformations, as twenty-three of the twenty-four we test, including crops, text overlays, and JPEG compression, keep generated images below $\tau_{\text{safety}}$ (\Cref{tab:semantic_attacks} in \Cref{app:semantic}).
Spatial transformations produce the lowest scores, as a horizontal flip reaches an A-index of $0.0045$, a $20\%$ crop reaches $0.0067$, and a $15^\circ$ rotation reaches $0.0074$, since spatial misalignment degrades SSIM and LPIPS simultaneously.
Across the four spatial transformations, including a $10\%$ crop, SSIM falls to $0.607$-$0.671$ while CLIP similarity remains at $0.916$-$0.963$, which moves the A-index away from the threshold rather than toward it.
Photometric transformations, i.e., brightness, contrast, color shift, and saturation, remain in the range $0.0115$ to $0.0166$, less than half of the safety threshold.
JPEG compression at quality $10$ produces the highest score below the threshold, $0.0272$, which remains $25.5\%$ under the safety threshold.
Gaussian noise ($\sigma=25$) reaches $0.0366$, marginally above the safety threshold but below the security threshold of $0.038$, and reduces SSIM to $0.33$ at this severity.

\subsection{Medium-Resource Attacker}
\label{sec:medium-attacker}

\tit{Best-of-$100$ Search Remains Below Both Thresholds}
\tinytit{Protocol}
An attacker with unbounded resources could synthesize and invert a very large number of candidates, select the candidate with the highest A-index, and adversarially refine it until it exceeds any fixed threshold.
We bound this search: the attacker generates $N=100$ images from one prompt using SD3 Medium, inverts every candidate, and selects the image with the highest A-index.
The attacker then applies projected gradient ascent with $\epsilon=8/255$ to increase the selected candidate's score.
At each iteration, the attacker differentiates the score through the approximated inversion pipeline, takes an ascent step on $\delta$, projects back onto the admissible set, and clips to the valid pixel range.
At the measured processing time of $11.67$ seconds per image and generator, scoring the candidate set requires approximately $1{,}167$ seconds, i.e., $19.5$ minutes when executed sequentially, excluding generation and PGD.
The prompt and six of the sampled candidates are shown in \Cref{sec:medium-attacker-full-details}.

\begin{figure}[!t]
    \centering
        \vspace{0pt}
        \includegraphics[width=\linewidth]{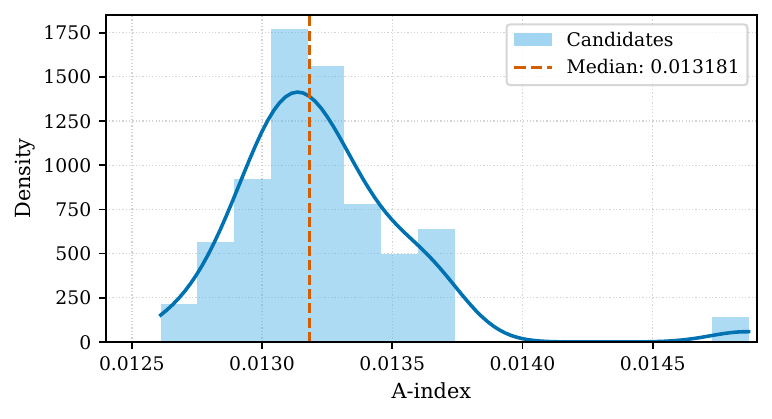}
        \caption{
    \textbf{Best-of-$100$ search does not reach the calibrated thresholds.}
    The attacker samples $100$ candidates from one prompt and selects the highest A-index before PGD refinement.
    The selected candidate increases from $0.0148$ to $0.0154$ after PGD, remaining below the safety and security thresholds.
    }
        \label{fig:medium_attacker}
\end{figure}
\begin{figure}[!b]
\centering
        \vspace{0pt}
        \includegraphics[width=.8\linewidth]{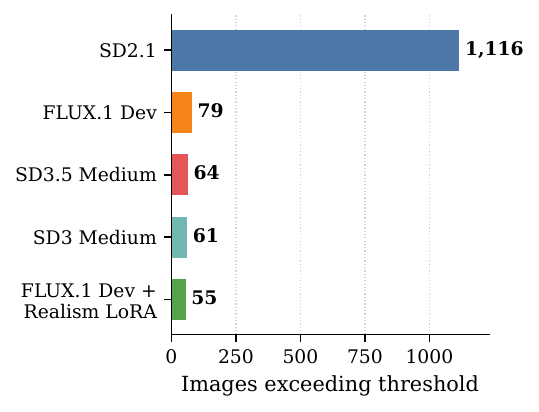}
        \caption{
    \textbf{Calibrated thresholds retain different subsets of $3{,}000$ unverified internet images.}
    We apply five generator-specific thresholds to the same SD3 inversion-score distribution: $1{,}116$ images exceed the SD2.1 threshold, compared with $55$-$79$ for the four newer configurations.
    This comparison measures threshold sensitivity on one score distribution and does not compare generator-specific reconstructions of the corpus.
    Examples from the corpus are shown in \Cref{fig:reddit_examples}.
    }
    \label{fig:scrapped_image}
\end{figure}

\tinytit{Results}
\Cref{fig:medium_attacker} shows the score distribution over the $100$ generated candidates.
The highest-scoring candidate reaches an A-index of $0.0148$, i.e., $40.5\%$ of the safety threshold, and PGD increases its score by $0.0006$ to $0.0154$.
The optimized score remains below the SD3 Medium safety threshold of $0.0365$ and security threshold of $0.038$, causing the detector to abstain.
The attack therefore closes less than $3\%$ of the candidate's remaining gap to the safety threshold.
This result applies to one prompt, $100$ candidates, and the evaluated perturbation budget; it does not establish robustness against larger searches or different optimization procedures.

\subsection{Social-Media Study and the Effect of Adapters}
\label{sec:social-media}

\tit{Generator-Specific Thresholds Retain Different Image Subsets}
\tinytit{Protocol}
We collect $3{,}000$ public Reddit images using the keyword groups described in \Cref{app:reddit} and caption each image with BLIP-2~\cite{blip2}.
The corpus spans nine keyword groups, e.g., \myverbatim{ai_art}, \myverbatim{people_faces}, and \myverbatim{misinformation_graphics}, and contains photographs, screenshots, memes, and digital artwork.
We invert every image with SD3 Medium and compute one SD3-based A-index distribution.
We then apply five safety thresholds that were separately calibrated on outputs from SD2.1, SD3 Medium, SD3.5 Medium, FLUX.1 Dev, and FLUX.1 Dev with a Realism LoRA adapter.
For each generator $G$, we count the images that satisfy $a_{\mathrm{SD3}}(x) \geq \tau_G^{\mathrm{safety}}$, i.e., that remain above the calibrated boundary.

\tinytit{Results}
\Cref{fig:scrapped_image} reports the retained counts, and \Cref{fig:reddit_examples} in the appendix shows examples from the sampled corpus.
The SD2.1 threshold retains $1{,}116$ images ($37.2\%$), while the thresholds for the four newer configurations retain between $55$ and $79$ images ($1.83\%$-$2.63\%$).
The per-generator counts are $79$ for FLUX.1 Dev, $64$ for SD3.5 Medium, $61$ for SD3 Medium, and $55$ for FLUX.1 Dev with the Realism LoRA.
Adding the Realism LoRA changes the FLUX.1 threshold from $0.035$ to $0.038$ and reduces the retained subset from $79$ to $55$ images on the same SD3 score distribution.
A lightweight adaptation of the same base generator therefore changes the calibrated boundary and the set of images that remains above it.
Because all five thresholds are applied to one SD3 inversion-score distribution, these counts measure sensitivity to the calibrated boundary rather than differences among generator-specific inversion pipelines.
The collection is keyword-driven, so the counts apply to the sampled distribution rather than to internet imagery in general (\Cref{app:reddit}).

\begin{figure*}[!t]
    \centering
    \includegraphics[width=0.9\textwidth]{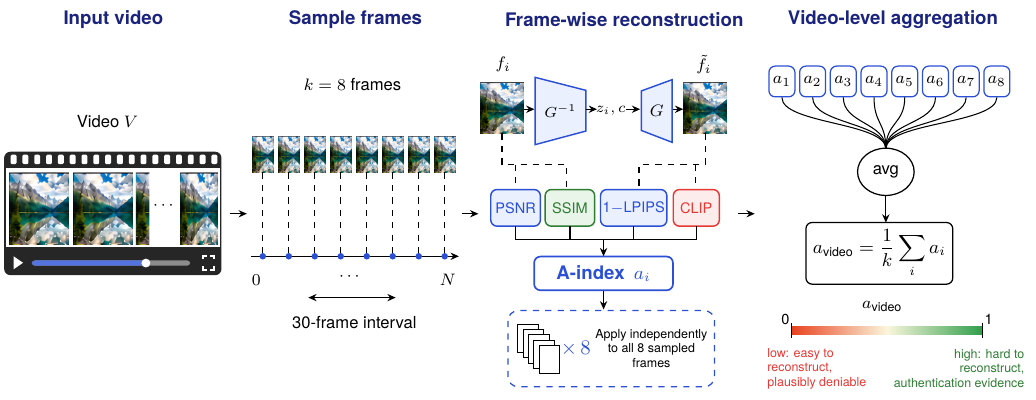}
    \caption{
    \textbf{Frame-aggregated video authentication.}
    We sample eight frames at $30$-frame intervals, reconstruct and score each frame independently, and average their A-index values to obtain $a_{\mathrm{video}}$.
    Lower values indicate that the sampled frames are easier to reconstruct, while higher values provide stronger authentication evidence.
    }
    \label{fig:video_pipeline}
\end{figure*}

\subsection{Video Modality}
\label{sec:video}

\tit{Frame Aggregation Preserves a Reconstruction Signal}
\tinytit{Protocol}
We evaluate a balanced subset of $100$ videos from Deepfake-Eval-2024~\cite{chandra2025deepfakeeval2024multimodalinthewildbenchmark}, containing $50$ authentic and $50$ generated videos.
The benchmark contains recent in-the-wild videos from social-media platforms, on which its authors report that open-source video detectors lose approximately $50\%$ AUC relative to earlier academic datasets.
For each video, we sample eight frames separated by $30$ frames, apply the image pipeline independently, and average their A-index values as defined in \Cref{eq:video}.
The evaluation covers only the visual component of each video and does not use the audio channel.
We compare against GenConViT~\cite{genconvit}, FTCN~\cite{ftcn}, and StyleFlow~\cite{styleflow} using their original code and weights without fine-tuning.
FTCN and StyleFlow explicitly model temporal information (\Cref{sec:background}), which our frame-averaging procedure does not use.
The processing steps and video-level aggregation are summarized in \Cref{fig:video_pipeline}.

\begin{table}[!t]
\centering
\caption{
\textbf{Video detectors transfer poorly to Deepfake-Eval-2024.}
Results on $50$ authentic and $50$ generated videos; no baseline exceeds $0.615$ AUC or $0.59$ precision.
}
\label{tab:video_baseline_failures}
\setlength{\tabcolsep}{3pt}
\renewcommand{\arraystretch}{0.92}
\footnotesize
\begin{tabular*}{0.72\linewidth}{@{\extracolsep{\fill}}lcccc}
\toprule
\textbf{Model} & \textbf{AUC} & \textbf{Prec.} & \textbf{Rec.} & \textbf{F1} \\
\midrule
GenConViT~\cite{genconvit} & 0.615 & 0.59 & 0.49 & 0.53 \\
FTCN~\cite{ftcn}           & 0.483 & 0.50 & 0.64 & 0.40 \\
StyleFlow~\cite{styleflow} & 0.509 & 0.53 & 0.42 & 0.47 \\
\bottomrule
\end{tabular*}
\end{table}

\begin{figure}[!t]
    \centering
    \includegraphics[width=\linewidth]{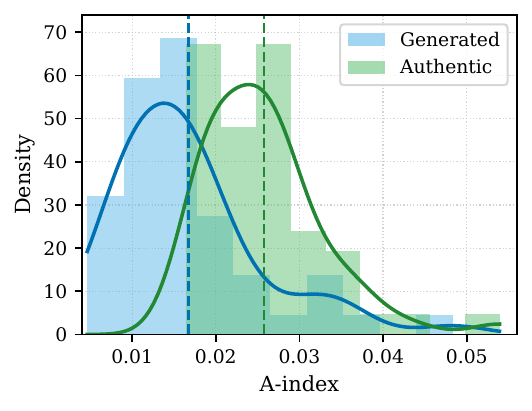}
    \caption{
    \textbf{Frame-aggregated A-index scores retain the expected ordering on video.}
    Each score averages eight independently processed frames from one of $100$ Deepfake-Eval-2024 videos.
    Authentic videos generally receive higher scores and generated videos generally receive lower scores, although the distributions overlap.
    }
    \label{fig:video_graph}
\end{figure}

\tinytit{Results}
\Cref{tab:video_baseline_failures} reports the baseline metrics, while \Cref{fig:video_graph} reports the frame-aggregated A-index distributions.
GenConViT obtains the highest baseline AUC of $0.615$, while FTCN and StyleFlow obtain AUCs of $0.483$ and $0.509$, respectively.
FTCN obtains the highest recall of $0.64$, but its AUC falls below $0.5$, so the higher recall does not correspond to reliable separation between the two classes.
StyleFlow reaches $0.53$ precision, $0.42$ recall, and an F1 score of $0.47$, so no baseline exceeds $0.59$ precision under this distribution shift.
The frame-aggregated A-index retains the image-level ordering, as authentic videos concentrate at higher scores and generated videos at lower scores, but the two distributions overlap and no video-specific threshold is calibrated.
Sequential processing requires approximately $8\times11.67=93.36$ seconds per video and generator on one NVIDIA RTX 5000 Ada, and this cost grows linearly with the number of sampled frames and evaluated generators.
This preliminary result measures frame-level visual evidence only and does not evaluate temporal consistency, motion, audio, or TPR at a calibrated video-specific FPR.
A direct comparison against the video baselines requires calibrating a video-specific threshold on generated calibration videos, which we leave to future work.
Additional dataset, baseline, runtime, and modality details are provided in \Cref{app:video_details}.

\section{Discussion}

\tit{Meaning and Deployment}
A certificate states that every tested generator assigned the sample an A-index above its calibrated threshold on the stated date.
It is not proof of origin because unknown or withheld generators remain outside $\mathcal{G}$, but anyone with the named generators can recompute the result.
When $\mathcal{G}$ changes, the defender recalibrates and re-checks prior certificates, revoking those that a new generator can reproduce.
Public weights therefore help the defender calibrate stronger generators, while private generators remain an unavoidable coverage gap.

\tit{Why Reconstruction Works}
Generated samples lie within a distribution their generator can synthesize, whereas authentic images often contain fine text, low contrast, or complex structure that current inversion priors fail to reproduce.
CLIP similarity alone separates the distributions with an effective AUC of 0.8546, while PSNR receives a near-zero weight, indicating that the strongest signal is semantic rather than pixel-level (\Cref{app:score}).
As generators cover more of the authentic distribution, however, fewer images remain certifiable, consistent with our social-media study (\Cref{sec:social-media}).

\tit{Operating Point and Fundamental Limits}
Detectors should report TPR at a fixed FPR (\Cref{eq:fpr,eq:tpr}) because accuracy hides the deployment threshold: our strongest baseline reaches 93\% accuracy yet certifies almost nothing at 1\% FPR.
This evaluation follows adjacent security fields such as membership inference~\cite{carlini2022membership}.
Our method is consistent with detection impossibility results~\cite{saberi2024robustness}: it does not recover provenance, but bounds false certificates by abstaining.
We restrict this bound to efficient adversaries because an unbounded attacker can search until a generated candidate crosses any fixed threshold (\Cref{sec:medium-attacker}).

\tit{Limitations}
Certification takes 11.67 seconds per image and generator on one NVIDIA RTX 5000 Ada, which suits forensic verification rather than feed-scale screening.

The security threshold covers only the attacks it was calibrated with, and a stronger optimizer or a different attack may exceed it.
Covering a new attack requires no retraining: the defender applies it to the generated calibration samples, re-scores them, and recomputes the $(1-\alpha)$-quantile.
The difficulty lies in coverage.
Each transformation family and parameter shifts the score distribution of generated content and therefore needs its own calibration samples, each costing one inversion, and compositions of transformations grow combinatorially.
Open-ended edits, such as merging images, splicing regions, taking screenshots, or image-to-image editing, cannot be enumerated at all and may introduce structures that the tested generators cannot reproduce, raising generated content above the threshold.
Our image-level score also cannot localize generated regions in such composites.
Certificates also exclude generators that cannot be inverted, including private fine-tunes, and certify less content as generator coverage improves.
Our Reddit results apply to keyword-driven samples from nine content groups rather than internet imagery at large (\Cref{app:reddit}).
Finally, the video study averages eight frames sampled 30 frames apart, models no temporal structure, and contains only 100 videos (\Cref{app:video_details}).

\tit{Outlook}
Future detectors should report TPR at a fixed FPR, evaluate adaptive attackers, and track how the certifiable share of public content changes as generators improve.

\section{Conclusion}

We introduce sound deepfake detection, which certifies content as \emph{authentic} only when no evaluated generator faithfully reconstructs it and otherwise abstains with the reconstruction as checkable evidence.
Our method calibrates its error rate before deployment using a safety threshold for clean content and a stricter security threshold for adversarially perturbed content, both at \(1\%\) FPR.
Across four years of generators, the best of twenty baselines declines from \(99.5\%\) to \(76\%\) accuracy, all fall below \(2\%\) accuracy under adversarial perturbations, and their recall approaches zero at \(1\%\) FPR, while our calibrated thresholds withstand the same attacks.
A best-of-100 attacker refined with PGD raises its score only from 0.0148 to 0.0154, below both thresholds, and none of twenty-four semantic transformations crosses the security threshold.
The procedure extends to video by averaging sampled-frame scores and incorporates each new generator or adapter through one recalibration rather than retraining.
Certificates remain relative to the evaluated generator set, and coverage erodes as generators improve: 1{,}116 of 3{,}000 Reddit images resist a 2022 generator, but only 55 to 79 resist 2024 generators.
These results challenge accuracy as a reliability measure and identify post-hoc verifiability as a shrinking resource that recalibration can track.
We release our code to add future generators through recalibration and measure how much content remains verifiable.

\FloatBarrier
\clearpage
\appendices
\crefalias{section}{appendix}

\section*{Ethical Considerations}

This work studies post-hoc authentication of synthetic media and therefore has dual-use implications. A reliable verifier helps investigators, platforms, and journalists assess whether an image can be reproduced by a known generator and avoid unsupported accusations. However, knowledge of the scoring function, calibration procedure, and decision thresholds may also help an adversary design content that is harder to authenticate. We limit this risk by evaluating adaptive attackers with full knowledge of our method and by calibrating a separate security threshold against the attacks we consider. Publishing the procedure also benefits defenders, who can recalibrate this threshold whenever a stronger attack or a new generator appears.

As discussed above, a certificate remains relative to the evaluated generator set, decision thresholds, and calibration date, and is not proof that a camera captured the image. Because authentication errors have unequal consequences, the output should not be the sole basis for legal, disciplinary, journalistic, or content-moderation decisions and warrants human review.

Our social-media study measures aggregate score distributions over 3{,}000 publicly accessible Reddit images, and public availability does not remove privacy or consent concerns. The analysis does not attribute images to individual users. We do not redistribute the images or any usernames, post text, or other identifying metadata, and we bound the sample's representativeness in \Cref{app:reddit}.

\section*{Open Science}

We publicly release our code at \url{https://github.com/Sarim-MBZUAI/content-authentication}. It contains the RF-Inversion reconstruction for SD2.1, SD3, and SD3.5, the A-index scoring, adapters for all twenty baseline detectors, the PGD attack for each detector, and the scripts that normalize images, score detectors, and aggregate the results.
Hyperparameters and random seeds are set as defaults in the code.
Model weights and datasets are not redistributed and are available from their original sources under their respective licenses.
We do not redistribute the Reddit images to protect user privacy.

\section*{LLM usage considerations}

We used LLMs to revise manuscript prose and assist with LaTeX, plotting code, and schematic figures.
The authors remain responsible for the accuracy and originality of all text, code, figures, results, and references.
Our inversion pipeline also uses BLIP-2~\cite{blip2} to generate captions that condition reconstruction, automating prompt construction for images and sampled video frames.
Caption errors can affect reconstruction fidelity and therefore the authentication score.
LLMs were used for editorial purposes in this manuscript, and all outputs were inspected by the authors to ensure accuracy and originality.

\raggedbottom
\bibliographystyle{IEEEtran}
\bibliography{main}
\newpage

\section{Evaluation Metric}
\label{app:score}
\subsection{Similarity Metrics Formulation}
In this section, we provide the mathematical formulations for all similarity metrics used in computing the Authenticity Index.
\tit{Peak Signal-to-Noise Ratio (PSNR)}
PSNR measures the pixel-level fidelity between the original image $\mathbf{x}$ and its reconstruction $\tilde{\mathbf{x}}$:

\begin{equation}
\text{PSNR}(\mathbf{x}, \tilde{\mathbf{x}}) = 10 \cdot \log_{10}\left(\frac{\text{MAX}_I^2}{\text{MSE}(\mathbf{x}, \tilde{\mathbf{x}})}\right)
\end{equation}
where $\text{MSE}(\mathbf{x}, \tilde{\mathbf{x}}) = \frac{1}{N} \sum_{i=1}^{N} (x_i - \tilde{x}_i)^2$, $\text{MAX}_I$ is the maximum possible pixel value (255 for 8-bit images), and $N$ is the total number of pixels. PSNR is measured in decibels (dB), with higher values indicating better reconstruction quality.
\tit{Structural Similarity Index (SSIM)} SSIM~\cite{ssim} evaluates structural information preservation between images:
\begin{equation}
\text{SSIM}(I, \hat{I}) = \frac{(2\mu_I \mu_{\hat{I}} + C_1)(2\sigma_{I\hat{I}} + C_2)}{(\mu_I^2 + \mu_{\hat{I}}^2 + C_1)(\sigma_I^2 + \sigma_{\hat{I}}^2 + C_2)},
\end{equation}
where $\mu$, $\sigma^2$, and $\sigma_{I\hat{I}}$ are the local means, variances, and cross-covariance of $I$ and $\hat{I}$, and $C_1$, $C_2$ are stabilizing constants.
\tit{Learned Perceptual Image Patch Similarity (LPIPS)} LPIPS~\cite{lpips} computes perceptual distance using deep features extracted from a pretrained network:
\begin{equation}
\text{LPIPS}(\mathbf{x}, \tilde{\mathbf{x}}) = \sum_{\ell} \frac{1}{H_\ell W_\ell} \sum_{h,w} \|\mathbf{w}_\ell \odot (\phi_\ell(\mathbf{x})_{h,w} - \phi_\ell(\tilde{\mathbf{x}})_{h,w})\|_2^2
\end{equation}
where $\phi_\ell$ represents features from layer $\ell$ of a pretrained network, $\mathbf{w}_\ell$ are learned weights, and $H_\ell, W_\ell$ are spatial dimensions at layer $\ell$. Since LPIPS measures distance, we use $(1 - \text{LPIPS})$ in our formulation to align with other similarity metrics.
\tit{CLIP Similarity}
CLIP \cite{radford2021learning} measures semantic consistency in the joint vision-language embedding space:

\begin{equation}
\text{CLIP}(\mathbf{x}, \tilde{\mathbf{x}}) = \frac{E(\mathbf{x}) \cdot E(\tilde{\mathbf{x}})}{\|E(\mathbf{x})\| \cdot \|E(\tilde{\mathbf{x}})\|}
\end{equation}
where $E(\cdot)$ is the CLIP vision encoder, and the similarity is computed as the cosine similarity between the normalized embeddings.
\subsection{Authenticity Index Calibration}
\tit{Weight Optimization for the composite similarity score}
Given our Similarity score:
\begin{equation}
\begin{split}
s(x,\tilde{x}) =\;& \alpha_1 \cdot \text{PSNR}(x,\tilde{x}) + \alpha_2 \cdot \text{SSIM}(x,\tilde{x}) \\
&+ \alpha_3 \cdot (1 - \text{LPIPS}(x,\tilde{x})) + \alpha_4 \cdot \text{CLIP}(x,\tilde{x}).
\end{split}
\end{equation}
and our Authenticity index:
\begin{equation}
\text{A-index}(\mathbf{x}, \tilde{\mathbf{x}}) = \frac{\exp(-\sigma \cdot s(\mathbf{x}, \tilde{\mathbf{x}}))}{1 + \exp(-\sigma \cdot s(\mathbf{x}, \tilde{\mathbf{x}}))}
\end{equation}
we aim to optimize the weights $\alpha_1, \alpha_2, \alpha_3, \alpha_4$ via using Differential Evolution to maximize separation between real and fake image distributions. The objective function minimizes the overlap between distributions:

\begin{equation}
\begin{aligned}
\text{minimize}\quad &\text{overlap}(\mathcal{D}_{\text{real}},
\mathcal{D}_{\text{fake}}) \\
&= \int \min\bigl(p_{\text{real}}(s),p_{\text{fake}}(s)\bigr)\,ds.
\end{aligned}
\end{equation}

\tit{Differential Evolution Parameters}
\begin{itemize}
    \item Population size: 20
    \item Mutation factor ($F$): 0.6
    \item Crossover probability ($CR$): 0.7
    \item Maximum iterations: 300
    \item Bounds: $\alpha_i \in [-10.0, 10.0]$ for $i = 1,2,3,4$
    \item Convergence tolerance: $1 \times 10^{-10}$
\end{itemize}

\tit{Optimized Weights}
\begin{itemize}
    \item $\alpha_1$ (PSNR): -0.0181
    \item $\alpha_2$ (SSIM): 1.380
    \item $\alpha_3$ (1-LPIPS): -4.058
    \item $\alpha_4$ (CLIP): 8.066
    \item $\sigma$ (sigmoid scale): 0.9
\end{itemize}

\section{Inversion Hyperparameters}
We employ Rectified Flow Inversion~\cite{rf_inversion} for all experiments. The key hyperparameters are described below.
\tit{General Inversion Parameters}
\begin{table}[H]
\centering
\caption{General inversion hyperparameters used across all experiments.}
\resizebox{\columnwidth}{!}{%
\begin{tabular}{ccc}
\toprule
\textbf{Parameter} & \textbf{Value} & \textbf{Description} \\
\midrule
Number of inference steps & 28 & Denoising timesteps \\
Guidance scale & 3.5 & Classifier-free guidance strength \\
Scheduler & FlowMatchEulerDiscrete & Sampling scheduler used during inversion \\
DDIM $\eta$ base & 0.95 & Base eta for interpolated denoise \\
Eta trend & constant & Trend of eta values (e.g., constant, linear) \\
Eta start step & 0 & Step where eta schedule starts \\
Eta end step & 9 & Step where eta schedule ends \\
Gamma & 0.5 & Gamma parameter for rectified inversion \\
Seed & 42 & Random seed for reproducibility \\
Data type & float16 & Precision for model computations \\
\bottomrule
\end{tabular}
}

\end{table}

\section{Detector Benchmark}
\label{app:benchmark}
We evaluate twenty detectors released between 2023 and 2026 against ten generators released between December 2022 and May 2026.
Each detector--generator pair uses $100$ authentic and $100$ prompt-matched generated images (seed $42$), all resized and center-cropped to $512\times512$ to remove resolution as a cue.
Each detector runs with its official code, checkpoint, and test-time preprocessing, and with its native decision threshold ($0.5$ on probabilities, $0$ on logits).
\Cref{tab:benchmark_acc} reports balanced accuracy.
AEROBLADE, WaRPAD, and FIRE assign nearly all images to one class at their native threshold, so their accuracy stays at chance.

\begin{table*}[!t]
\centering
\caption{\textbf{Balanced accuracy (\%) of twenty detectors on ten generators.} Each cell uses $100$ authentic and $100$ generated images at $512\times512$. Bold marks the best detector per generator.}
\label{tab:benchmark_acc}
\setlength{\tabcolsep}{3pt}
\footnotesize
\resizebox{\textwidth}{!}{%
\begin{tabular}{lc*{10}{c}}
\toprule
 & & SD2.1 & SD3 & SD3.5 & FLUX.1 & FLUX.1 & Boreal & FLUX.2 & Boreal & HiDream & GPT \\
 & & \cite{rombach2021ldm} & \cite{esser2024sd3} & \cite{stability_sd35} & \cite{flux2024} & +LoRA\cite{xlabs_flux_realismlora} & FLUX.1\cite{boreal_flux1} & \cite{flux2_dev} & FLUX.2\cite{boreal_flux2} & O1\cite{hidream_o1} & Image-2\cite{openai2026chatgptimages2} \\
\textbf{Detector} & \textbf{Year} & 2022 & 2024 & 2024 & 2024 & 2024 & 2024 & 2025 & 2025 & 2026 & 2026 \\
\midrule
UFD~\cite{ufd} & 2023 & 49.5 & 50.0 & 50.5 & 48.5 & 49.0 & 49.0 & 49.0 & 48.5 & 49.0 & 50.0 \\
AEROBLADE~\cite{ricker2024aeroblade} & 2024 & 50.0 & 50.0 & 50.0 & 50.0 & 50.0 & 50.0 & 50.0 & 50.0 & 50.0 & 50.0 \\
FatFormer~\cite{fatformer} & 2024 & 49.5 & 49.0 & 63.0 & 49.0 & 51.0 & 59.5 & 52.5 & 52.0 & 57.0 & 60.5 \\
FreqNet~\cite{freqnet} & 2024 & 46.5 & 44.5 & 90.0 & 82.5 & 84.0 & 81.0 & 83.0 & 74.5 & 78.5 & 81.0 \\
NPR~\cite{npr} & 2024 & 35.0 & 37.0 & 74.5 & 80.0 & 75.5 & 80.0 & 80.0 & 80.0 & 80.0 & 79.5 \\
C2PClip~\cite{c2pclip} & 2025 & 50.0 & 50.0 & 60.0 & 52.5 & 50.5 & 51.5 & 50.0 & 50.5 & 52.5 & 51.0 \\
D3~\cite{d3} & 2025 & 81.0 & 82.5 & \textbf{98.5} & 96.0 & 87.5 & 81.0 & 93.0 & 76.0 & 91.5 & 84.0 \\
DDA~\cite{dda} & 2025 & 97.0 & 66.5 & 97.5 & 68.0 & 61.0 & 62.0 & 85.0 & 69.5 & 82.5 & 56.5 \\
FIRE~\cite{chu2025fire} & 2025 & 49.5 & 50.0 & 50.0 & 50.0 & 50.0 & 50.0 & 50.0 & 50.0 & 50.0 & 50.0 \\
FerretNet~\cite{ferretnet} & 2025 & 50.5 & 49.0 & \textbf{98.5} & 93.5 & \textbf{95.0} & \textbf{98.0} & 87.0 & \textbf{98.5} & \textbf{98.0} & \textbf{99.5} \\
WaRPAD~\cite{warpad} & 2025 & 50.0 & 50.0 & 50.0 & 50.0 & 50.0 & 50.0 & 50.0 & 50.0 & 50.0 & 50.0 \\
AllPatchesMatter~\cite{allpatchesmatter} & 2026 & 52.5 & 55.0 & 97.5 & 59.5 & 58.0 & 63.5 & 85.5 & 64.5 & 74.5 & 92.0 \\
DEAR~\cite{dear} & 2026 & 80.5 & 52.0 & 94.0 & 82.0 & 69.0 & 56.5 & 70.5 & 57.0 & 71.0 & 51.0 \\
DGS-Net~\cite{dgsnet} & 2026 & 45.5 & 42.5 & 94.0 & 92.5 & 66.5 & 84.5 & \textbf{95.0} & 89.0 & 91.0 & 93.0 \\
ForensicConcept~\cite{forensicconcept} & 2026 & 55.0 & 58.5 & 89.0 & 62.5 & 54.0 & 63.5 & 68.0 & 58.0 & 62.0 & 74.0 \\
IAPL~\cite{iapl} & 2026 & 50.0 & 51.5 & 62.5 & 50.0 & 49.0 & 53.5 & 50.0 & 50.0 & 50.5 & 61.0 \\
OmniAID~\cite{omniaid} & 2026 & 98.0 & 91.5 & 92.5 & 89.5 & 86.0 & 57.0 & 72.0 & 57.5 & 90.5 & 50.5 \\
PGC~\cite{pgc} & 2026 & 81.0 & 57.5 & 74.0 & 54.0 & 53.5 & 54.0 & 56.0 & 49.5 & 69.5 & 59.0 \\
PROBE~\cite{probe} & 2026 & \textbf{99.5} & 66.0 & 93.0 & 49.5 & 51.0 & 49.5 & 55.5 & 51.0 & 56.5 & 54.0 \\
SICA~\cite{sica} & 2026 & 95.5 & \textbf{96.0} & \textbf{98.5} & \textbf{97.0} & 90.5 & 79.5 & \textbf{95.0} & 70.5 & \textbf{98.0} & 70.0 \\
\bottomrule
\end{tabular}}
\end{table*}

\section{Baseline Detectors}
\label{app:baselines}

We evaluate twenty passive detectors.
PROBE~\cite{probe}, SICA~\cite{sica}, and D3~\cite{d3} are described in \Cref{sec:background}.
We briefly summarize the remaining seventeen, grouped by the taxonomy of \Cref{sec:background}.

\tinytit{Artifact methods}
\begin{itemize}
    \item \textbf{FreqNet}~\cite{freqnet} learns in the frequency domain and forces the detector to focus on high-frequency information.
    \item \textbf{NPR}~\cite{npr} detects traces of the up-sampling operations shared by generator architectures through neighboring pixel relationships.
    \item \textbf{FerretNet}~\cite{ferretnet} detects synthetic images efficiently from local pixel dependencies.
    \item \textbf{APM}~\cite{allpatchesmatter} treats every image patch as carrying artifacts and trains the detector on all patches instead of a global view.
    \item \textbf{WaRPAD}~\cite{warpad} is training-free and scores how robust an image's features are to cropping.
\end{itemize}

\tinytit{Feature classifiers}
\begin{itemize}
    \item \textbf{UFD}~\cite{ufd} performs nearest-neighbor and linear probing in the feature space of a frozen CLIP model instead of training a deep classifier.
    \item \textbf{FatFormer}~\cite{fatformer} adapts CLIP with forgery-aware adapters and language-guided alignment.
    \item \textbf{C2P-CLIP}~\cite{c2pclip} injects category-common prompts into the CLIP text encoder.
    \item \textbf{DGS-Net}~\cite{dgsnet} fine-tunes CLIP with distillation-guided gradient surgery to keep pretrained knowledge while learning forgery cues.
    \item \textbf{IAPL}~\cite{iapl} learns prompts that adapt to each input image.
    \item \textbf{ForensicCpt}~\cite{forensicconcept} localizes decision-critical patches, clusters them into a codebook of forensic concepts, and transfers these concepts across backbones.
    \item \textbf{OmniAID}~\cite{omniaid} decouples semantic content from generation artifacts for detection in the wild.
    \item \textbf{PGC}~\cite{pgc} aggregates the most discriminative local clues through peak-sensitive aggregation and calibrates the global decision with them.
    \item \textbf{DEAR}~\cite{dear} prunes feature channels aligned with spurious cues, identified through inpainted regions, to keep only features that capture generative artifacts.
    \item \textbf{DDA}~\cite{dda} aligns real and synthetic training data in the pixel and frequency domains to remove spurious correlations.
\end{itemize}

\tinytit{Reconstruction methods}
\begin{itemize}
    \item \textbf{AEROBLADE}~\cite{ricker2024aeroblade} is training-free and thresholds the reconstruction error of the latent diffusion autoencoder.
    \item \textbf{FIRE}~\cite{chu2025fire} detects diffusion-generated images through frequency-guided reconstruction error.
\end{itemize}

\section{Zero-Shot Detection Graphs for All Baseline Methods}

We report the prediction distributions of all baseline detectors under a zero-shot setting.These detectors include UFD~\cite{ufd}, FreqNet~\cite{freqnet}, NPR~\cite{npr}, FatFormer~\cite{fatformer}, D3~\cite{d3}, and C2P-CLIP~\cite{c2pclip}. The evaluation was conducted on test samples generated by models unseen during training, namely Stable Diffusion 2.1, Stable Diffusion XL, Stable Diffusion 3 (medium), Stable Diffusion 3.5 (medium), DALL·E 3, and Midjourney 6.

\begin{figure*}[!p]
      \centering
      \begin{subfigure}[b]{0.32\textwidth}
          \includegraphics[width=\linewidth]{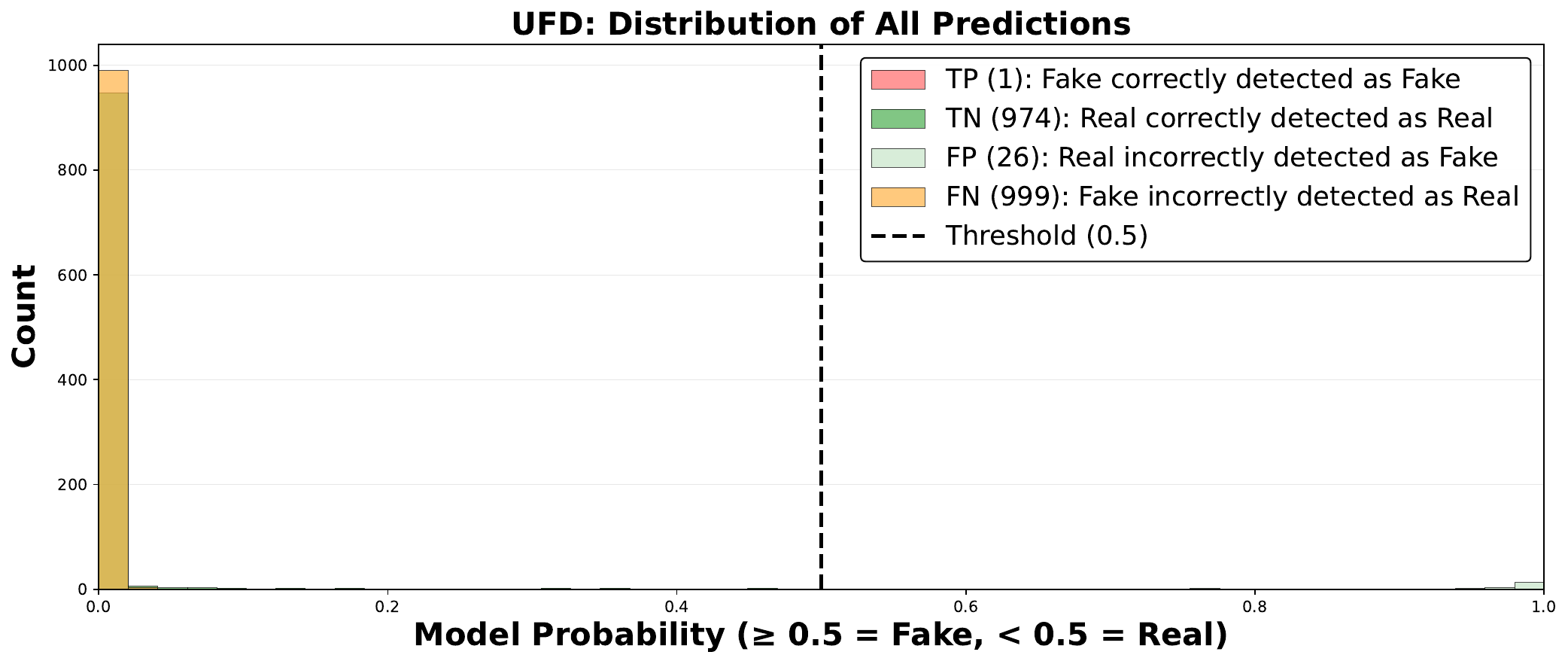}
          \caption{UFD}
      \end{subfigure}
      \hfill
      \begin{subfigure}[b]{0.32\textwidth}
          \includegraphics[width=\linewidth]{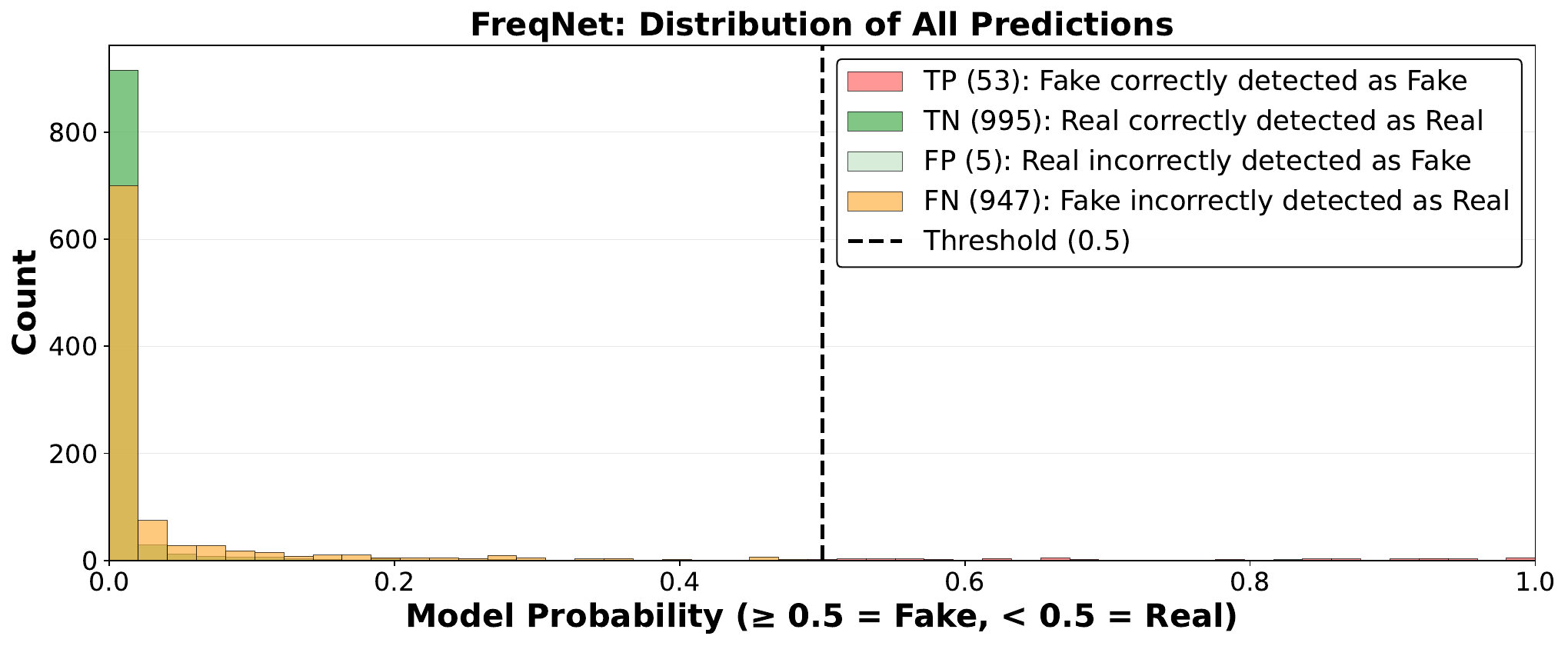}
          \caption{FreqNet}
      \end{subfigure}
      \hfill
      \begin{subfigure}[b]{0.32\textwidth}
          \includegraphics[width=\linewidth]{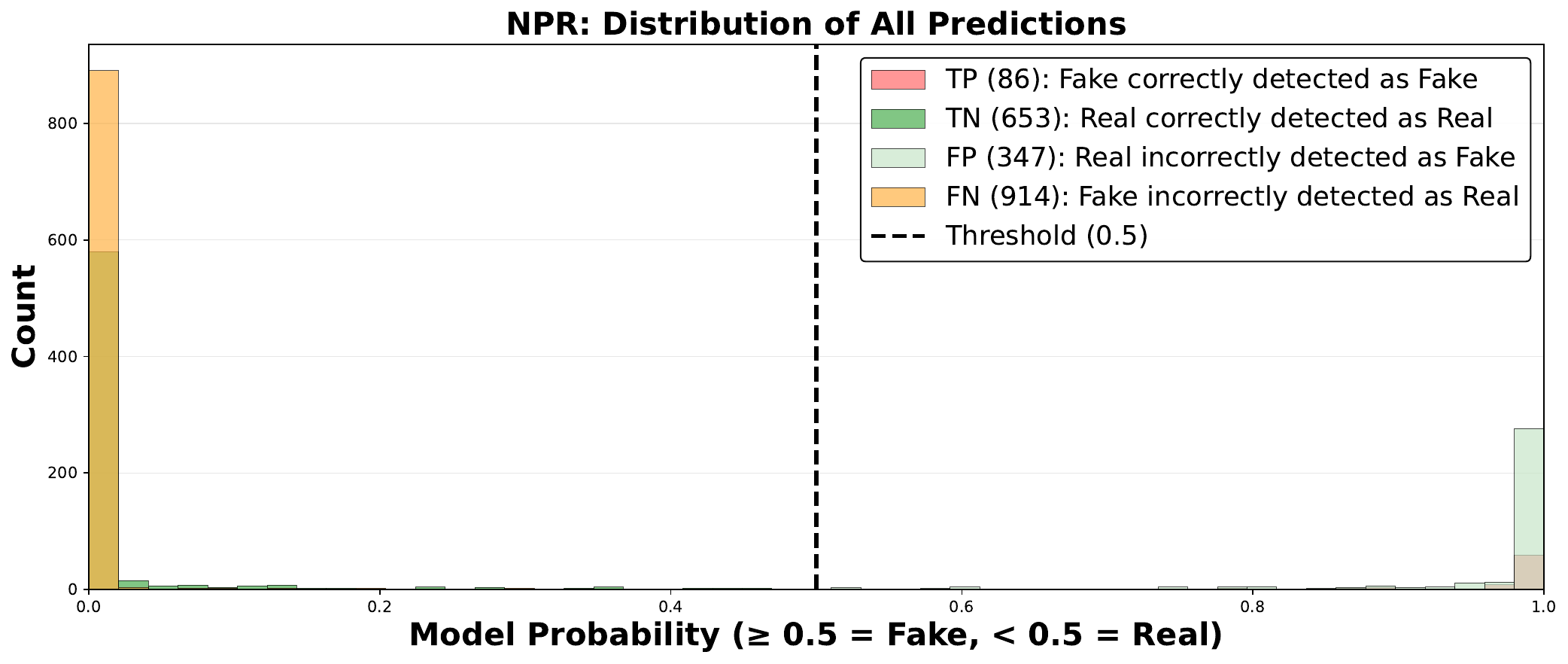}
          \caption{NPR}
      \end{subfigure}

      \vspace{0.04cm}

      \begin{subfigure}[b]{0.32\textwidth}
          \includegraphics[width=\linewidth]{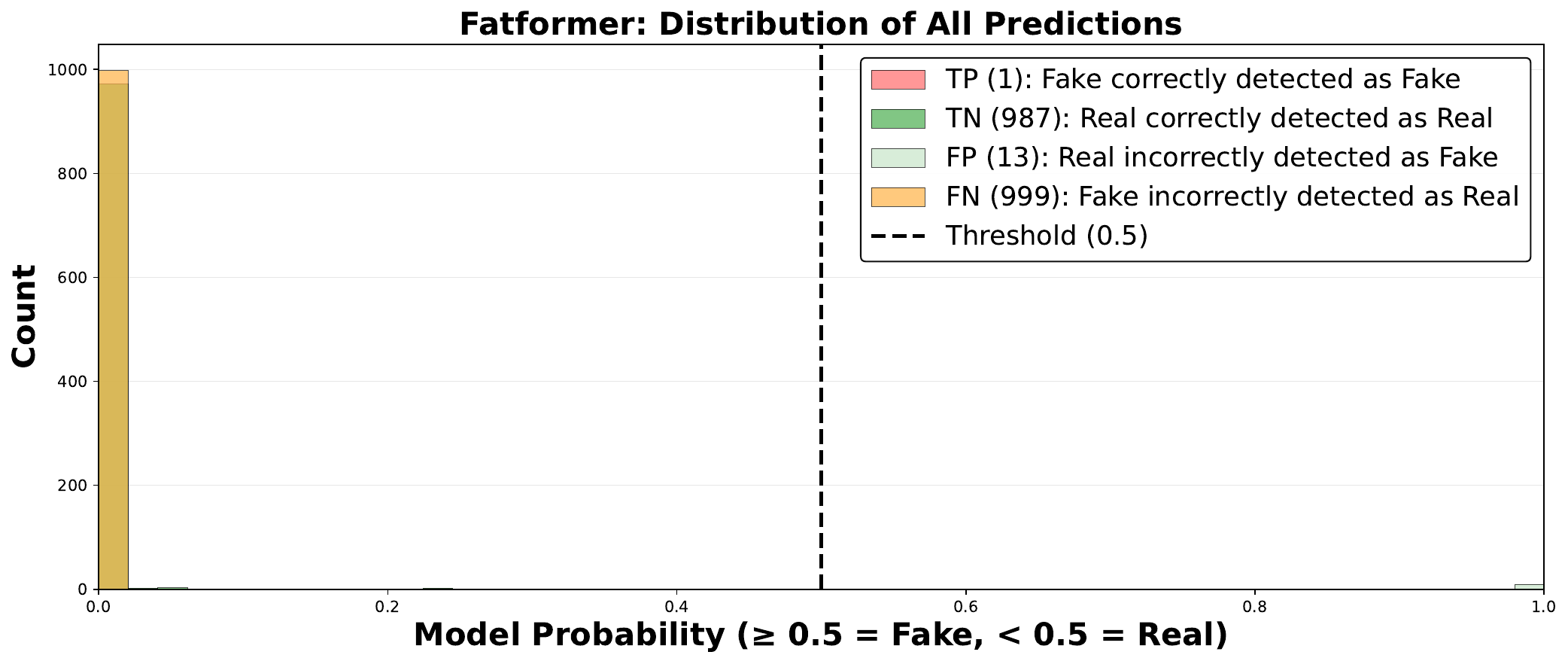}
          \caption{FatFormer}
      \end{subfigure}
      \hfill
      \begin{subfigure}[b]{0.32\textwidth}
          \includegraphics[width=\linewidth]{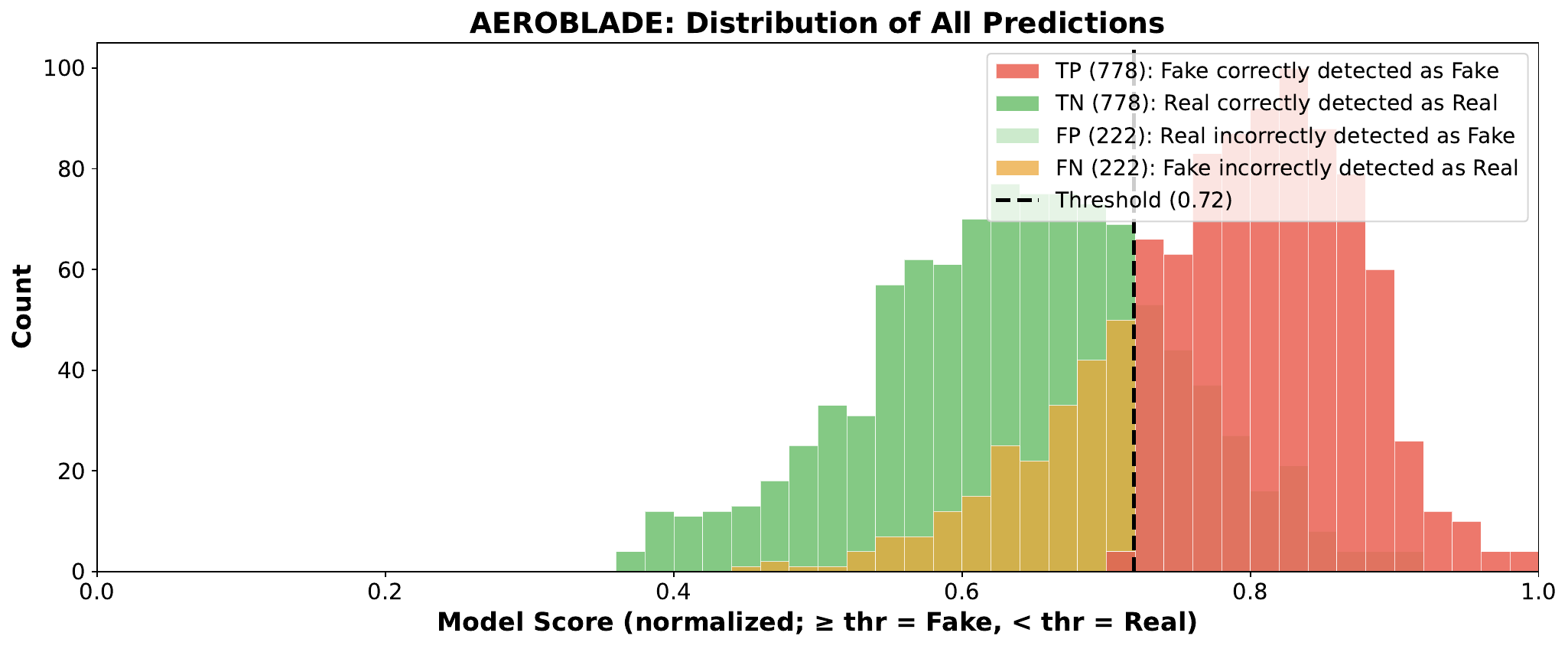}
          \caption{AEROBLADE}
      \end{subfigure}
      \hfill
      \begin{subfigure}[b]{0.32\textwidth}
          \includegraphics[width=\linewidth]{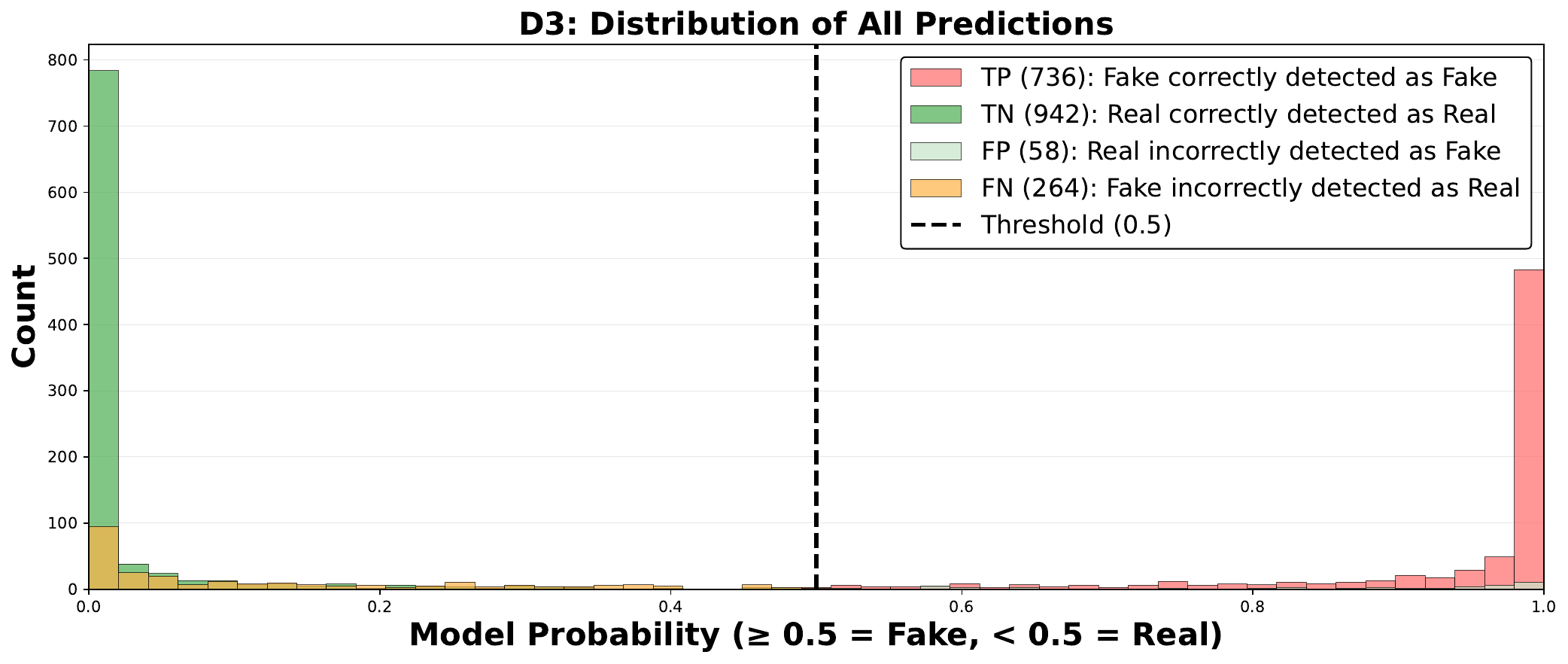}
          \caption{D3}
      \end{subfigure}

      \vspace{0.04cm}

      \begin{subfigure}[b]{0.32\textwidth}
          \includegraphics[width=\linewidth]{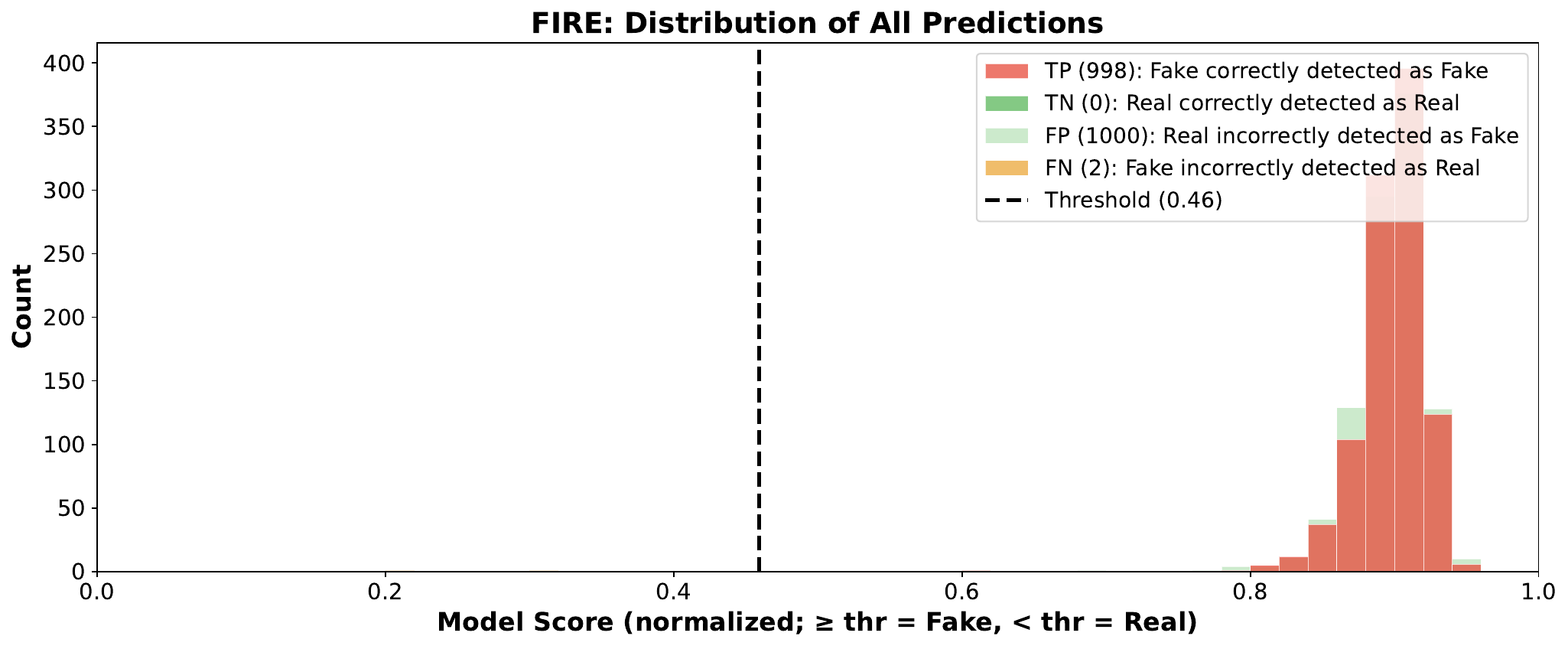}
          \caption{FIRE}
      \end{subfigure}
      \hfill
      \begin{subfigure}[b]{0.32\textwidth}
          \includegraphics[width=\linewidth]{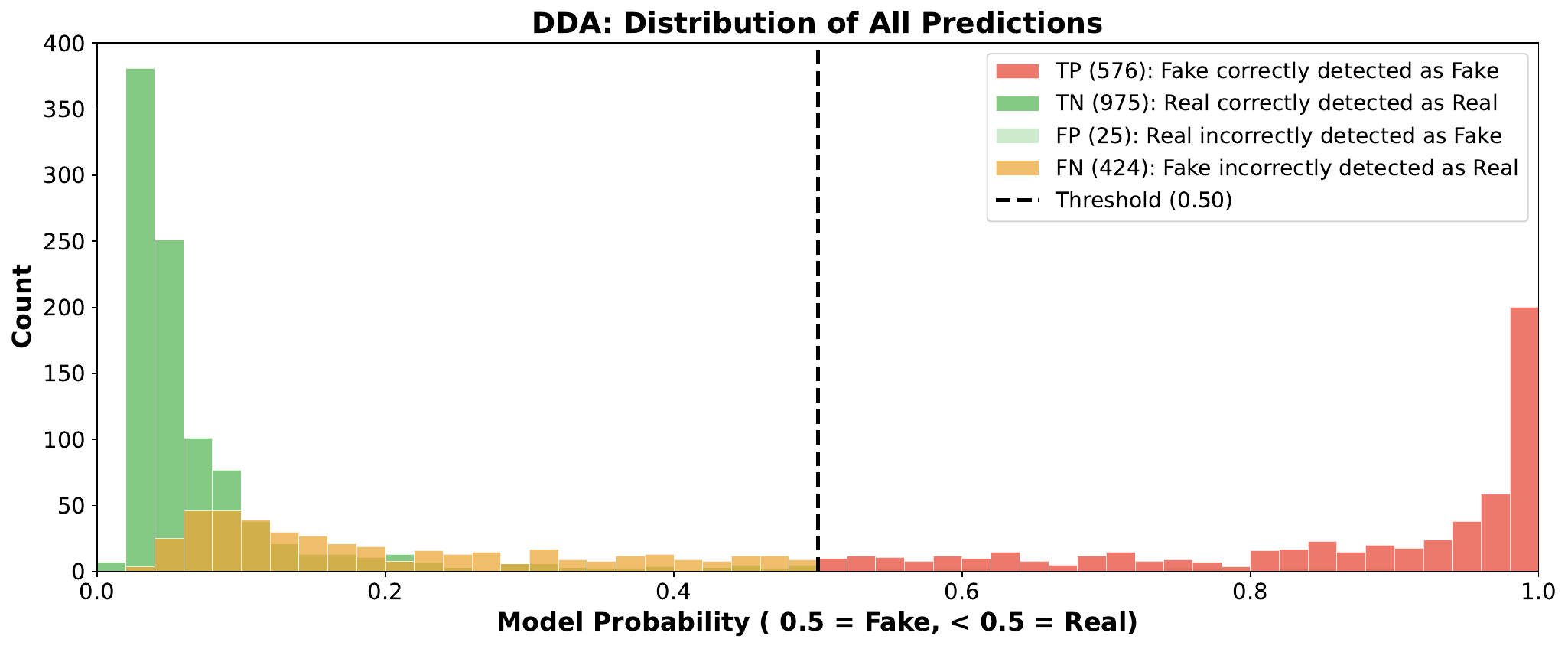}
          \caption{DDA}
      \end{subfigure}
      \hfill
      \begin{subfigure}[b]{0.32\textwidth}
          \includegraphics[width=\linewidth]{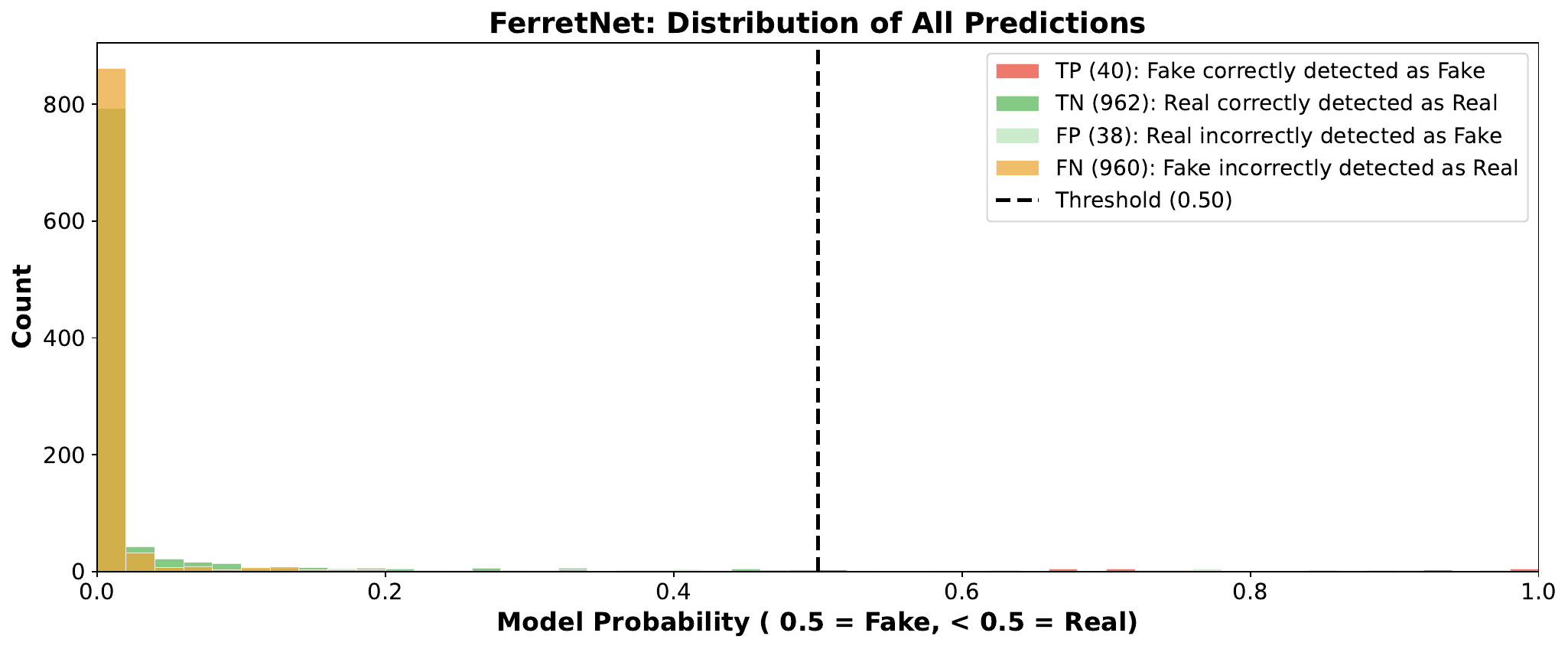}
          \caption{FerretNet}
      \end{subfigure}

      \vspace{0.04cm}

      \begin{subfigure}[b]{0.32\textwidth}
          \includegraphics[width=\linewidth]{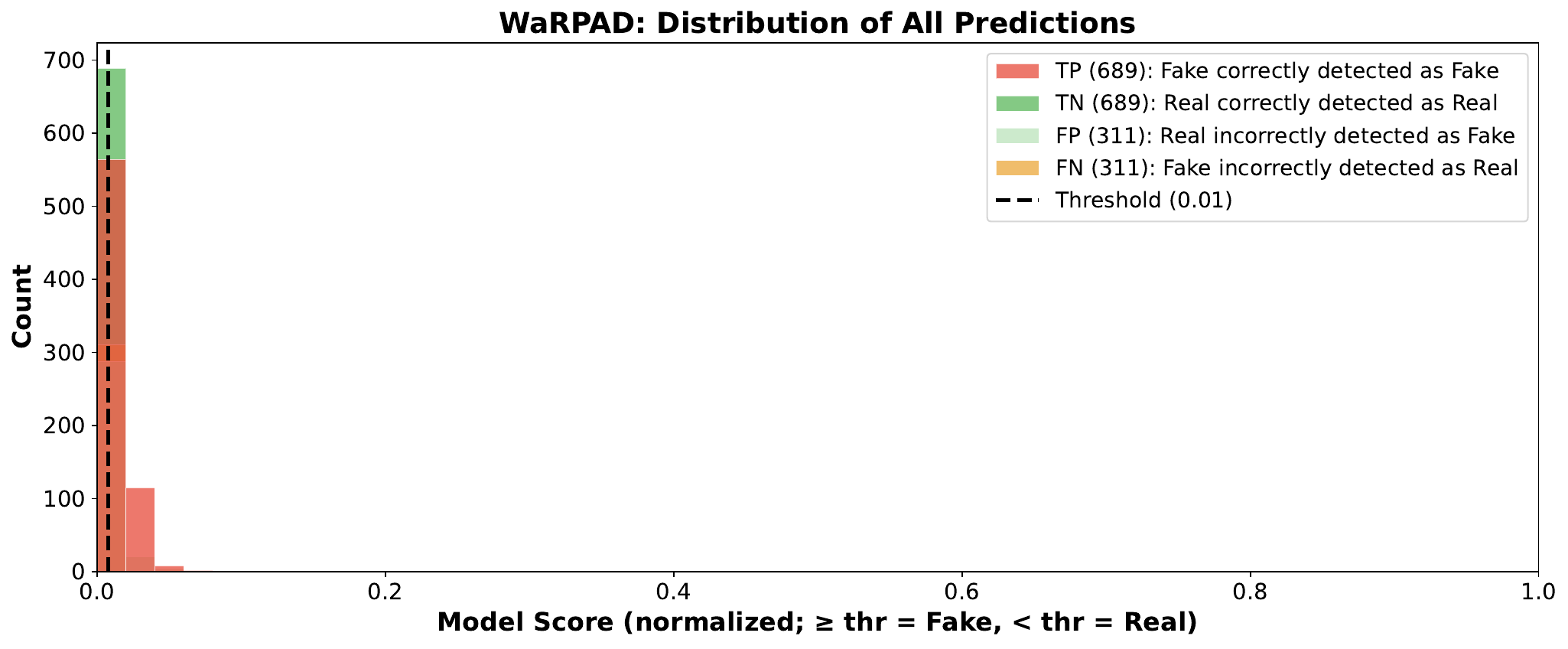}
          \caption{WaRPAD}
      \end{subfigure}
      \hfill
      \begin{subfigure}[b]{0.32\textwidth}
          \includegraphics[width=\linewidth]{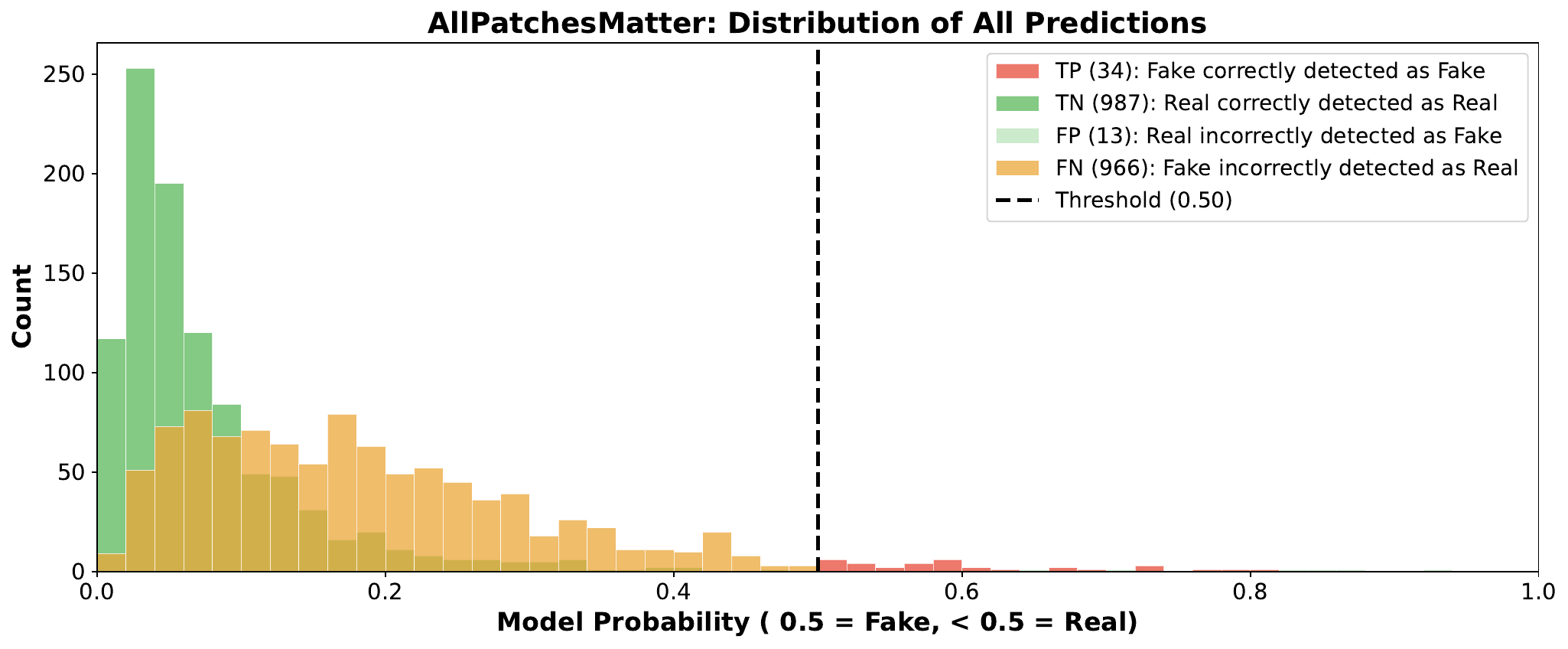}
          \caption{AllPatchesMatter}
      \end{subfigure}
      \hfill
      \begin{subfigure}[b]{0.32\textwidth}
          \includegraphics[width=\linewidth]{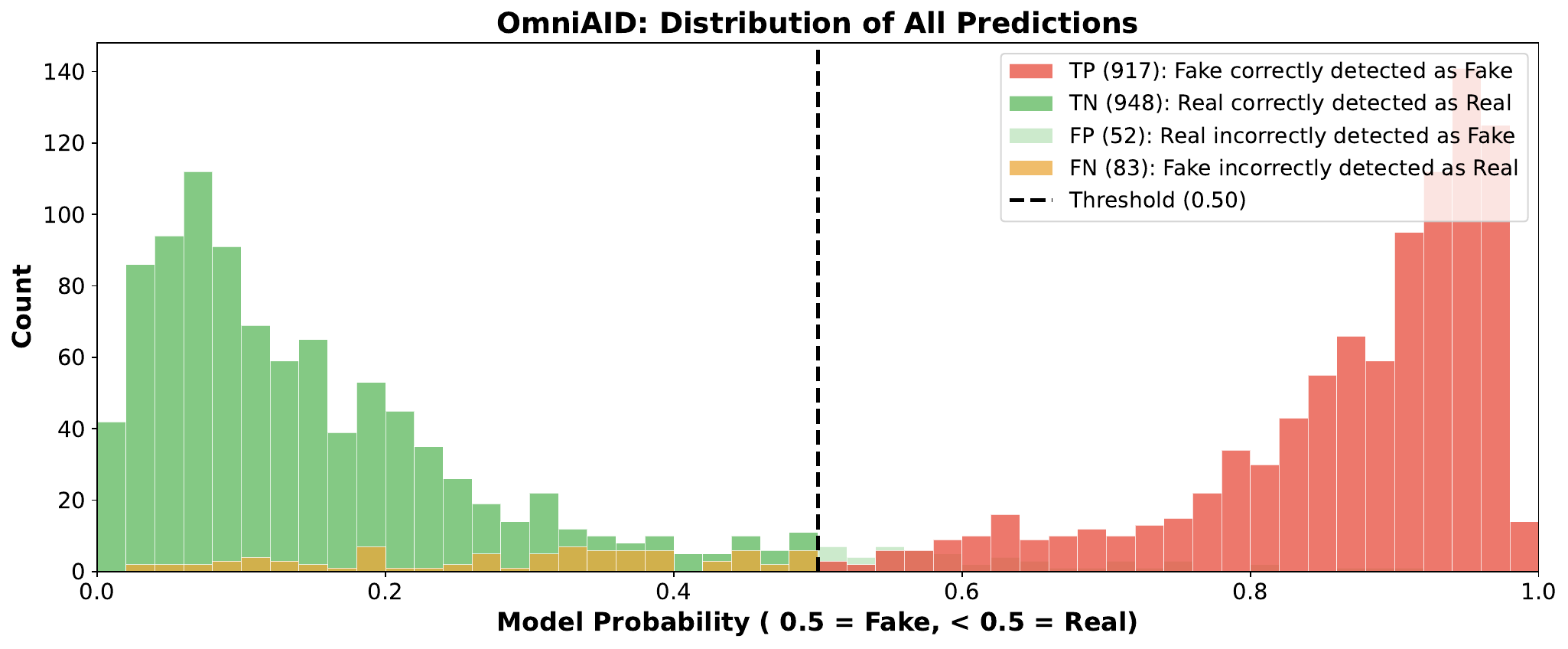}
          \caption{OmniAID}
      \end{subfigure}

      \vspace{0.04cm}

      \begin{subfigure}[b]{0.32\textwidth}
          \includegraphics[width=\linewidth]{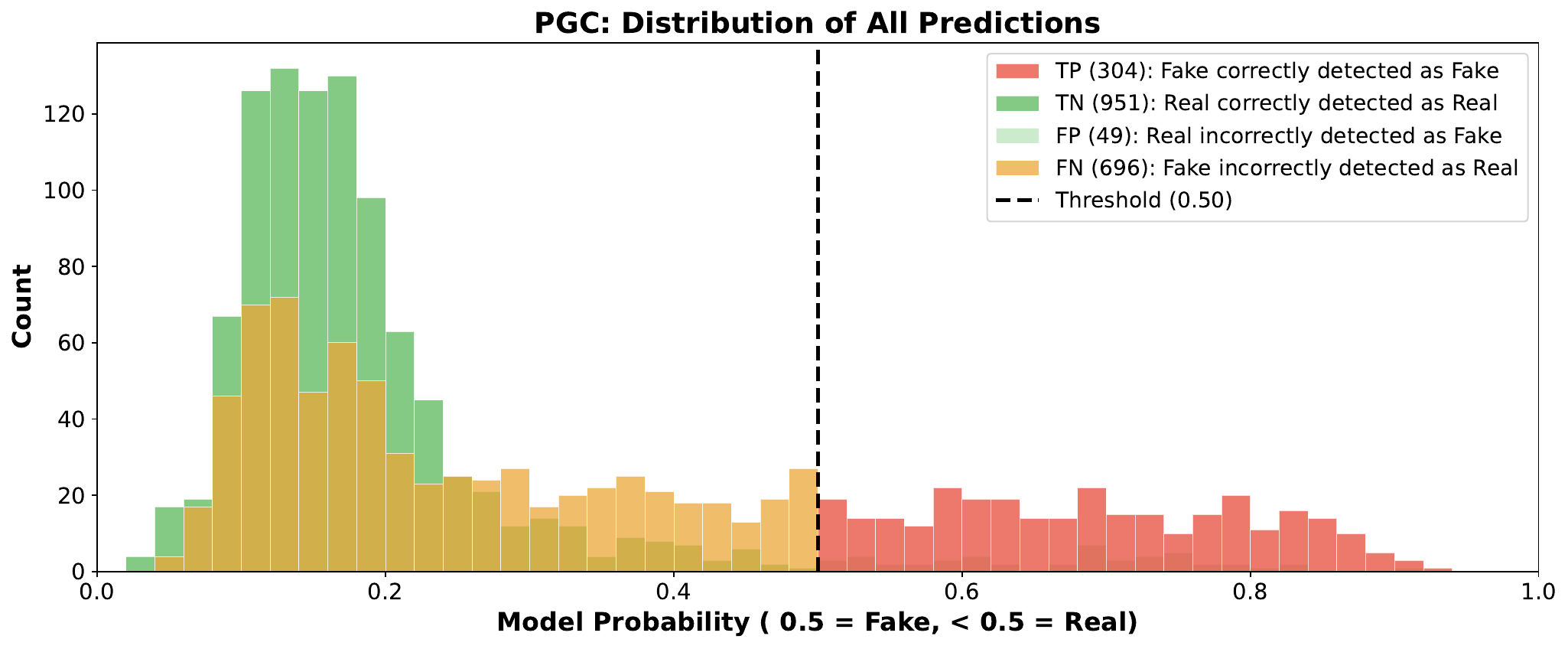}
          \caption{PGC}
      \end{subfigure}
      \hfill
      \begin{subfigure}[b]{0.32\textwidth}
          \includegraphics[width=\linewidth]{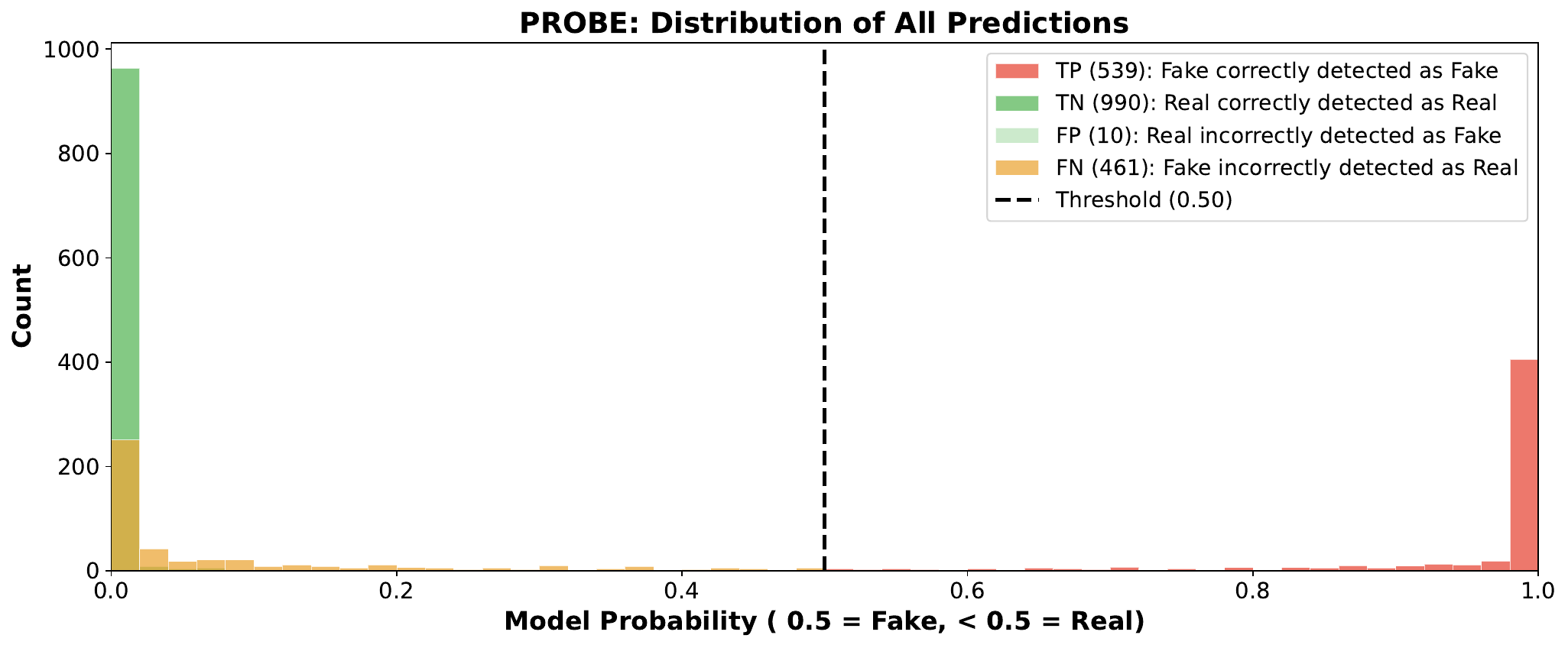}
          \caption{PROBE}
      \end{subfigure}
      \hfill
      \begin{subfigure}[b]{0.32\textwidth}
          \includegraphics[width=\linewidth]{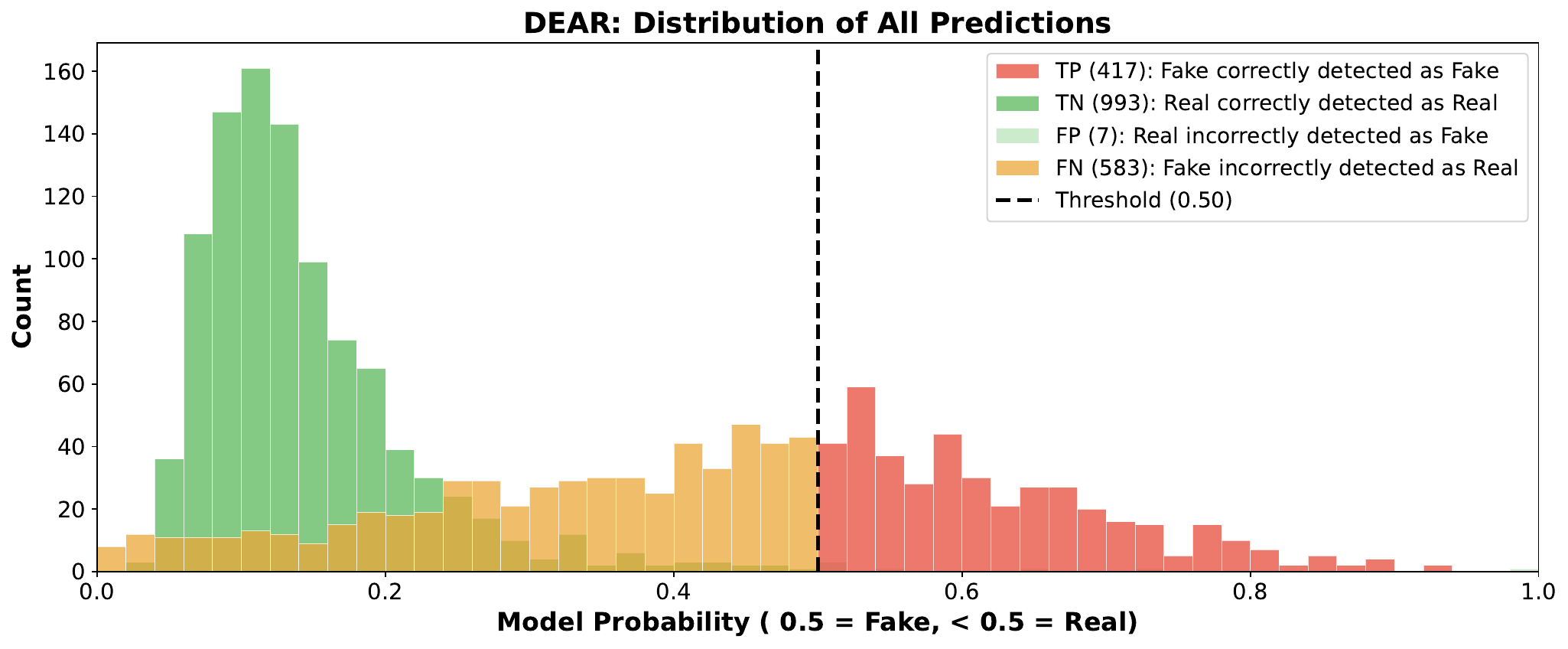}
          \caption{DEAR}
      \end{subfigure}

      \vspace{0.04cm}

      \begin{subfigure}[b]{0.32\textwidth}
          \includegraphics[width=\linewidth]{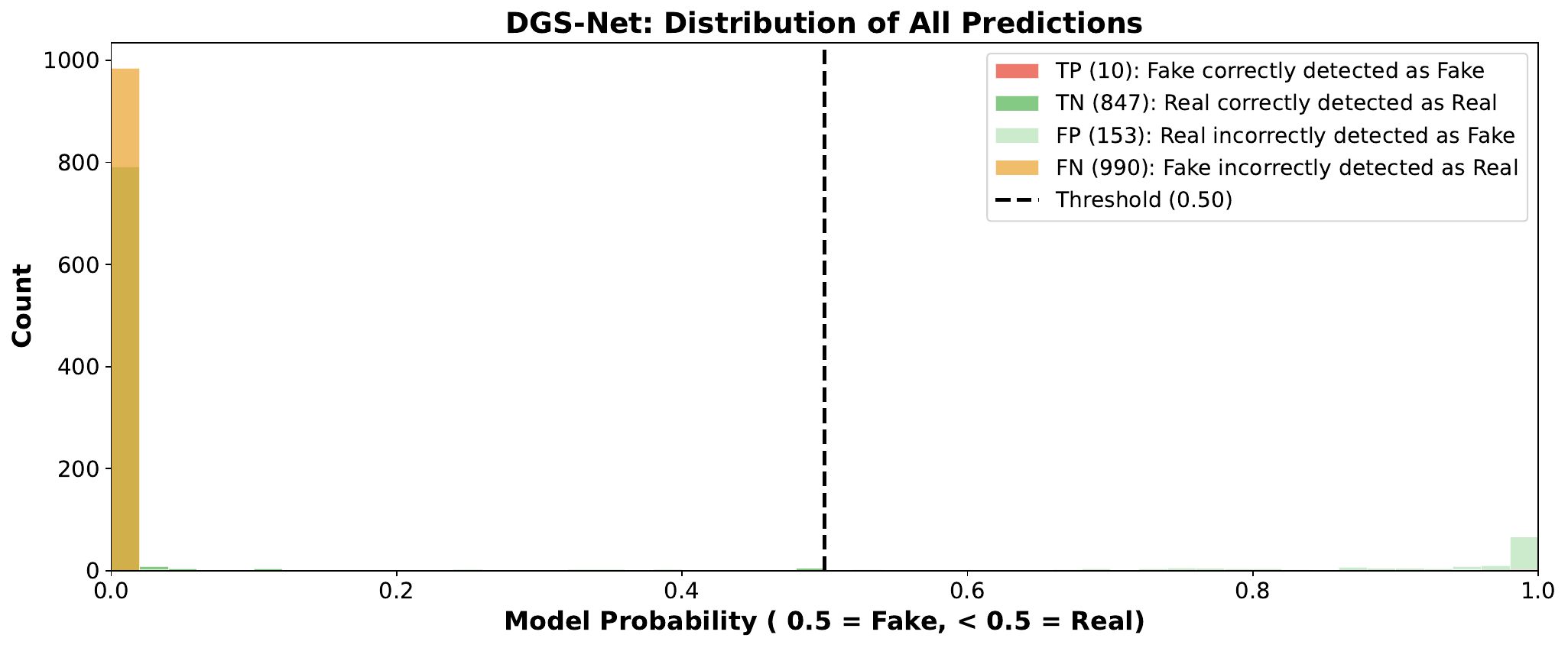}
          \caption{DGS-Net}
      \end{subfigure}
      \hfill
      \begin{subfigure}[b]{0.32\textwidth}
          \includegraphics[width=\linewidth]{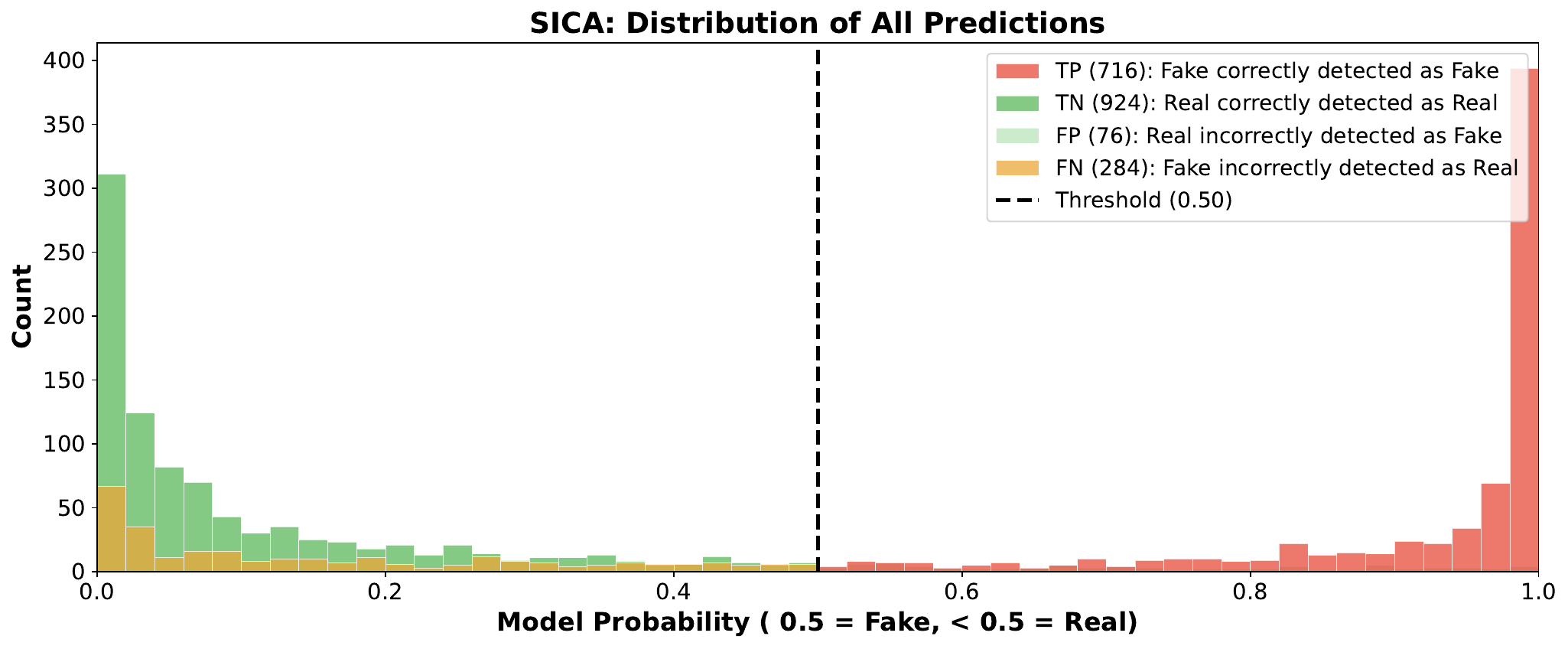}
          \caption{SICA}
      \end{subfigure}
      \hfill
      \begin{subfigure}[b]{0.32\textwidth}
          \includegraphics[width=\linewidth]{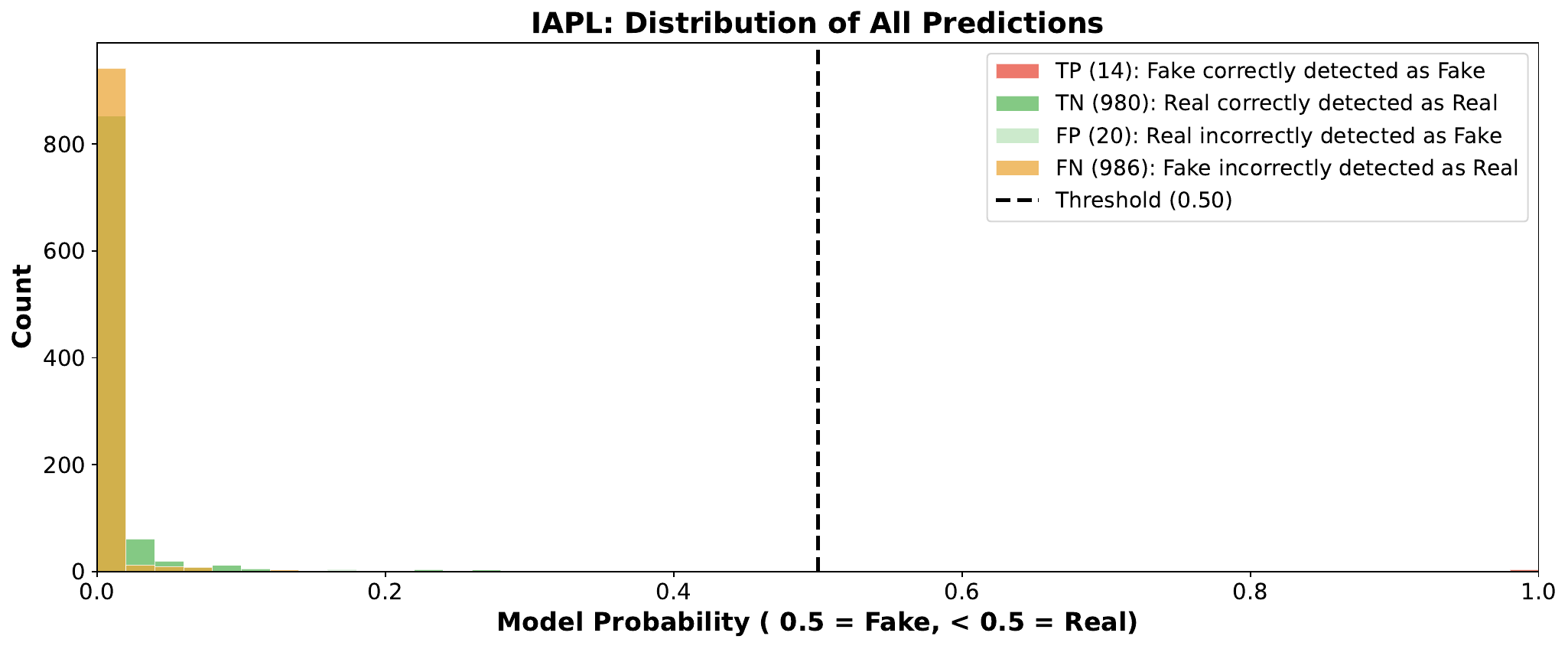}
          \caption{IAPL}
      \end{subfigure}

      \vspace{0.04cm}

      \begin{subfigure}[b]{0.32\textwidth}
          \includegraphics[width=\linewidth]{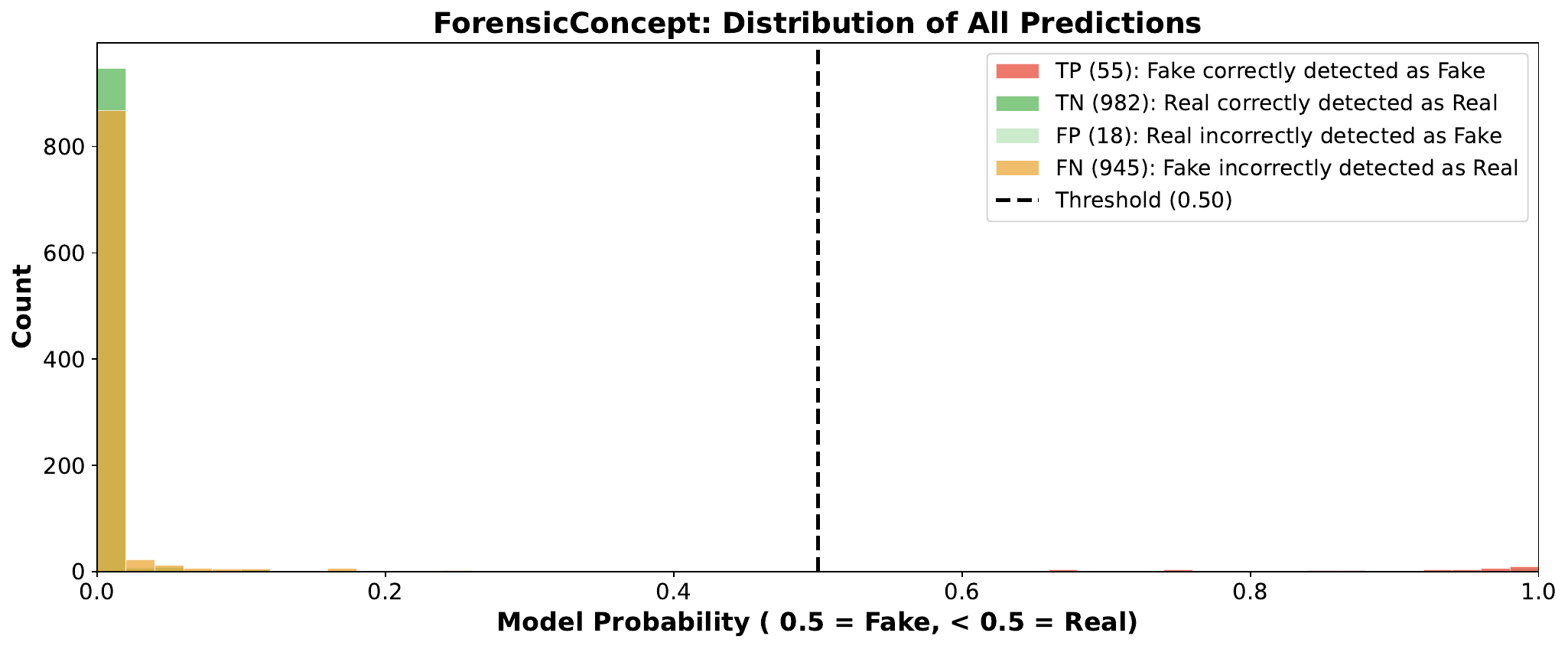}
          \caption{ForensicConcept}
      \end{subfigure}
      \caption{Zero-shot prediction score distributions for each evaluated detector. Each subplot shows the
  histogram of classifier outputs for real and fake images. Vertical dashed lines indicate classification
  thresholds. Most methods struggle to generalize to unseen generators, misclassifying a significant number
  of fake images as real.}
      \label{fig:zero_shot_all}
  \end{figure*}

Our observations indicate that most methods exhibit substantial prediction score overlap between real and fake classes. As shown in Figure \ref{fig:zero_shot_all}, models like UFD, FreqNet, and NPR tend to overfit to familiar artifacts and thus default to high-confidence predictions of “real” even when presented with synthetic samples from unseen generators. Although D3 performs slightly better, it still suffers from notable misclassification rates. These plots underline the generalization weakness of current binary detectors and reinforce the motivation behind adopting a calibrated authenticity framework.

\section{PGD Attack Graphs for All Baseline Methods}

To further assess robustness, we visualize how the prediction distributions of five baseline detectors change under adversarial perturbation. Specifically, we perform projected gradient descent (PGD) attacks~\cite{pgd} with $\ell_\infty$-norm bounded perturbations ($\epsilon = 8/255$), targeting both real and fake inputs. The objective is to push predictions across the decision boundary using imperceptible noise.

\begin{figure*}[!p]
      \centering
      \begin{subfigure}[b]{0.30\textwidth}
          \includegraphics[width=\linewidth]{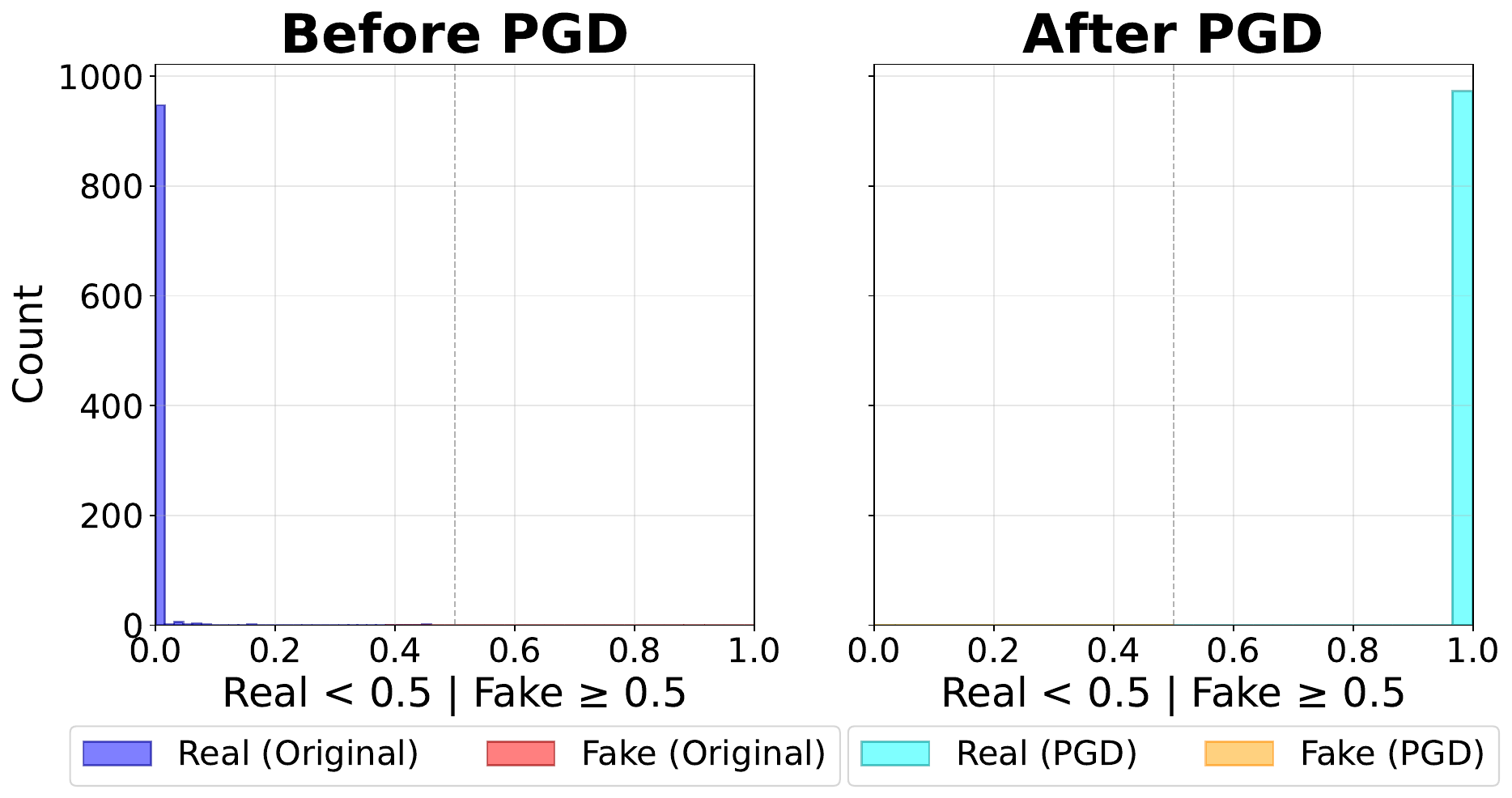}
          \caption{UFD}
      \end{subfigure}
      \hfill
      \begin{subfigure}[b]{0.30\textwidth}
          \includegraphics[width=\linewidth]{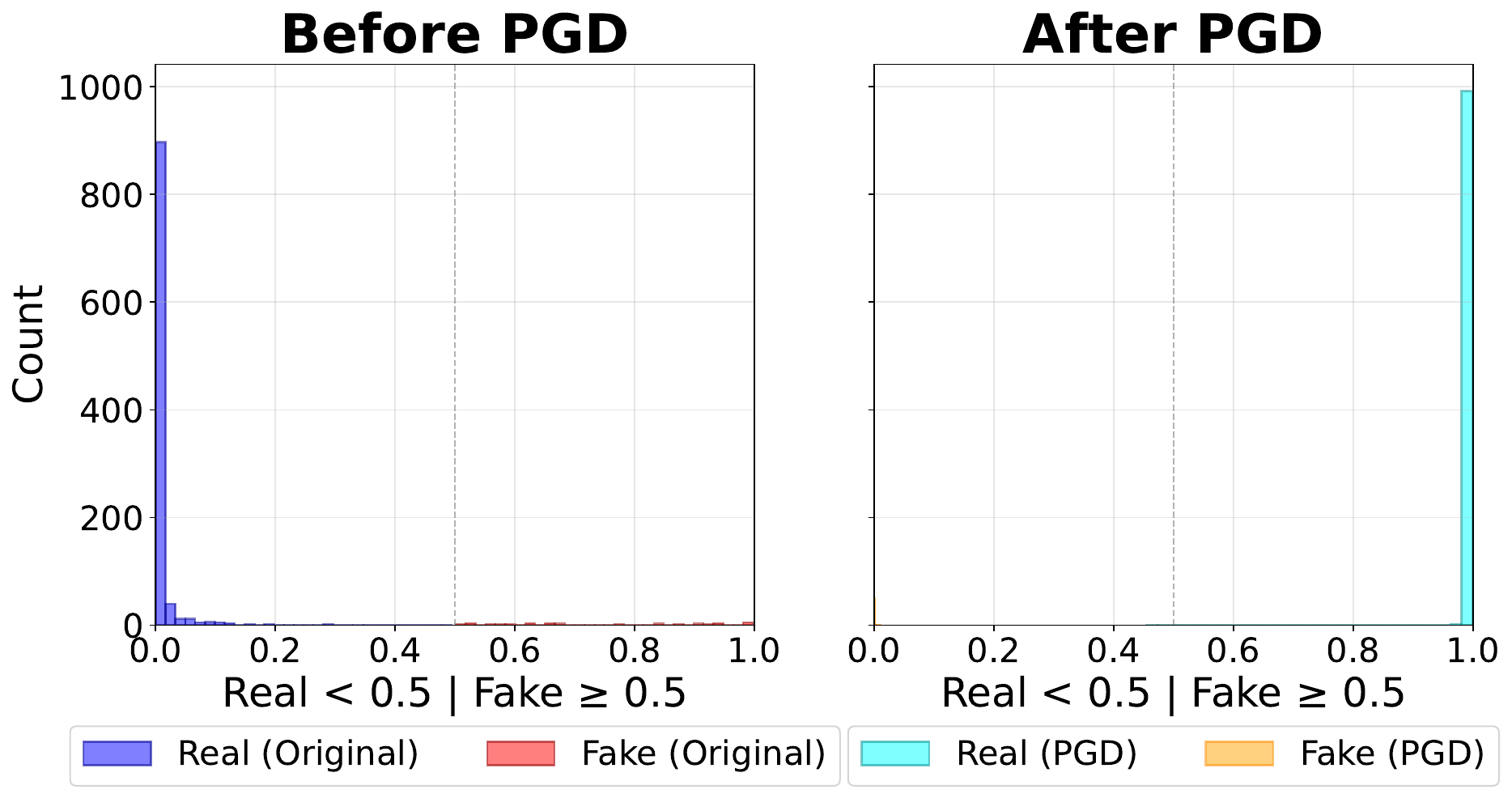}
          \caption{FreqNet}
      \end{subfigure}
      \hfill
      \begin{subfigure}[b]{0.30\textwidth}
          \includegraphics[width=\linewidth]{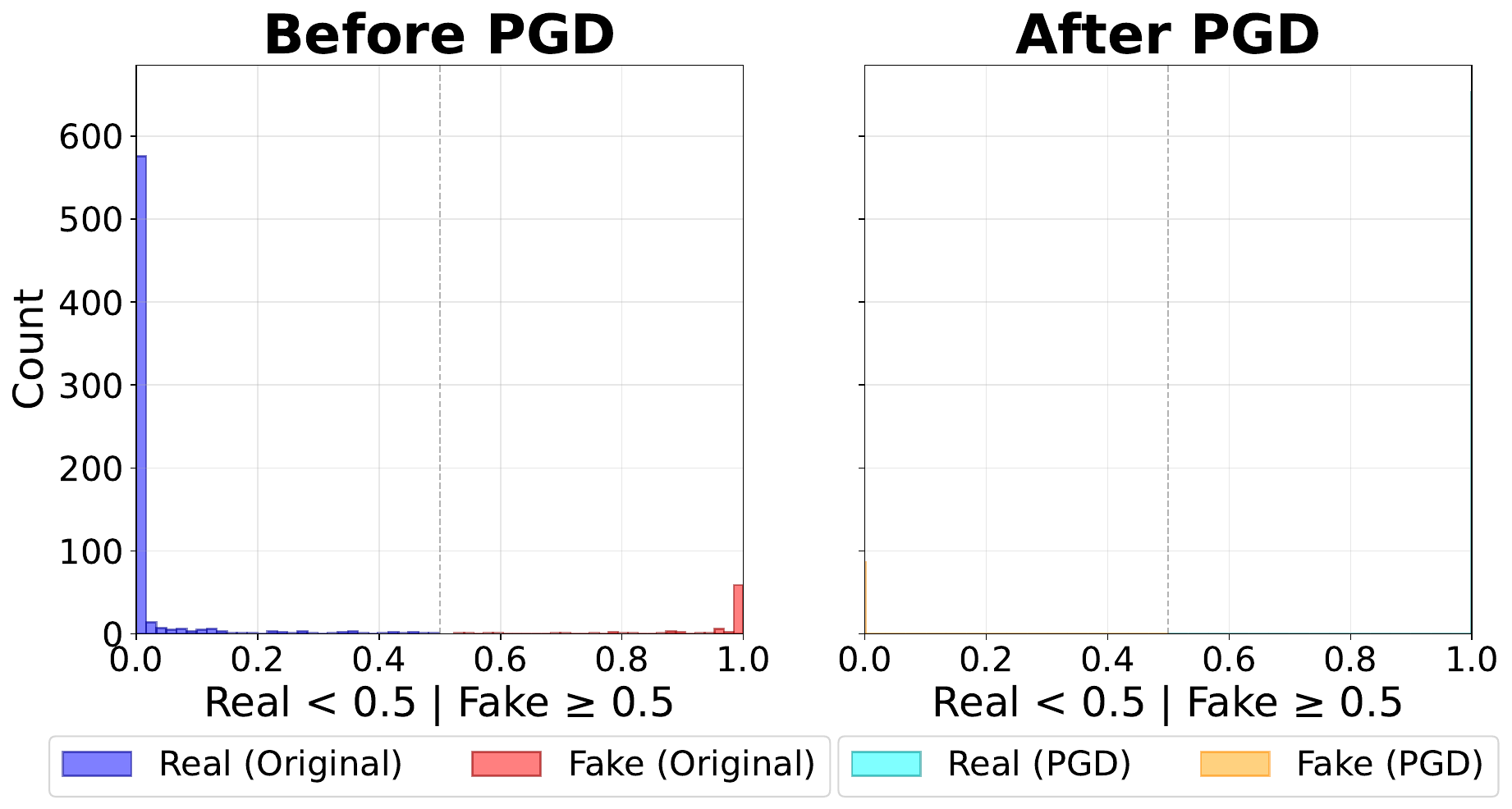}
          \caption{NPR}
      \end{subfigure}

      \vspace{0.05cm}

      \begin{subfigure}[b]{0.30\textwidth}
          \includegraphics[width=\linewidth]{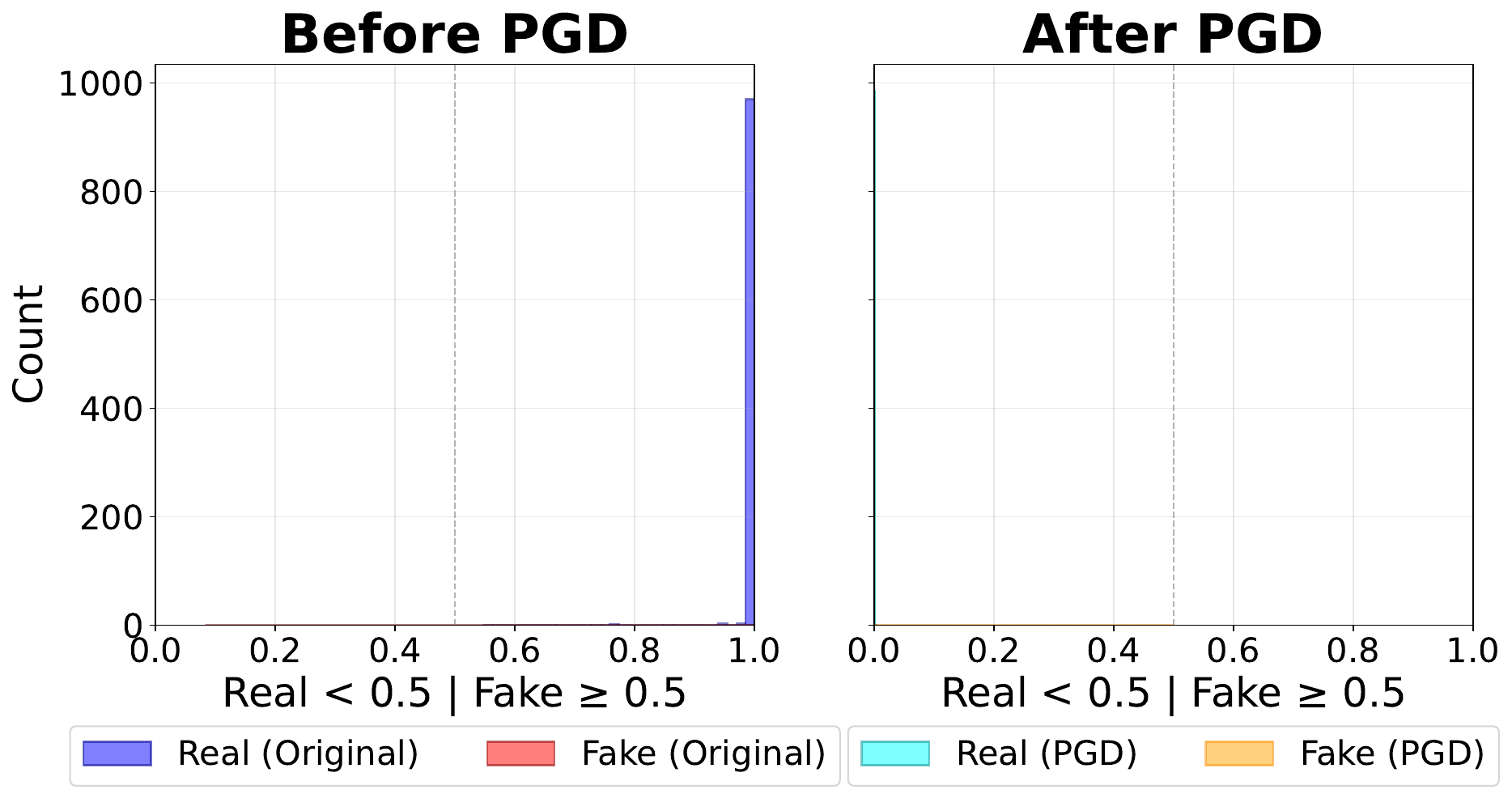}
          \caption{FatFormer}
      \end{subfigure}
      \hfill
      \begin{subfigure}[b]{0.30\textwidth}
          \includegraphics[width=\linewidth]{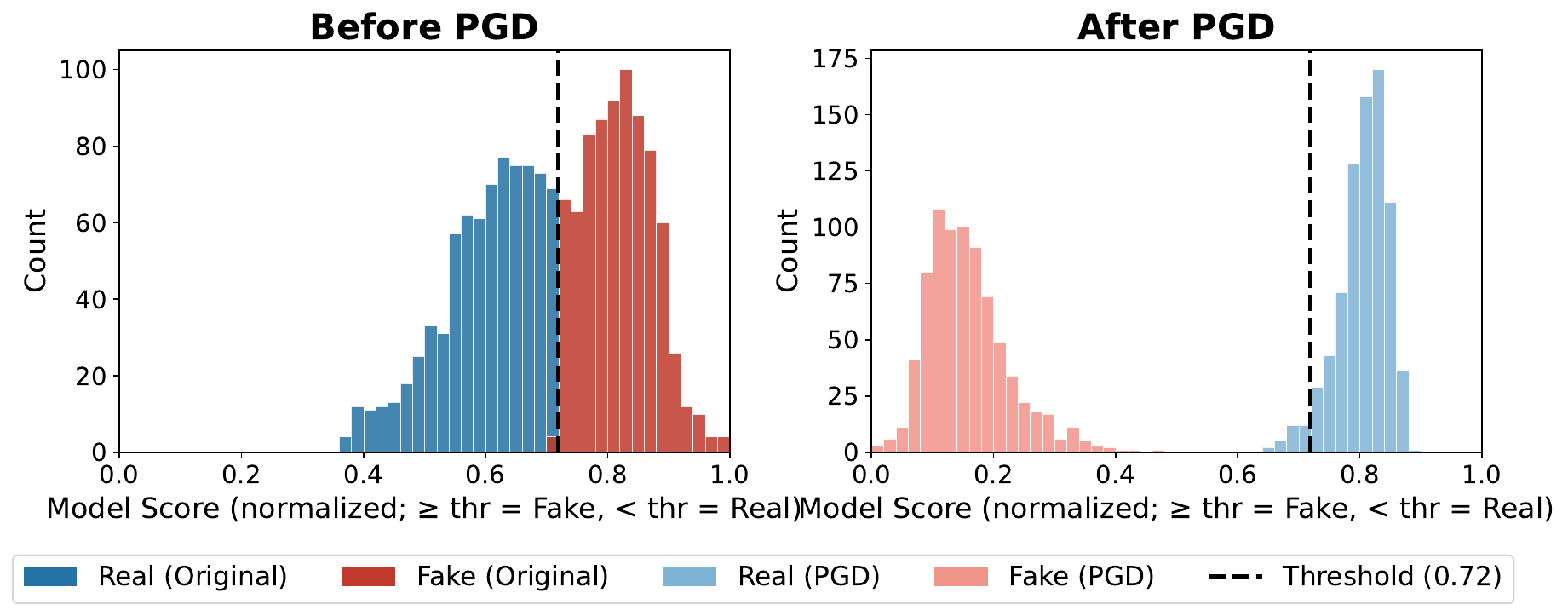}
          \caption{AEROBLADE}
      \end{subfigure}
      \hfill
      \begin{subfigure}[b]{0.30\textwidth}
          \includegraphics[width=\linewidth]{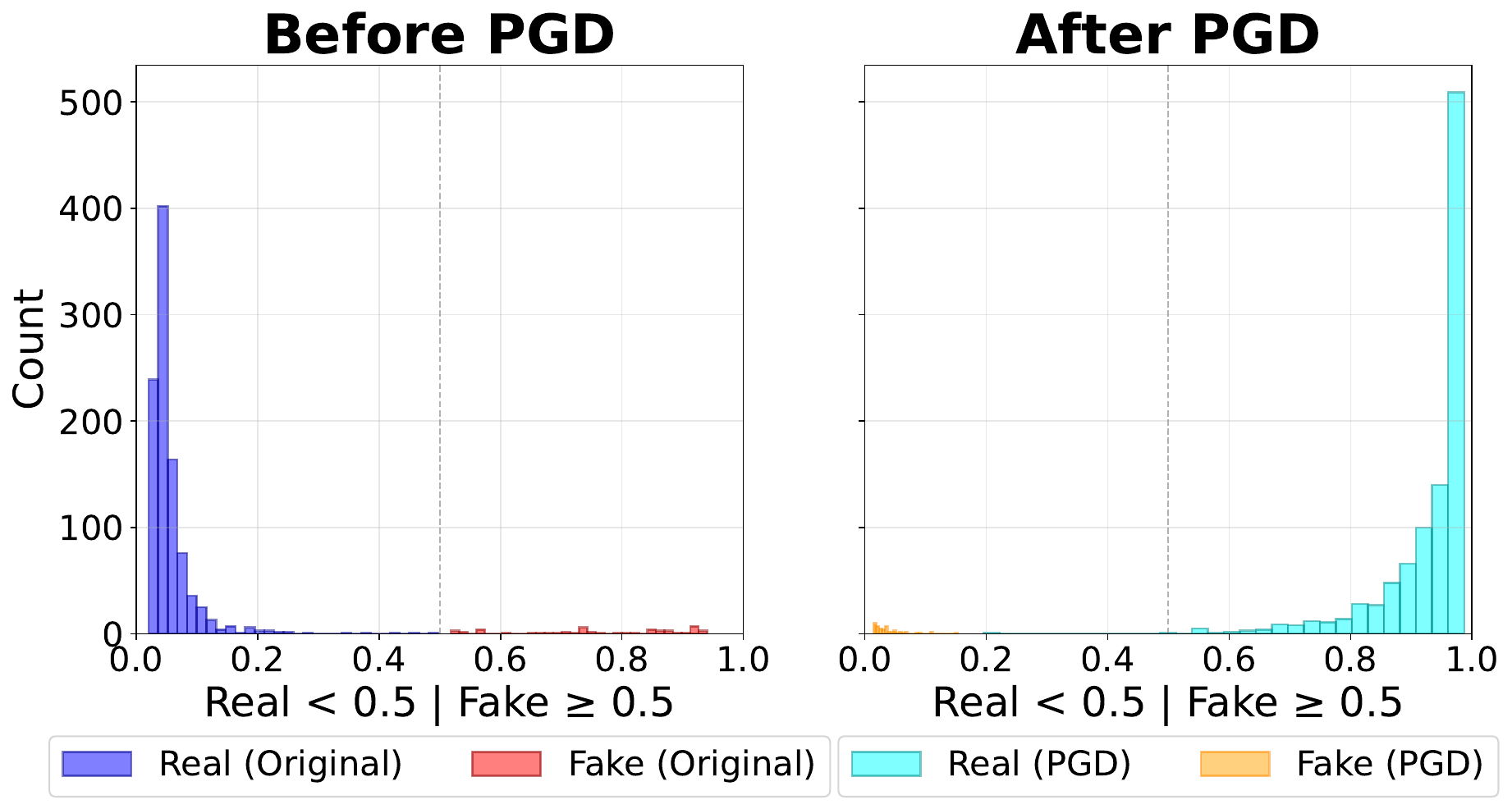}
          \caption{C2P-CLIP}
      \end{subfigure}

      \vspace{0.05cm}

      \begin{subfigure}[b]{0.30\textwidth}
          \includegraphics[width=\linewidth]{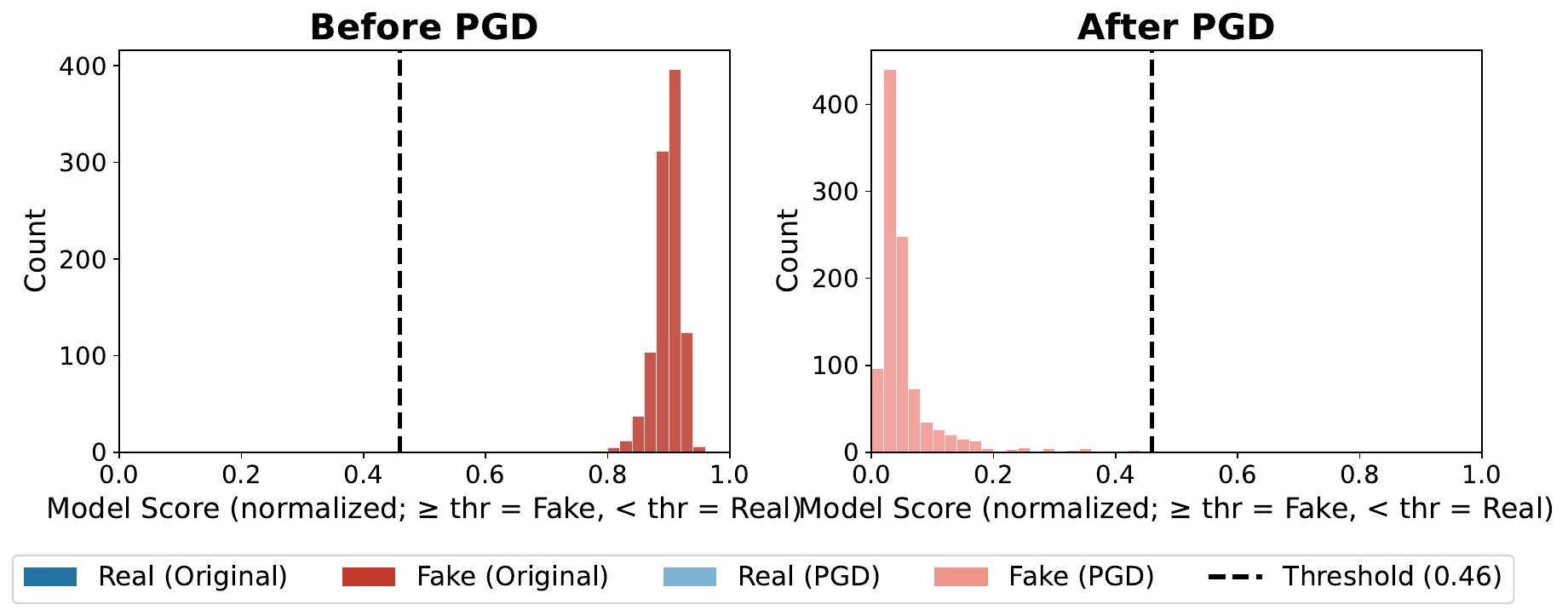}
          \caption{FIRE}
      \end{subfigure}
      \hfill
      \begin{subfigure}[b]{0.30\textwidth}
          \includegraphics[width=\linewidth]{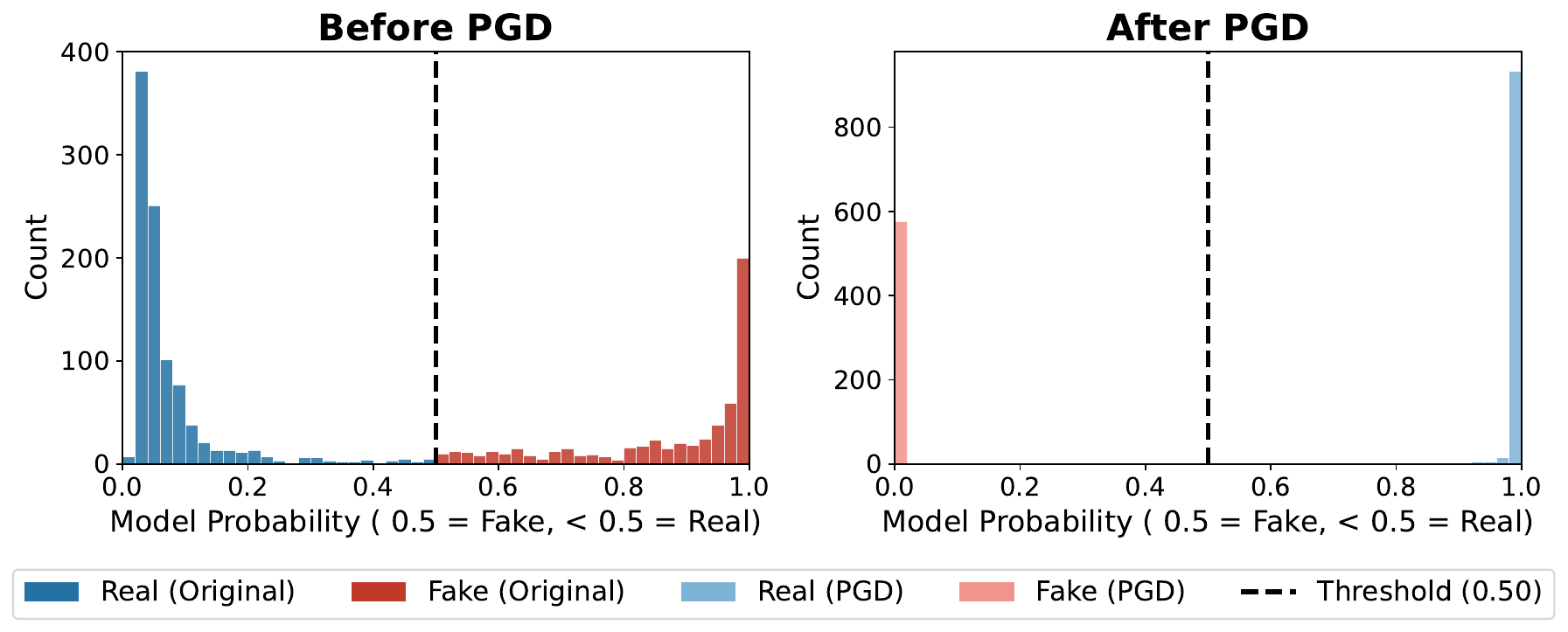}
          \caption{DDA}
      \end{subfigure}
      \hfill
      \begin{subfigure}[b]{0.30\textwidth}
          \includegraphics[width=\linewidth]{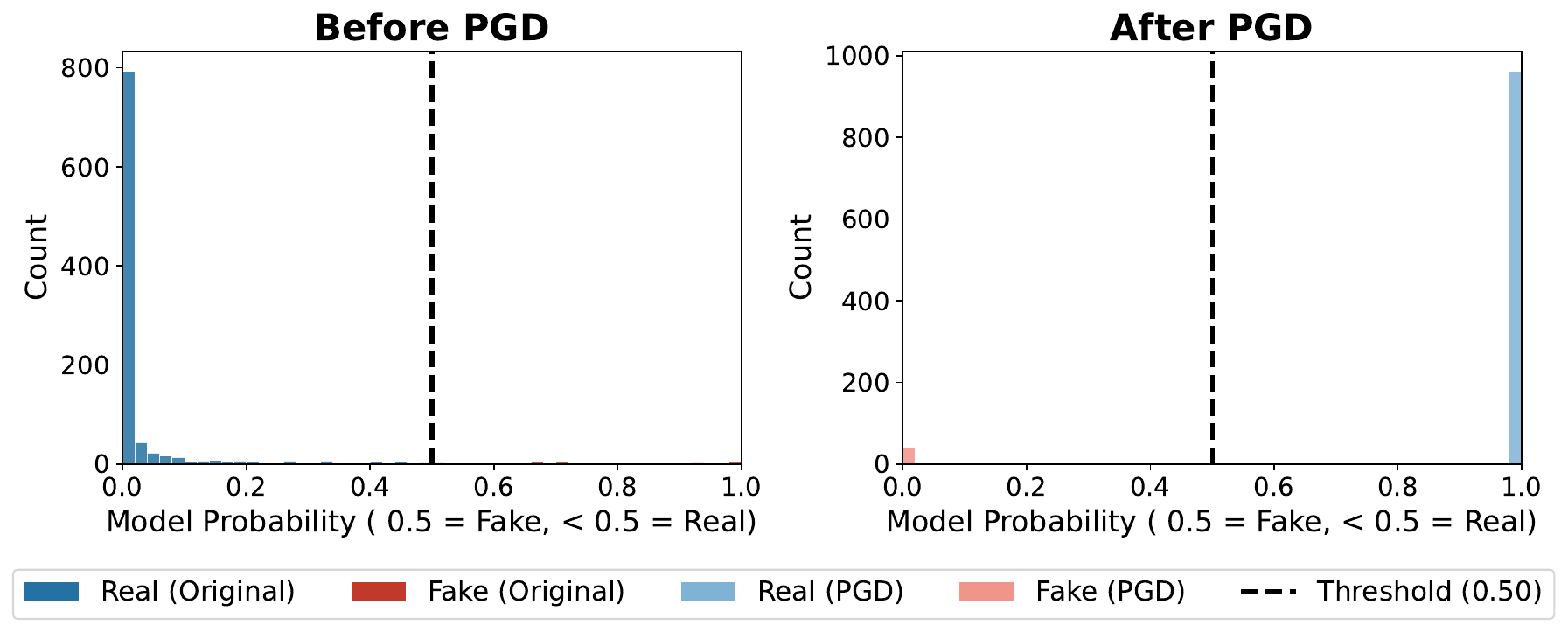}
          \caption{FerretNet}
      \end{subfigure}

      \vspace{0.05cm}

      \begin{subfigure}[b]{0.30\textwidth}
          \includegraphics[width=\linewidth]{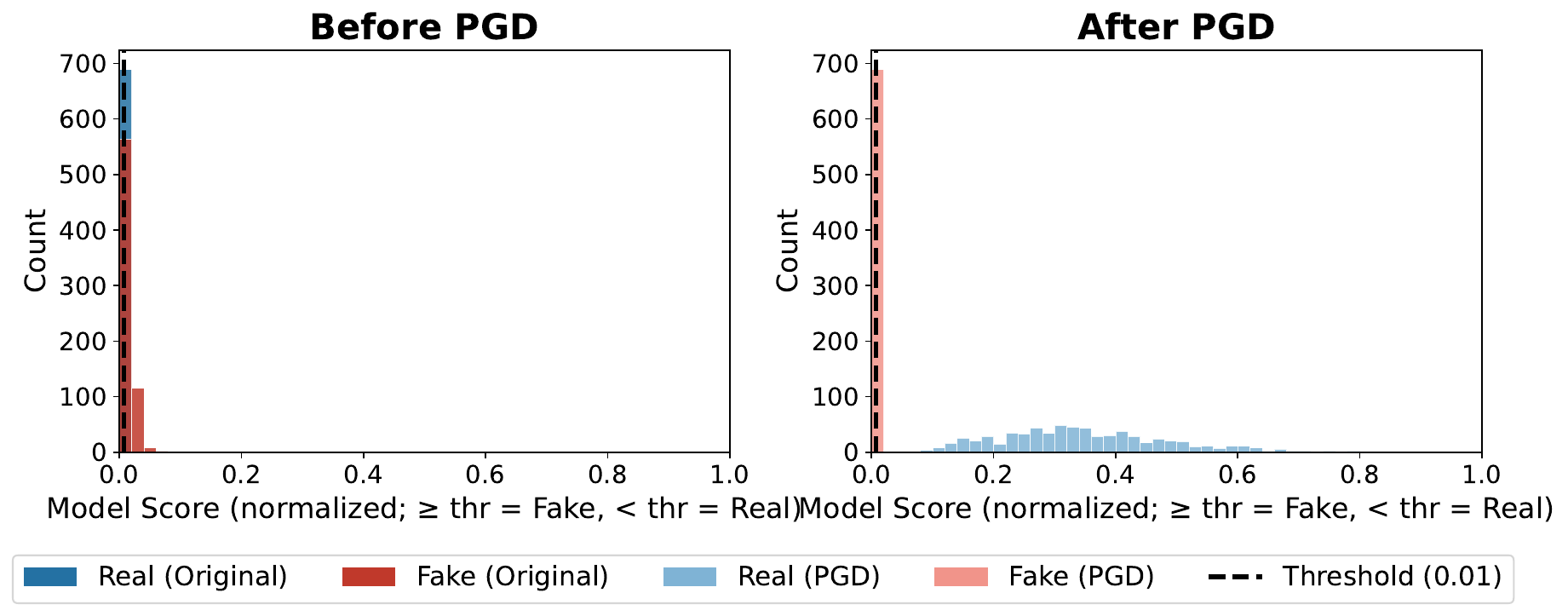}
          \caption{WaRPAD}
      \end{subfigure}
      \hfill
      \begin{subfigure}[b]{0.30\textwidth}
          \includegraphics[width=\linewidth]{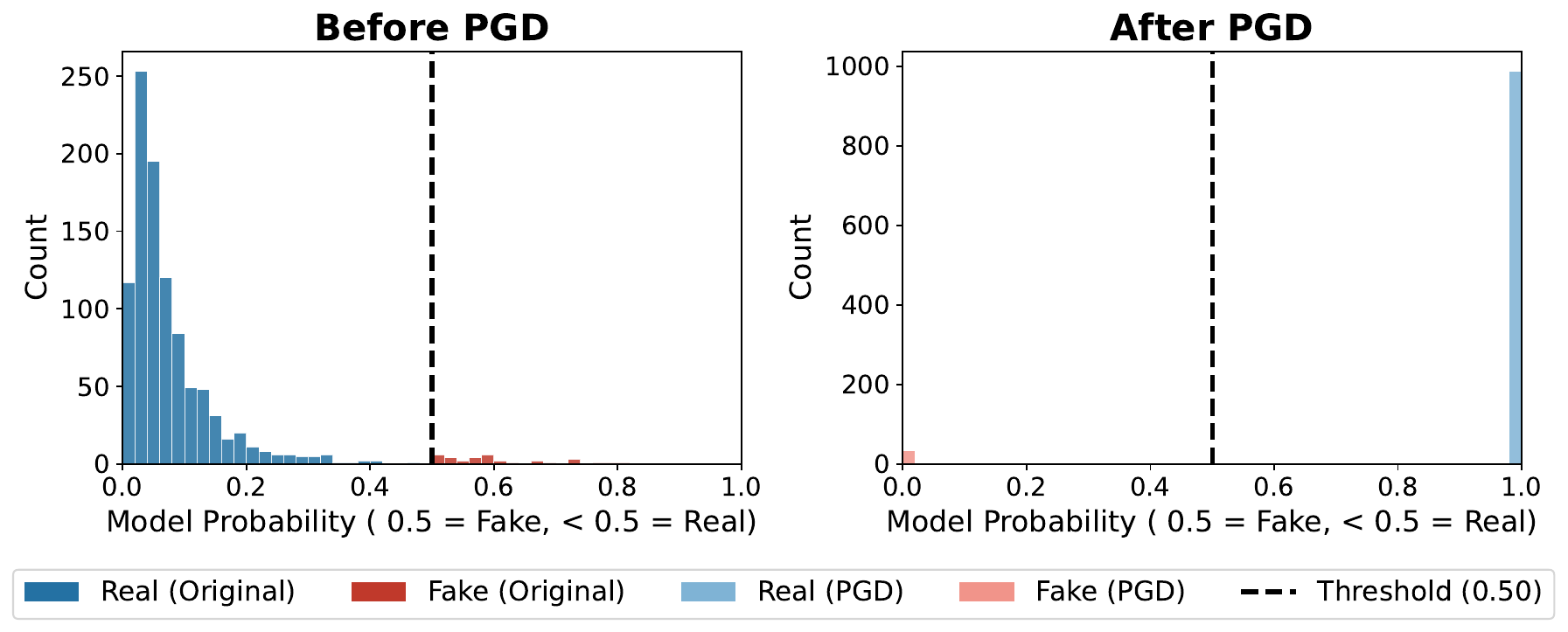}
          \caption{AllPatchesMatter}
      \end{subfigure}
      \hfill
      \begin{subfigure}[b]{0.30\textwidth}
          \includegraphics[width=\linewidth]{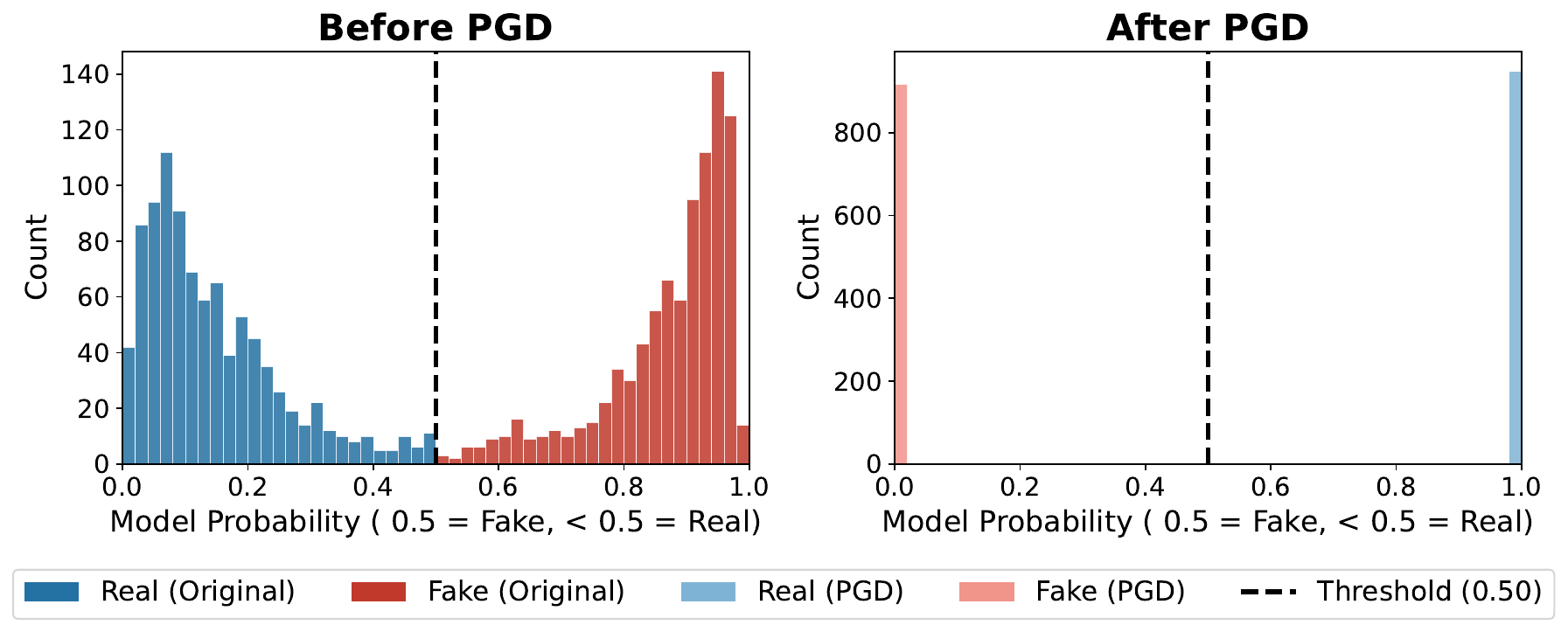}
          \caption{OmniAID}
      \end{subfigure}

      \vspace{0.05cm}

      \begin{subfigure}[b]{0.30\textwidth}
          \includegraphics[width=\linewidth]{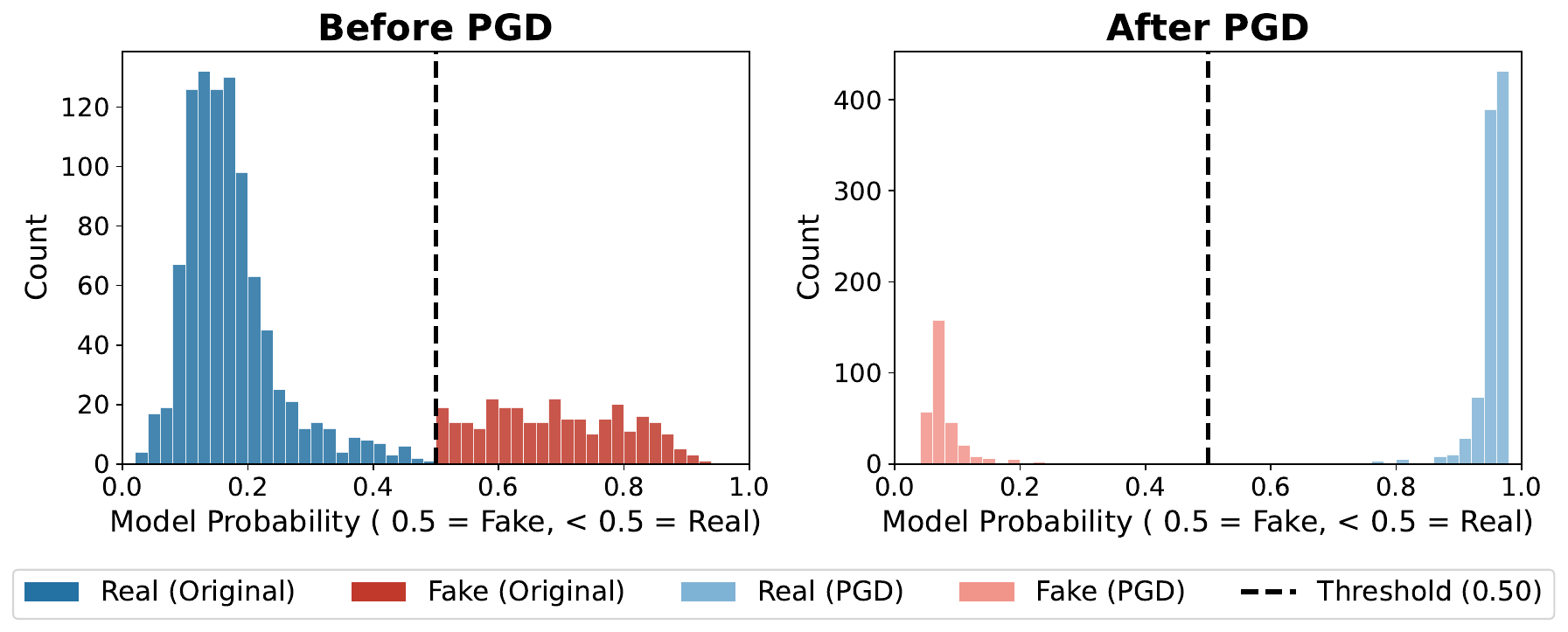}
          \caption{PGC}
      \end{subfigure}
      \hfill
      \begin{subfigure}[b]{0.30\textwidth}
          \includegraphics[width=\linewidth]{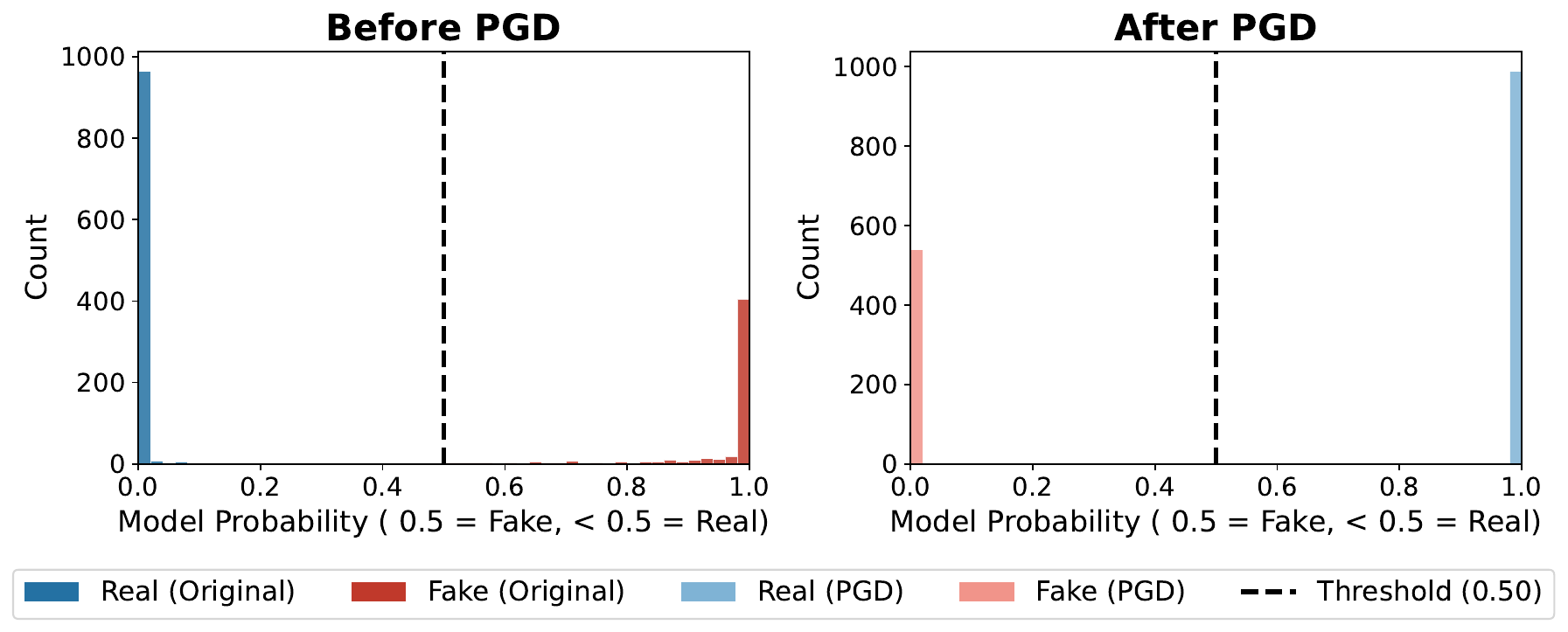}
          \caption{PROBE}
      \end{subfigure}
      \hfill
      \begin{subfigure}[b]{0.30\textwidth}
          \includegraphics[width=\linewidth]{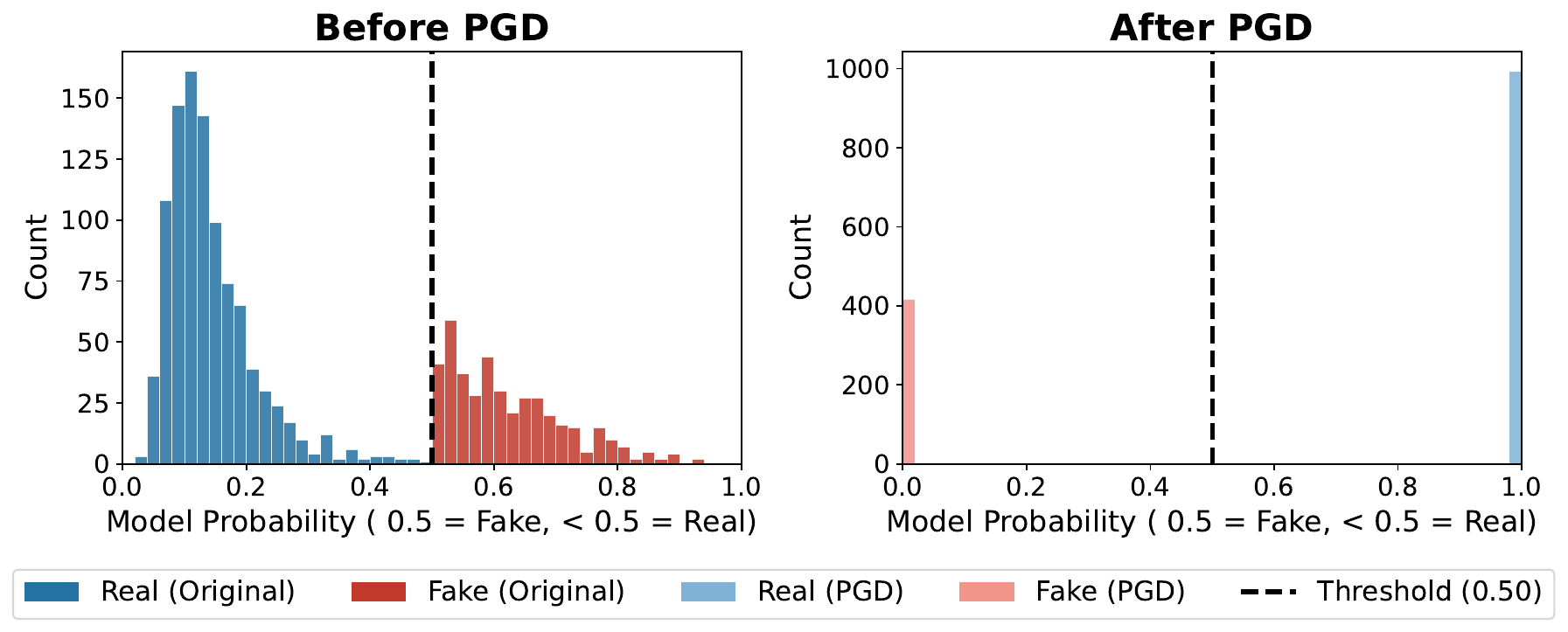}
          \caption{DEAR}
      \end{subfigure}

      \vspace{0.05cm}

      \begin{subfigure}[b]{0.30\textwidth}
          \includegraphics[width=\linewidth]{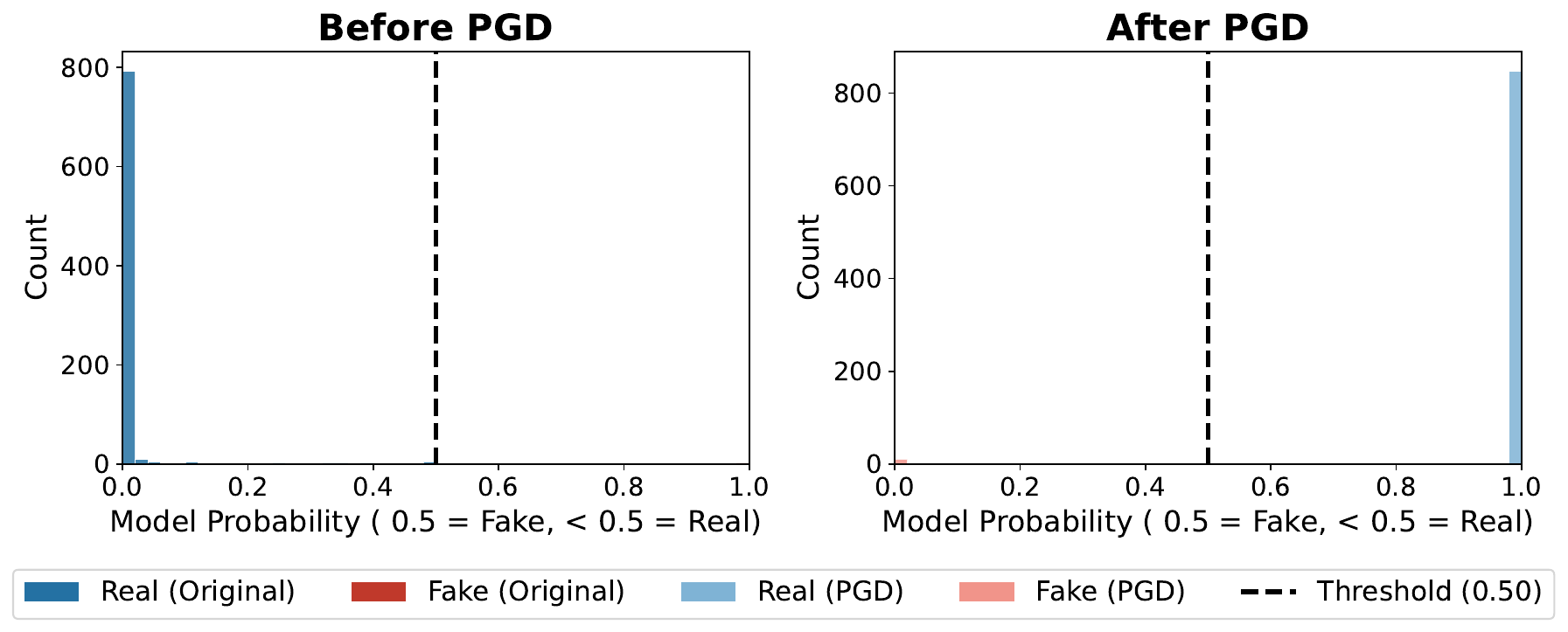}
          \caption{DGS-Net}
      \end{subfigure}
      \hfill
      \begin{subfigure}[b]{0.30\textwidth}
          \includegraphics[width=\linewidth]{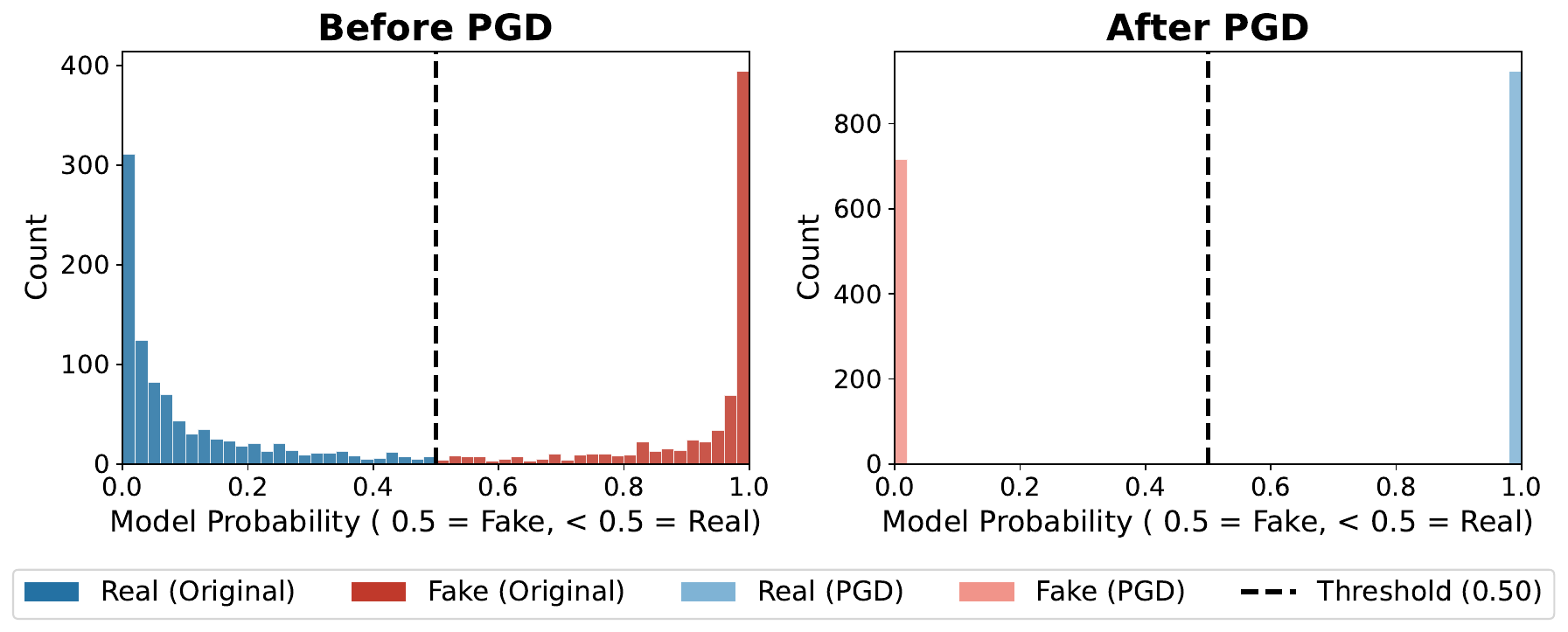}
          \caption{SICA}
      \end{subfigure}
      \hfill
      \begin{subfigure}[b]{0.30\textwidth}
          \includegraphics[width=\linewidth]{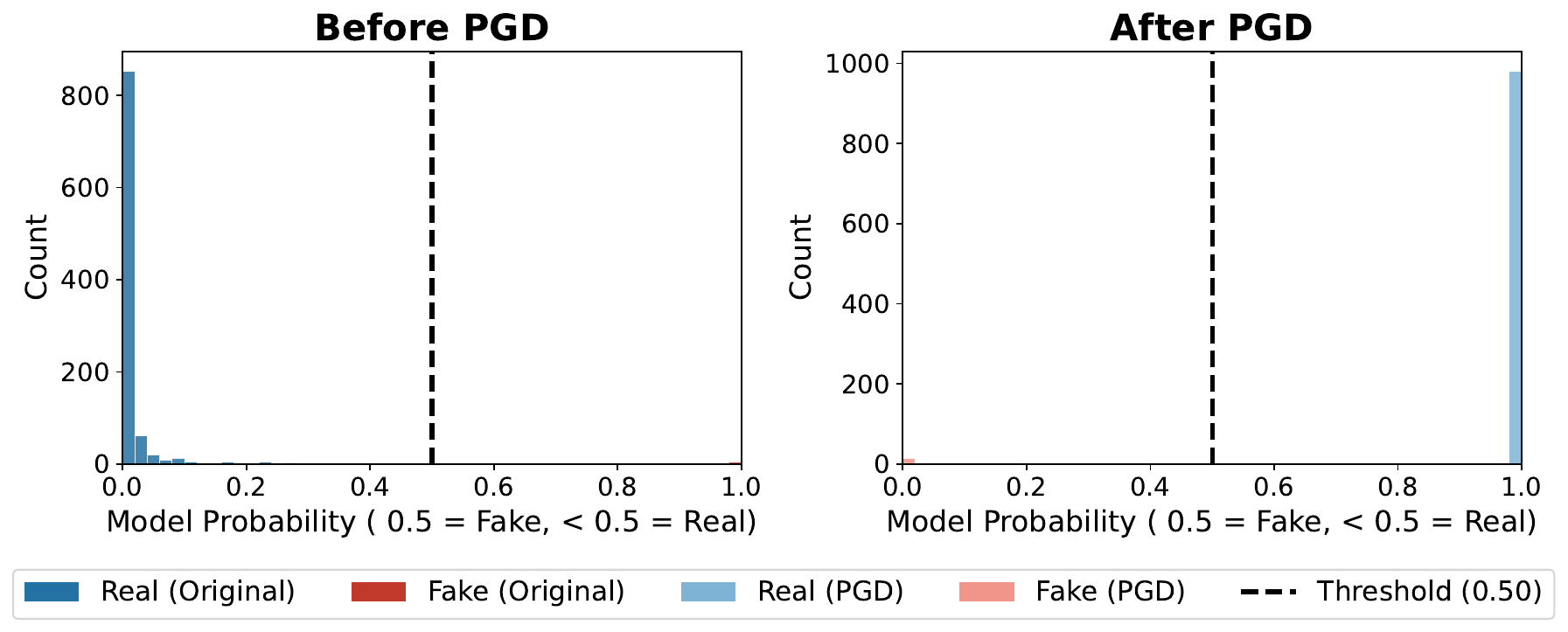}
          \caption{IAPL}
      \end{subfigure}

      \vspace{0.05cm}

      \begin{subfigure}[b]{0.30\textwidth}
          \includegraphics[width=\linewidth]{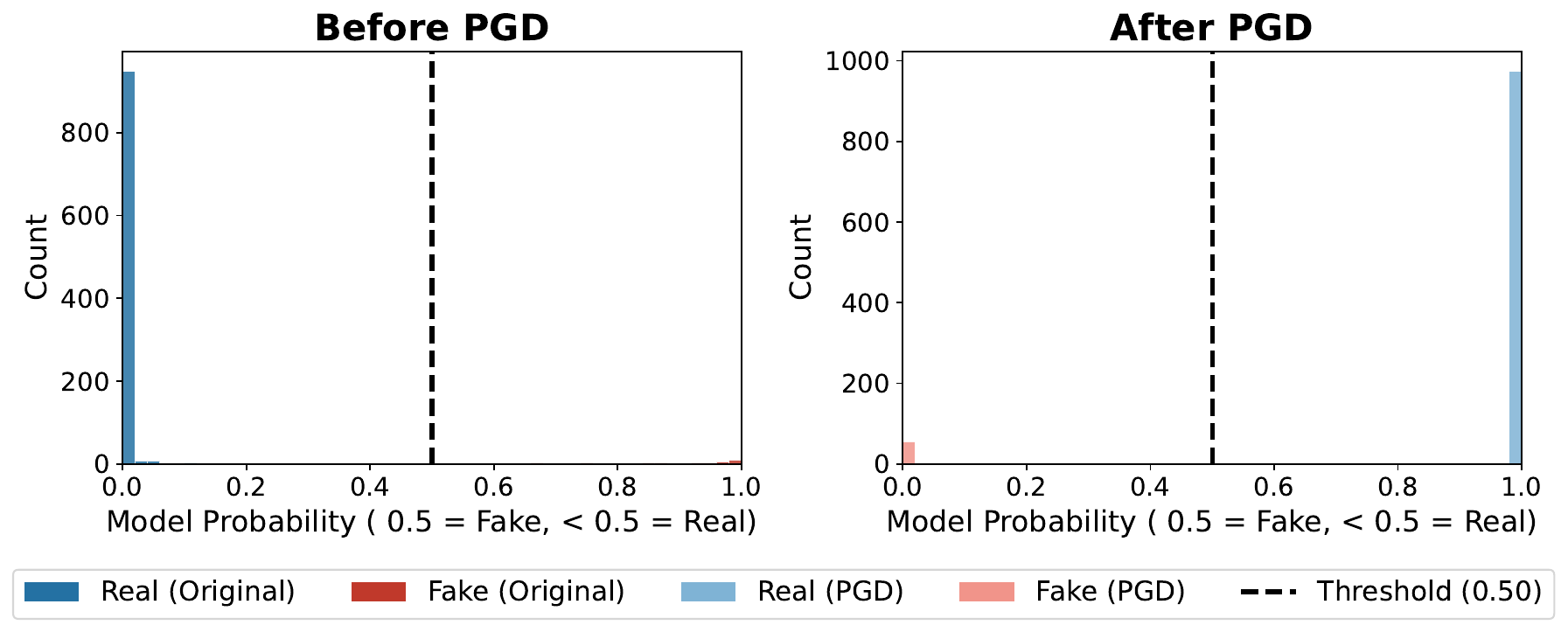}
          \caption{ForensicConcept}
      \end{subfigure}
      \caption{Prediction score distributions before and after PGD attack for each evaluated detector
  (excluding D3, shown separately in \Cref{fig:d3_pgd}). Each plot shows how adversarial perturbation
  erodes separability between real and fake inputs, often collapsing confidence entirely.}
      \label{fig:pgd_attacks}
  \end{figure*}

The results show a complete collapse in performance across all methods. Models like UFD~\cite{ufd}, FreqNet~\cite{freqnet}, and NPR~\cite{npr} fail entirely, with prediction distributions collapsing into indistinguishability between real and fake classes. FatFormer~\cite{fatformer}
and C2P-CLIP~\cite{c2pclip} also misclassify nearly all samples post-attack. These visualizations in Figure~\ref{fig:pgd_attacks} highlight the fragility of binary detectors and the lack of graceful degradation under adversarial pressure (a phenomenon that is particularly pronounced in the cases of NPR and FatFormer). In contrast, our proposed method maintains calibrated separation and abstains when confidence is insufficient, as demonstrated in the main paper.

To illustrate the imperceptibility of the attack, \Cref{fig:d3_pgd} shows an image before and after PGD perturbation. Despite the negligible visual difference, the prediction shifts significantly post-attack, causing traditional detectors to fail.

\section{Medium Resource Attacker Full Details}
\label{sec:medium-attacker-full-details}

To operationalize our medium-resource attacker scenario , we simulate a threat model where the adversary is granted access to a single prompt and a fixed computational budget. The attacker samples \(N = 100\) images by drawing random seeds and generating samples \(x_i = G_\theta(z_i;\text{prompt})\) for \(i = 1,\dots,100\), where \(G_\theta\) is the target generative model (Stable Diffusion 3 medium in our case). Each generated image is inverted using our reconstruction-free inversion pipeline \(\widetilde{G}^{-1}\) and scored using the Authenticity Index:
\begin{equation}
\text{A-index}(x_i, \widetilde{x}_i) = \text{A-index}(x_i, \widetilde{G}^{-1}(x_i)).
\end{equation}

The attacker then selects the highest-scoring candidate \(x^\star\) and optimizes a small perturbation \(\delta\), constrained by \(\|\delta\|_\infty \leq 8/255\), to maximize the A-index after inversion of the perturbed image.

The prompt used in this experiment is:
\begin{quote}
\myverbatim{A hungry man standing outside a real pizza shop at night, mouth slightly open, drooling, pointing toward the glowing neon pizza sign. Warm light from the shop window reveals fresh cheesy pizzas inside with steam, the scene looks realistic and cinematic with natural lighting and lifelike details.}
\label{quote:prompt_hungry_man_real_pizza}
\end{quote}

\begin{figure*}[!t]
    \centering
    \begin{minipage}{0.82\textwidth}
        \centering
        \includegraphics[width=\linewidth]{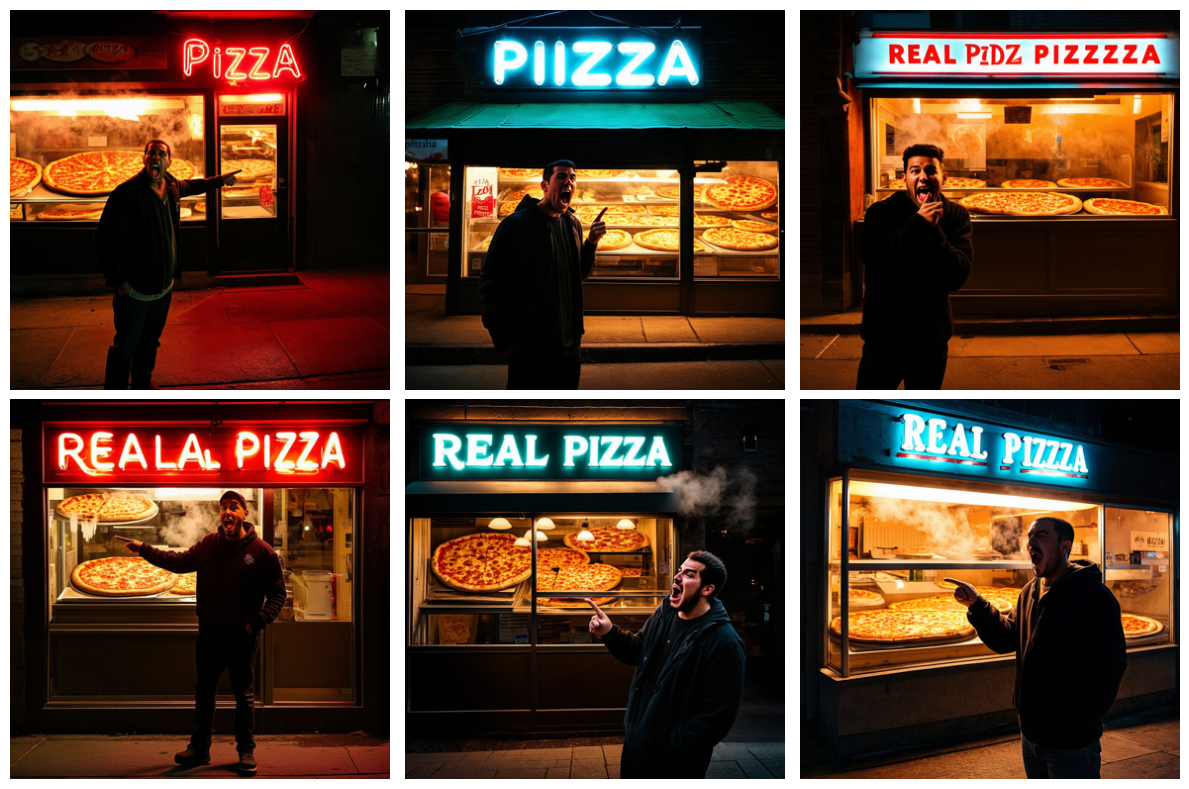}
    \end{minipage}
    \caption{Six image samples generated from the same prompt (\Cref{quote:prompt_hungry_man_real_pizza}), with different random seeds. All were generated under the medium-resource attacker's sampling budget. These highlight the diversity in candidate outputs that the attacker can choose from.}
    \label{fig:medium_attacker_variants}
\end{figure*}

To illustrate the range of outputs available under the attacker's sampling budget, we visualize six randomly sampled images generated from this prompt using different seeds in Figure~\ref{fig:medium_attacker_variants}. These images are unmodified generations prior to inversion or adversarial refinement. While visually diverse, the top-scoring image among the 100 candidates achieved an \(\text{A-index}\) of 0.0148. After applying PGD to perturb \(x^\star\), the \(\text{A-index}(x^\star, \widetilde{G}^{-1}(x^\star + \delta))\) increased modestly to 0.0154. This slight improvement remains below both the safety threshold (\(\tau_\text{safety} = 0.0365\)) and the adversarial security threshold (\(\tau_\text{security} = 0.038\)), indicating the limited effectiveness of a medium-resource attacker.

\section{Authenticity Index (A-index): Design, Ablations, and Sensitivity}
\label{sec:supp_aindex}

This section provides additional analysis of the Authenticity Index (A-index), including metric statistics under inversion, learned weights, subset ablations over all metric combinations, robustness to weight perturbations, and comparisons to alternative aggregation strategies. 

\subsection{Metric Statistics Under Inversion}
\label{sec:supp_metric_stats}

Table~\ref{tab:supp_metric_stats} summarizes the inversion similarity statistics for real and synthetic images. Across all four metrics, synthetic images are reconstructed more faithfully on average, consistent with the interpretation that ``resynthesizability'' under inversion is a measurable property that differs between authentic and generated content.

\begin{table}[H]
\centering
\caption{Inversion metric statistics on the calibration split. Synthetic images reconstruct more faithfully on average, consistent with the hypothesis that generated images are easier to re-synthesize under inversion.}
\label{tab:supp_metric_stats}
\small
\setlength{\tabcolsep}{6pt}
\begin{tabular}{lccc}
\toprule
\textbf{Metric} & \textbf{Fake mean} & \textbf{Real mean} & \textbf{Fake $-$ Real} \\
\midrule
PSNR      & 23.6922 & 22.0028 & +1.6894 \\
SSIM      & 0.7207  & 0.6036  & +0.1171 \\
$1-\mathrm{LPIPS}$ & 0.9240  & 0.9126  & +0.0114 \\
CLIP similarity & 0.9666  & 0.9293  & +0.0373 \\
\bottomrule
\end{tabular}

\end{table}

\subsection{A-index Definition and Learned Weights}
\label{sec:supp_weights}

We form a weighted combination of the four inversion metrics and map it through a sigmoid. The learned weights (optimized to minimize distribution overlap on the calibration split) and the sigmoid scale used throughout experiments are:

\begin{equation}
\begin{aligned}
x ={}& -0.0181\cdot \mathrm{PSNR}
+ 1.380\cdot \mathrm{SSIM} \\
&- 4.058\cdot (1-\mathrm{LPIPS})
+ 8.066\cdot \mathrm{CLIP},
\qquad \sigma = 0.9.
\end{aligned}
\end{equation}
A positive weight means higher similarity increases the score, while a negative weight means higher similarity decreases it. Notably, $(1-\mathrm{LPIPS})$ receives a negative weight: perceptual reconstruction can be disproportionately faithful for synthetic images under inversion, making high perceptual similarity indicative of synthetic origin in this calibration distribution.

\subsection{Single-Metric Performance}
\label{sec:supp_single_metric}

Table~\ref{tab:supp_single_metric} reports separability when each inversion metric is used alone. CLIP similarity provides the strongest individual signal ($\mathrm{AUC}_{\mathrm{eff}}=0.8546$), suggesting that semantic alignment between the image and its inversion is highly informative. SSIM also offers meaningful separation, indicating that structural fidelity under inversion differs between real and synthetic images.

In contrast, PSNR and $(1-\mathrm{LPIPS})$ perform substantially worse as standalone predictors. Pixel-level reconstruction error alone does not reliably distinguish image origin, and perceptual similarity can remain high even for generated images. These results motivate combining complementary metrics in the A-index rather than relying on any single similarity measure.
\begin{table}[t]
\centering
\caption{Single-metric separability under inversion (weights optimized independently for each metric). CLIP similarity provides the strongest standalone signal.}
\label{tab:supp_single_metric}
\small
\setlength{\tabcolsep}{6pt}
\begin{tabular}{lccc}
\toprule
\textbf{Metric (alone)} & \textbf{$\mathrm{AUC}_{\mathrm{eff}} \uparrow$} & \textbf{Overlap $\downarrow$} & \textbf{Cohen's $d \uparrow$} \\
\midrule
PSNR & 0.6515 & 0.6942 & 0.4896 \\
SSIM & 0.7284 & 0.6040 & 0.8630 \\
$(1-\mathrm{LPIPS})$ & 0.5967 & 0.8062 & 0.3378 \\
CLIP similarity & \textbf{0.8546} & \textbf{0.4401} & \textbf{1.2996} \\
\bottomrule
\end{tabular}

\end{table}

\subsection{Leave-One-Out Ablations}
\label{sec:supp_loo}

Table~\ref{tab:supp_loo} evaluates the contribution of each metric by removing it from the full A-index and re-optimizing the remaining weights. Removing CLIP causes the largest degradation, increasing overlap from $0.3436$ to $0.5284$ and reducing Cohen's $d$ substantially. This confirms that semantic similarity is the most informative component.

Removing SSIM or $(1-\mathrm{LPIPS})$ leads to moderate performance drops, indicating that structural and perceptual cues provide complementary information. In contrast, removing PSNR has minimal impact, consistent with its near-zero weight in the optimized model.

\begin{table*}[t]
\centering
\caption{Leave-one-out ablation of A-index metrics. Removing CLIP causes the largest degradation in separability, while removing PSNR has minimal impact.}
\label{tab:supp_loo}
\begin{tabular}{lcccc}
\toprule
\textbf{Removed metric} &
\textbf{$\mathrm{AUC}_{\mathrm{eff}} \uparrow$} &
\textbf{Overlap $\downarrow$} &
\textbf{Cohen's $d \uparrow$} &
\textbf{$\Delta$Overlap / $\Delta d$} \\
\midrule
Remove PSNR & 0.8841 & 0.3476 & 1.5690 & +0.0040 / $-$0.0180 \\
Remove SSIM & 0.8609 & 0.3936 & 1.3956 & +0.0500 / $-$0.1913 \\
Remove $(1-\mathrm{LPIPS})$ & 0.8623 & 0.4035 & 1.4163 & +0.0599 / $-$0.1706 \\
Remove CLIP & 0.7693 & 0.5284 & 0.9619 & +0.1848 / $-$0.6251 \\
\midrule
\textbf{None (All 4)} & \textbf{0.8852} & \textbf{0.3436} & \textbf{1.5870} & baseline \\
\bottomrule
\end{tabular}

\end{table*}

\subsection{Exhaustive Subset Sweep Over All 15 Non-Empty Metric Subsets}
\label{sec:supp_all_subsets}

Tables~\ref{tab:supp_best_by_size} and~\ref{tab:supp_all_15_subsets} report results for all metric subsets. Performance improves consistently as additional complementary metrics are included. While CLIP alone provides the strongest single-metric signal, combining it with structural and perceptual metrics substantially reduces distribution overlap.

The best three-metric subset (SSIM + $(1-\mathrm{LPIPS})$ + CLIP) nearly matches the full model, while the complete four-metric formulation achieves the best overall separability (overlap $0.3436$, $d=1.5870$). These results confirm that aggregating multiple inversion signals improves discrimination compared to any individual metric.

\begin{table*}[t]
\centering
\caption{Best-performing subset at each subset size from the exhaustive metric sweep. The full four-metric combination achieves the strongest separability.}
\label{tab:supp_best_by_size}
\begin{tabular}{lccc}
\toprule
\textbf{\# Metrics} & \textbf{Best subset (by overlap)} & \textbf{Overlap $\downarrow$} & \textbf{Cohen's $d \uparrow$} \\
\midrule
1 & CLIP & 0.4401 & 1.2996 \\
2 & SSIM + CLIP & 0.4028 & 1.3596 \\
3 & SSIM + $(1-\mathrm{LPIPS})$ + CLIP & 0.3476 & 1.5690 \\
\textbf{4} & \textbf{PSNR + SSIM + $(1-\mathrm{LPIPS})$ + CLIP} & \textbf{0.3436} & \textbf{1.5870} \\
\bottomrule
\end{tabular}

\end{table*}

\begin{table}[t]
\centering
\caption{Performance of all 15 non-empty metric subsets with independently optimized weights. Combining complementary inversion signals consistently improves separability, with the full four-metric combination achieving the strongest performance.}
\label{tab:supp_all_15_subsets}
\scriptsize
\setlength{\tabcolsep}{5pt}
\begin{tabular}{lccc}
\toprule
\textbf{Subset} & \textbf{$\mathrm{AUC}_{\mathrm{eff}} \uparrow$} & \textbf{Overlap $\downarrow$} & \textbf{Cohen's $d \uparrow$} \\
\midrule
\multicolumn{4}{c}{\textit{Single metrics}} \\
PSNR & 0.6515 & 0.6942 & 0.4896 \\
SSIM & 0.7284 & 0.6040 & 0.8630 \\
$(1-\mathrm{LPIPS})$ & 0.5967 & 0.8062 & 0.3378 \\
CLIP & 0.8546 & 0.4401 & 1.2996 \\
\midrule
\multicolumn{4}{c}{\textit{Two metrics}} \\
PSNR + SSIM & 0.7405 & 0.5818 & 0.9099 \\
PSNR + $(1-\mathrm{LPIPS})$ & 0.6512 & 0.6806 & 0.4824 \\
PSNR + CLIP & 0.8576 & 0.4146 & 1.3528 \\
SSIM + $(1-\mathrm{LPIPS})$ & 0.7434 & 0.5704 & 0.9016 \\
SSIM + CLIP & 0.8611 & 0.4028 & 1.3596 \\
$(1-\mathrm{LPIPS})$ + CLIP & 0.8581 & 0.4163 & 1.3285 \\
\midrule
\multicolumn{4}{c}{\textit{Three metrics}} \\
PSNR + SSIM + $(1-\mathrm{LPIPS})$ & 0.7693 & 0.5284 & 0.9619 \\
PSNR + SSIM + CLIP & 0.8623 & 0.4035 & 1.4163 \\
PSNR + $(1-\mathrm{LPIPS})$ + CLIP & 0.8609 & 0.3936 & 1.3956 \\
SSIM + $(1-\mathrm{LPIPS})$ + CLIP & 0.8841 & 0.3476 & 1.5690 \\
\midrule
\multicolumn{4}{c}{\textit{Four metrics}} \\
\textbf{PSNR + SSIM + $(1-\mathrm{LPIPS})$ + CLIP} & \textbf{0.8852} & \textbf{0.3436} & \textbf{1.5870} \\
\bottomrule
\end{tabular}

\end{table}

\subsection{Sensitivity to Weight Perturbations}
\label{sec:supp_weight_sensitivity}

We evaluate robustness of the learned weights by applying multiplicative Gaussian perturbations proportional to each weight magnitude (100 trials per noise level). Table~\ref{tab:weight_sensitivity} shows that separability degrades smoothly under moderate perturbations, with substantial degradation only under extreme noise.

\begin{table*}[t]
\centering
\caption{Sensitivity of separability to perturbations of the learned A-index weights. Performance degrades gradually under moderate perturbations, indicating robustness of the learned weighting scheme.}
\label{tab:weight_sensitivity}
\setlength{\tabcolsep}{6pt}
\renewcommand{\arraystretch}{1.05}
\begin{tabular}{lccc}
\hline
\textbf{Perturbation level} &
\textbf{Overlap (mean$\pm$std) $\downarrow$} &
\textbf{Cohen's $d$ (mean$\pm$std) $\uparrow$} &
\textbf{$\mathrm{AUC}_{\mathrm{eff}}$ $\uparrow$} \\
\hline
0\%   & $0.3436 \pm 0.0000$ & $1.5870 \pm 0.0000$ & 0.8852 \\
10\%  & $0.3781 \pm 0.0090$ & $1.5834 \pm 0.0172$ & 0.8843 \\
20\%  & $0.3830 \pm 0.0094$ & $1.5756 \pm 0.0302$ & 0.8821 \\
30\%  & $0.3917 \pm 0.0199$ & $1.5480 \pm 0.0660$ & 0.8772 \\
50\%  & $0.4380 \pm 0.0784$ & $1.4083 \pm 0.2395$ & 0.8495 \\
70\%  & $0.4932 \pm 0.1365$ & $1.2415 \pm 0.4086$ & 0.8129 \\
100\% & $0.5021 \pm 0.1303$ & $1.2280 \pm 0.4017$ & 0.8077 \\
\hline
\end{tabular}

\end{table*}

\subsection{Alternative Aggregation Strategies}
\label{sec:supp_alt_aggs}

We compare the linear+sigmoid A-index to a range of alternative aggregation approaches, including simple baselines, an unsupervised projection (PCA), and supervised models evaluated via 5-fold stratified cross-validation. Table~\ref{tab:alt_aggs} reports $\mathrm{AUC}_{\mathrm{eff}}$, overlap, Cohen's $d$, parameter counts, and cross-validation variability.
\begin{table*}[t]
\centering
\caption{Comparison to alternative aggregation strategies sorted by $\mathrm{AUC}_{\mathrm{eff}}$. Our method (\textbf{Linear + sigmoid}) achieves competitive performance while remaining interpretable and using only four parameters. Supervised models are evaluated via 5-fold cross-validation.}
\label{tab:alt_aggs}
\setlength{\tabcolsep}{5pt}
\renewcommand{\arraystretch}{1.05}
\begin{tabular}{lcccccc}
\hline
\textbf{Method} &
\textbf{$\mathrm{AUC}_{\mathrm{eff}}$ $\uparrow$} &
\textbf{Overlap $\downarrow$} &
\textbf{Cohen's $d$ $\uparrow$} &
\textbf{\#Params} &
\textbf{Interp.} &
\textbf{CV std (Overlap)} \\
\hline
SVM-RBF (5-fold)        & 0.9125 & 0.3003 & 2.0610 & $\mathcal{O}(n)$ & N & 0.0323 \\
Random forest (5-fold)  & 0.8948 & 0.3470 & 1.9023 & $>$1000 & N & 0.0504 \\
\textbf{Linear + sigmoid (ours)} & \textbf{0.8852} & \textbf{0.3436} & \textbf{1.5870} & \textbf{4} & \textbf{Y} & -- \\
Logistic reg. (5-fold)  & 0.8848 & 0.3629 & 1.7271 & 5 & Y & 0.0425 \\
Product (log-space)     & 0.8777 & 0.3489 & 1.5437 & 4 & Y & -- \\
MLP (16,8) (5-fold)     & 0.8518 & 0.4537 & 1.5169 & 225 & N & 0.0742 \\
PCA (PC1) + sigmoid     & 0.7328 & 0.5999 & 0.7467 & 4 & N & -- \\
Geometric mean + sigmoid& 0.7219 & 0.6162 & 0.8354 & 1 & Y & -- \\
Equal weights + sigmoid & 0.6587 & 0.7359 & 0.5602 & 0 & Y & -- \\
\hline
\end{tabular}

\end{table*}

\subsection{Summary.}
Across all 15 metric subsets, combining complementary inversion signals improves separability, with the full four-metric A-index achieving the lowest overlap (0.3436) and highest Cohen's $d$ (1.5870). Leave-one-out ablations show CLIP contributes the largest marginal gain, while PSNR has minimal impact (consistent with its near-zero learned weight). Weight perturbation experiments indicate smooth degradation under moderate noise, and comparisons to alternative aggregations show the linear+sigmoid formulation performs competitively while remaining compact and interpretable.

\section{Reddit Collection Methodology and Bias}
\label{app:reddit}
\label{sec:reddit_methodology_bias}
\subsection{Collection Methodology}
The motivation for this study is to understand how vulnerable real-world imagery is to resynthesis as generative models become increasingly powerful. If common internet images can be inverted with high fidelity, then their authenticity becomes difficult to establish, since an adversary can plausibly claim they were generated. To examine this risk in practice, we collected approximately 3{,}000 images from Reddit using keyword-driven queries spanning multiple content categories. The keyword sets were organized into the following groups: \texttt{ai\_art}, \texttt{clickbait\_thumbnails}, \texttt{conspiracy\_imagery}, \texttt{fake\_product\_ads}, \texttt{hateful\_memes}, \texttt{misinformation\_graphics}, \texttt{people\_faces}, \texttt{propaganda\_images}, and \texttt{protest\_and\_activism}. We captioned the collected images using Salesforce BLIP-2~\cite{blip2}. We then inverted the images using Stable Diffusion 3 (medium). Separately, we calibrate a safety threshold for each generative method (Stable Diffusion 2.1, SD3 (medium), SD3.5 (medium), Flux Dev, and Flux Dev with Realism LoRA) using distributions of real-inverted and fake-inverted images. We then apply the method-specific thresholds to the same set of SD3-inverted internet images to assess how the certifiable subset changes across generative methods.
\begin{figure*}[!t]
    \centering
    \includegraphics[width=0.9\linewidth]{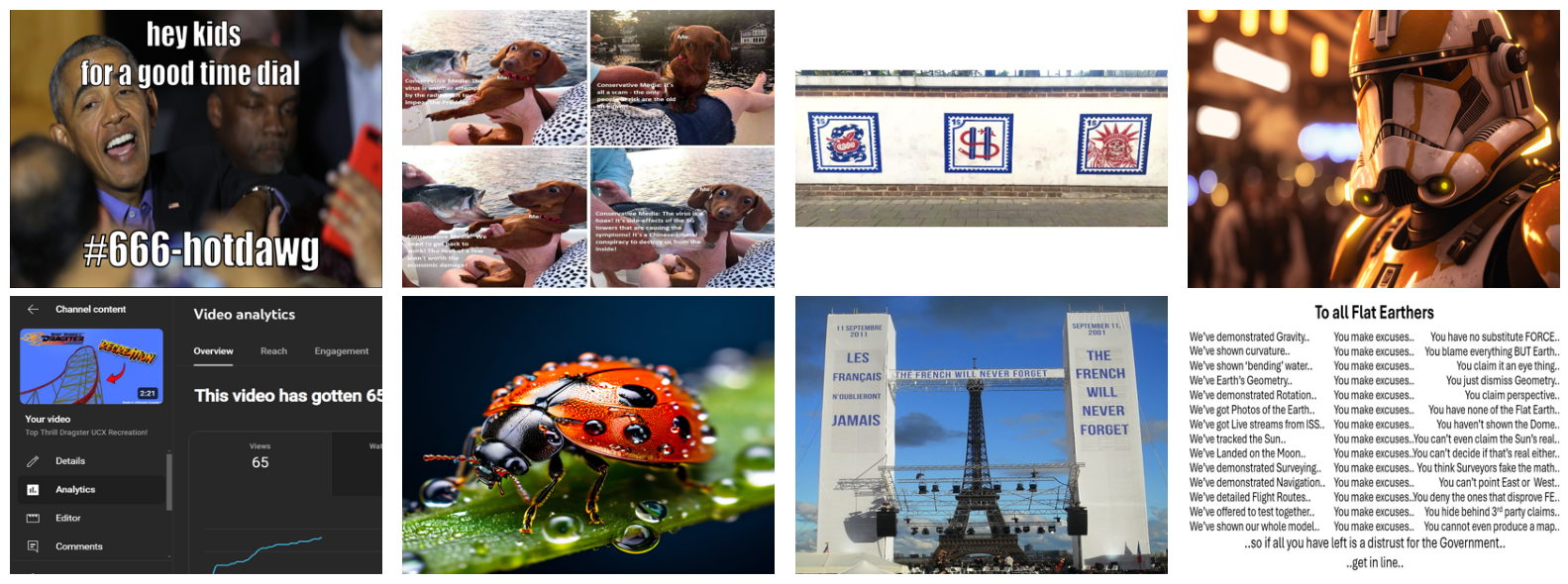}
    \caption{\textbf{Examples from the keyword-sampled Reddit corpus.} The corpus contains photographs, screenshots, memes, and digital artwork.}
    \label{fig:reddit_examples}
\end{figure*}

\subsection{Bias and Limitations}
This collection is keyword-driven and therefore not a representative sample of Reddit or of internet imagery. The resulting dataset may over-sample content that matches the chosen keyword groups and under-sample content that does not, and it is additionally shaped by Reddit community preferences, moderation, reposting behavior, and temporal trends at the time of collection. Some keyword groups (e.g., propaganda, misinformation, hateful memes) were used to increase semantic and stylistic diversity and should not be interpreted as claims about prevalence; conclusions drawn from this study should be interpreted as applying to the sampled distribution rather than the broader population of online images. Finally, because the data are drawn from public posts, we do not attempt to identify individuals or infer personal attributes; the study is used only to evaluate aggregate resynthesis vulnerability.

\section{Robustness to Semantic Attacks.}
\label{app:semantic}

We evaluate the A-index under a range of semantic transformations spanning photometric distortions (brightness, contrast, saturation, color shift), spatial transforms (crop, rotation, horizontal flip), compression artifacts (JPEG at quality 10 and 30), and additive noise (Gaussian noise and blur) applied to generated images. The calibrated safety threshold for SD3~(medium) is $\tau_{\text{safety}} = 0.0365$. A fake image can only evade detection if its A-index exceeds this threshold after transformation. Table~\ref{tab:semantic_attacks} reports results across these edits.

Across every photometric and spatial perturbation, the A-index of generated images remains well below $\tau_{\text{safety}}$. Spatial transforms produce the lowest scores: horizontal flip ($0.0045$), $20\%$ crop ($0.0067$), and $15^\circ$ rotation ($0.0074$). This is not coincidental: spatial distortions misalign the image with the generator's inversion prior, degrading both structural similarity (SSIM) and perceptual distance simultaneously, which pulls the A-index further from the threshold rather than toward it. Photometric attacks (brightness, contrast, color, saturation) similarly remain in the range $[0.0115,\, 0.0158]$, as color-space shifts do not improve a generated image's reconstruction consistency under the inversion pipeline.

The only attack that approaches the threshold is Gaussian noise ($\sigma = 25$), yielding an A-index of $0.0366$ marginally above $\tau_{\text{safety}}$ but still below the adversarially calibrated security threshold $\tau_{\text{security}} = 0.038$. Importantly, at this severity the \emph{perceptual quality degrades substantially} (SSIM $= 0.33$), which is the opposite of what a successful evasion attack must achieve: the slight A-index increase is driven by heavy noise disrupting pixel-level reconstruction rather than by the image becoming harder to authenticate. In other words, the only near-threshold case arises in a regime where the transformation itself introduces conspicuous quality loss, rather than improving plausibility of authenticity.

\begin{figure*}[!t]
    \centering
    \begin{subfigure}[b]{0.235\textwidth}
        \includegraphics[width=\linewidth]{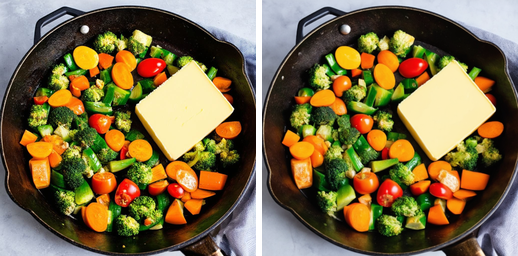}
        \caption*{\scriptsize Identity (control)}
    \end{subfigure}
    \hfill
    \begin{subfigure}[b]{0.235\textwidth}
        \includegraphics[width=\linewidth]{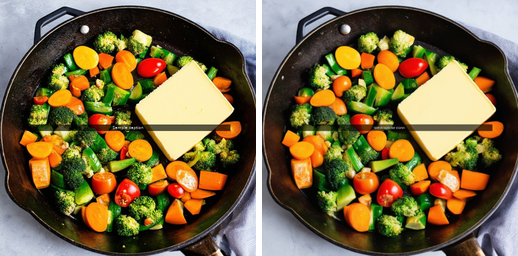}
        \caption*{\scriptsize Text 2\% center}
    \end{subfigure}
    \hfill
    \begin{subfigure}[b]{0.235\textwidth}
        \includegraphics[width=\linewidth]{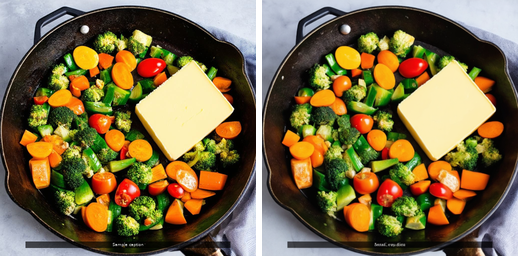}
        \caption*{\scriptsize Text 2\% bottom}
    \end{subfigure}
    \hfill
    \begin{subfigure}[b]{0.235\textwidth}
        \includegraphics[width=\linewidth]{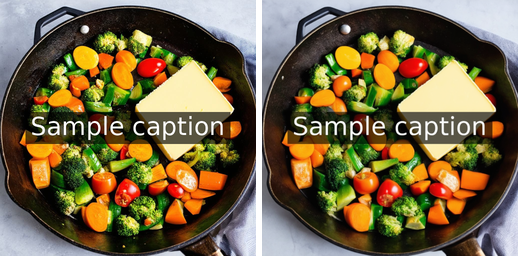}
        \caption*{\scriptsize Text 10\% center}
    \end{subfigure}

    \vspace{0.12cm}

    \begin{subfigure}[b]{0.235\textwidth}
        \includegraphics[width=\linewidth]{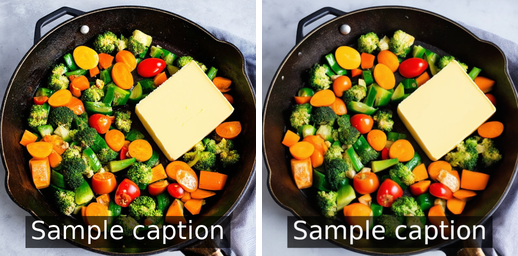}
        \caption*{\scriptsize Text 10\% bottom}
    \end{subfigure}
    \hfill
    \begin{subfigure}[b]{0.235\textwidth}
        \includegraphics[width=\linewidth]{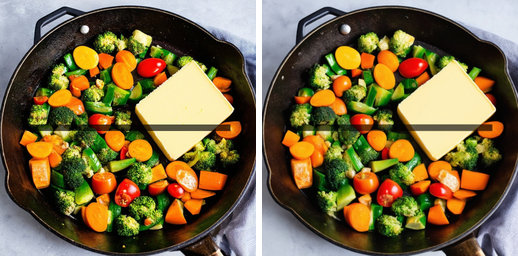}
        \caption*{\scriptsize Box 2\% center}
    \end{subfigure}
    \hfill
    \begin{subfigure}[b]{0.235\textwidth}
        \includegraphics[width=\linewidth]{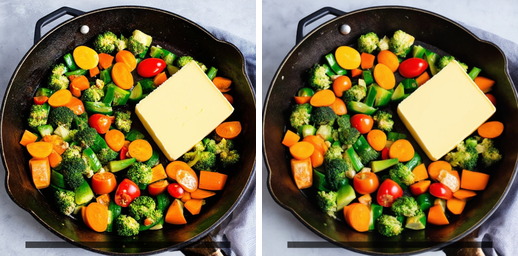}
        \caption*{\scriptsize Box 2\% bottom}
    \end{subfigure}
    \hfill
    \begin{subfigure}[b]{0.235\textwidth}
        \includegraphics[width=\linewidth]{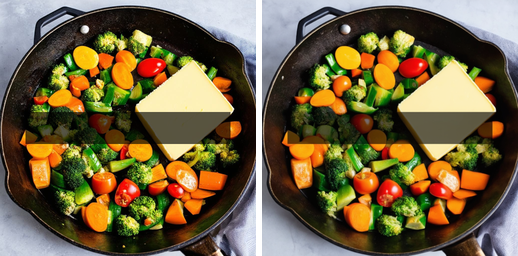}
        \caption*{\scriptsize Box 10\% center}
    \end{subfigure}

    \vspace{0.12cm}

    \begin{subfigure}[b]{0.235\textwidth}
        \includegraphics[width=\linewidth]{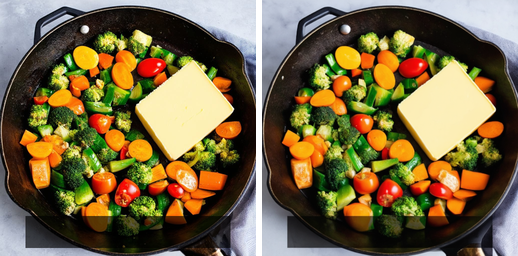}
        \caption*{\scriptsize Box 10\% bottom}
    \end{subfigure}
    \hfill
    \begin{subfigure}[b]{0.235\textwidth}
        \includegraphics[width=\linewidth]{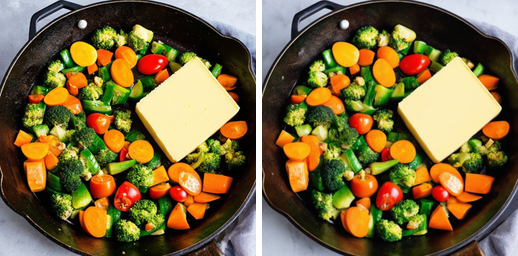}
        \caption*{\scriptsize Crop area 90\%}
    \end{subfigure}
    \hfill
    \begin{subfigure}[b]{0.235\textwidth}
        \includegraphics[width=\linewidth]{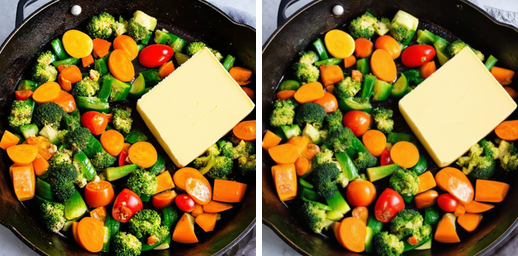}
        \caption*{\scriptsize Crop area 70\%}
    \end{subfigure}
    \hfill
    \begin{subfigure}[b]{0.235\textwidth}
        \includegraphics[width=\linewidth]{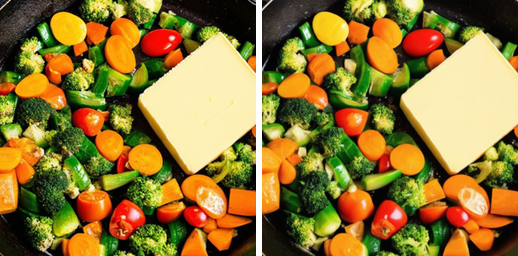}
        \caption*{\scriptsize Crop area 50\%}
    \end{subfigure}

    \vspace{0.12cm}

    \begin{subfigure}[b]{0.235\textwidth}
        \includegraphics[width=\linewidth]{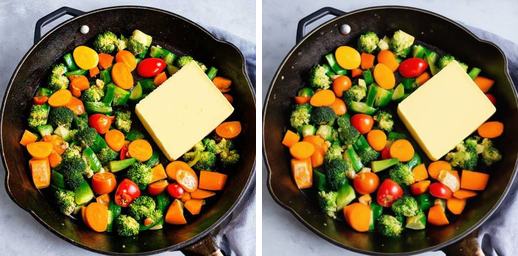}
        \caption*{\scriptsize JPEG q50}
    \end{subfigure}
    \hfill
    \begin{subfigure}[b]{0.235\textwidth}
        \includegraphics[width=\linewidth]{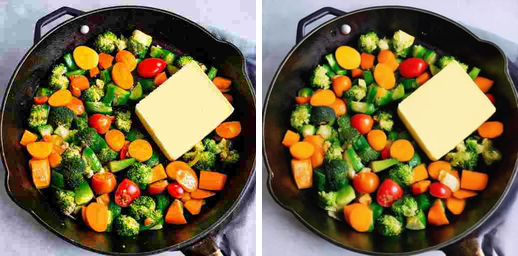}
        \caption*{\scriptsize JPEG q10}
    \end{subfigure}
    \hfill
    \begin{subfigure}[b]{0.235\textwidth}
        \includegraphics[width=\linewidth]{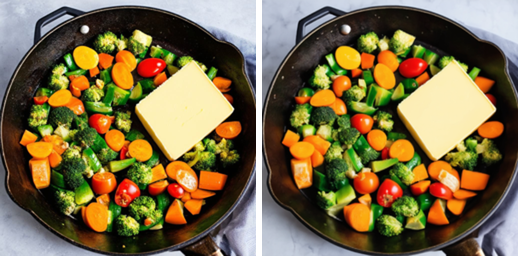}
        \caption*{\scriptsize Downsample $512{\to}256{\to}512$}
    \end{subfigure}
    \hfill
    \begin{subfigure}[b]{0.235\textwidth}
        \includegraphics[width=\linewidth]{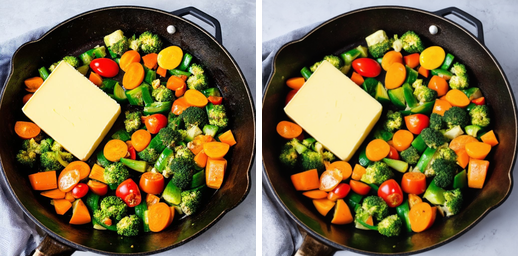}
        \caption*{\scriptsize Horizontal flip}
    \end{subfigure}

    \caption{\textbf{Qualitative examples of some transformations and their SD3 inversions.} Each panel shows the edited query on the left and its SD3 Medium inverted reconstruction on the right. Overlays, crops, JPEG compression, downsampling, and flipping leave the reconstruction faithful, so the A-index stays well below $\tau_{\text{safety}}$, consistent with \Cref{tab:semantic_attacks}.}
    \label{fig:sd3_edit_sweep_examples}
\end{figure*}

The underlying reason these attacks fail as evasion strategies is fundamental: semantic transformations do not change \emph{what the generator can synthesize}. A generated image, regardless of post-hoc distortion, was produced by the generator's own distribution and therefore inverts consistently. Our A-index measures inversion consistency a property tied to the image's origin under the specified inversion pipeline not surface-level statistics that can be manipulated by photometric transforms. In practice, pushing a generated image toward $\tau_{\text{safety}}$ via post-processing tends to require increasingly aggressive distortions that \emph{degrade quality substantially}, leading primarily to a degraded (and visibly corrupted) image rather than a meaningful bypass. These results confirm that the A-index is robust to this class of semantic attacks.
\Cref{fig:sd3_edit_sweep_examples} shows qualitative examples of some of these transformations and their inversions.

\begin{table}[!t]
\centering
\caption{A-index of generated SD3 images under semantic edits and transformations. The safety threshold is $\tau_{\text{safety}} = 0.0365$ and the security threshold is $\tau_{\text{security}} = 0.038$ (SD3 medium). Every edit remains below $\tau_{\text{safety}}$ except Gaussian noise, which remains below $\tau_{\text{security}}$.}
\label{tab:semantic_attacks}
\footnotesize
\setlength{\tabcolsep}{2.2pt}
\begin{tabular}{lcccc}
\toprule
\textbf{Attack} & \textbf{SSIM} & \textbf{CLIP} & \textbf{A-index} & \textbf{$\Delta\tau_s$} \\
\midrule
Horizontal flip     & 0.645 & 0.963 & 0.0045 & $-$87.7\% \\
Crop 20\%           & 0.638 & 0.916 & 0.0067 & $-$81.6\% \\
Rotation 15$^\circ$ & 0.607 & 0.926 & 0.0074 & $-$79.7\% \\
Crop 10\%           & 0.671 & 0.948 & 0.0086 & $-$76.4\% \\
Brightness $\downarrow$ (0.5$\times$) & 0.690 & 0.942 & 0.0115 & $-$68.5\% \\
Brightness $\uparrow$ (1.5$\times$)   & 0.810 & 0.946 & 0.0132 & $-$63.8\% \\
Color shift         & 0.845 & 0.935 & 0.0135 & $-$63.0\% \\
Downsample $512{\to}256{\to}512$ & 0.756 & 0.979 & 0.0143 & $-$60.8\% \\
Contrast $\downarrow$ (0.5$\times$)   & 0.813 & 0.933 & 0.0144 & $-$60.6\% \\
Saturation $\uparrow$ (2.0$\times$)   & 0.859 & 0.964 & 0.0151 & $-$58.6\% \\
Text 10\% bottom    & 0.693 & 0.981 & 0.0152 & $-$58.4\% \\
Box 2\% bottom      & 0.675 & 0.977 & 0.0155 & $-$57.5\% \\
Gaussian blur       & 0.847 & 0.946 & 0.0156 & $-$57.3\% \\
Box 10\% bottom     & 0.685 & 0.976 & 0.0156 & $-$57.3\% \\
Box 2\% center      & 0.672 & 0.977 & 0.0156 & $-$57.2\% \\
Text 10\% center    & 0.683 & 0.979 & 0.0157 & $-$56.9\% \\
Contrast $\uparrow$ (1.5$\times$)     & 0.819 & 0.954 & 0.0158 & $-$56.7\% \\
Text 2\% bottom     & 0.675 & 0.974 & 0.0159 & $-$56.5\% \\
Text 2\% center     & 0.671 & 0.973 & 0.0160 & $-$56.1\% \\
Box 10\% center     & 0.675 & 0.972 & 0.0161 & $-$55.8\% \\
Saturation $\downarrow$ (0.3$\times$) & 0.858 & 0.945 & 0.0166 & $-$54.5\% \\
JPEG Q=30           & 0.865 & 0.953 & 0.0178 & $-$51.2\% \\
JPEG Q=10           & 0.851 & 0.881 & 0.0272 & $-$25.5\% \\
\midrule
Gaussian noise ($\sigma$=25) & 0.330 & 0.909 & 0.0366 & $+$0.3\%$^\dagger$ \\
\bottomrule
\end{tabular}
\vspace{2pt}
\begin{flushleft}
\footnotesize $^\dagger$ Marginally exceeds $\tau_{\text{safety}}$ but remains below $\tau_{\text{security}} = 0.038$. The elevated score is caused by noise disrupting pixel-level reconstruction (SSIM $= 0.33$), not by increased inversion fidelity.
\end{flushleft}
\end{table}

\section{Additional Video Evaluation Details}
\label{app:video_details}

\tit{Dataset}
Deepfake-Eval-2024 contains recent in-the-wild videos collected from social-media platforms and deepfake-detection services~\cite{chandra2025deepfakeeval2024multimodalinthewildbenchmark}.
The benchmark authors report that open-source video detectors lose approximately $50\%$ AUC relative to earlier academic datasets.
We sample a balanced subset of $100$ videos, containing $50$ authentic and $50$ generated videos, and evaluate only their visual content.

\tit{Frame Sampling}
We sample eight frames from each video, with consecutive samples separated by $30$ frames.
Each frame is independently captioned, inverted, reconstructed, and scored using the image pipeline.
We average the eight frame-level scores to obtain the video-level score in \Cref{eq:video}.
This aggregation does not model motion, temporal consistency, frame transitions, or dependencies between neighboring frames.

\tit{Video Baselines}
GenConViT~\cite{genconvit} combines ConvNeXt and Swin Transformer features with autoencoder and variational-autoencoder components.
FTCN~\cite{ftcn} uses temporally extended convolutions and a Temporal Transformer to model relationships across frames.
StyleFlow~\cite{styleflow} represents changes in facial style-latent vectors with a StyleGRU and combines them with content features.
We evaluate all three methods using their original inference code, default thresholds, and publicly available weights without fine-tuning on Deepfake-Eval-2024.

\tit{Scope of the Comparison}
\Cref{tab:video_baseline_failures} reports threshold-free ranking and binary classification metrics for the three video baselines.
\Cref{fig:video_graph} instead measures whether frame-aggregated reconstruction scores retain the ordering observed for images.
We do not claim a direct numerical improvement because the video experiment lacks a separately calibrated threshold and therefore does not report TPR at $1\%$ FPR.
A direct comparison requires generated calibration videos, a video-specific threshold, and evaluation at the same operating point used for images.

\tit{Runtime}
The frame-averaging procedure invokes the image pipeline eight times for each video and generator.
At $11.67$ seconds per frame and generator, sequential evaluation requires approximately $93.36$ seconds per video and generator on one NVIDIA RTX 5000 Ada.
Frames and generators can be processed in parallel, but the total computation grows linearly with both quantities.

\tit{Limitations}
The subset contains only $100$ videos and cannot establish trends across generator families, release dates, content categories, or post-processing histories.
The method does not analyze temporal consistency, optical flow, motion trajectories, synchronization, or neighboring-frame dependencies.
Deepfake-Eval-2024 also contains audio, but our evaluation excludes this channel and makes no multimedia-authentication claim.
Extending the framework to audio requires an invertible audio generator, audio-specific similarity metrics, and separately calibrated safety and security thresholds.

\section{Limitations}

Despite its strengths, our approach has several limitations that highlight directions for future research.

First, the authenticity index relies on access to high-quality inversion pipelines and perceptual similarity models such as CLIP and LPIPS. While this enables a robust and calibrated score, it assumes white-box or partially open generative models. In scenarios where the generator is fully black-box or proprietary, the inversion step may not be feasible or reliable.

Second, the safety and security thresholds are specific to each generative model and require calibration using real and synthetic data. This introduces practical overhead, especially in dynamic environments where new models are rapidly emerging. Automating or generalizing this calibration remains an open challenge.

Finally, while our extension to the video domain is promising, it treats each frame independently and does not exploit temporal consistency or motion cues. Incorporating these temporal dynamics could improve performance in challenging video-based deepfake scenarios.

\end{document}